%% file: sample-sigconf.tex
\documentclass[sigconf]{acmart}
\renewcommand\footnotetextcopyrightpermission[1]{} 
\usepackage{booktabs} 
\usepackage{pgf} 
\usepackage{subfig} 
\usepackage{paralist}
\usepackage{placeins} 

\begin{document}
\title{GitGraph - Architecture Search Space Creation through Frequent Computational Subgraph Mining}


\author{Kamil Bennani-Smires}
\affiliation{%
  \institution{Artificial intelligence and machine learning group, Swisscom}
}
\email{kamil.bennani-smires@swisscom.com}

\author{Claudiu Musat}
\affiliation{%
  \institution{Artificial intelligence and machine learning group, Swisscom}
}
\email{claudiu.musat@swisscom.com}

\author{Andreea Hossmann}
\affiliation{%
  \institution{Artificial intelligence and machine learning group, Swisscom}
}
\email{andreea.hossmann@swisscom.com}

\author{Michael Baeriswyl}
\affiliation{%
  \institution{Artificial intelligence and machine learning group, Swisscom}
}
\email{michael.baeriswyl@swisscom.com}


\begin{abstract}
The dramatic success of deep neural networks across multiple application areas often relies on experts painstakingly designing a network architecture specific to each task. To simplify this process and make it more accessible, an emerging research effort seeks to automate the design of neural network architectures, using e.g. evolutionary algorithms or reinforcement learning or simple search in a constrained space of neural modules.

Considering the typical size of the search space (e.g. \textasciitilde$10^{10}$ candidates for a $10$-layer network) and the cost of evaluating a single candidate, current architecture search methods are very restricted. They either rely on static pre-built modules to be recombined for the task at hand, or they define a static hand-crafted framework within which they can generate new architectures from the simplest possible operations.

In this paper, we relax these restrictions, by capitalizing on the collective wisdom contained in the plethora of neural networks published in online code repositories. Concretely, we \begin{inparaenum}[(a)] \item extract and publish GitGraph, a corpus of neural architectures and their descriptions; \item we create problem-specific neural architecture search spaces, implemented as a textual search mechanism over GitGraph; \item we propose a method of identifying \textbf{\textit{unique}} common subgraphs within the architectures solving each problem (e.g., image processing, reinforcement learning), that can then serve as modules in the newly created problem specific neural search space\end{inparaenum}.




\end{abstract}

%
%


\maketitle

\input{samplebody-conf}

\bibliographystyle{ACM-Reference-Format}
\bibliography{sample-bibliography} 

\input{appendix}

\end{document}

%% file: samplebody-conf.tex

\section{Introduction}

The work of a deep learning practitioner is more repetitive than we care to admit. A lot of the automation drive has focused on reducing the amount of repetitive work office workers do. Typically, however, this has not included AI researchers or data scientists. We automate the repetitive work of the customer support agent \cite{goldberg1999automated} who gives the same advice to hundreds of customers in need of a password reset. In a similar fashion, we should help the deep learning practitioner to avoid designing the same type of neural architecture every time they need to turn one type of sequence into another. While this is now a laborious manual process, attempts to automatically build new architectures from basic components are increasingly gaining traction.

Current automated neural architecture creation strategies rely on extensive expert knowledge and heavy handed supervision. They either use predefined modules \cite{deeparch} and the novelty lies in the recombination or they create new modules but within a very tightly controled structure \cite{google_reinf,Such2017a}. 
The reason for this heavy-handed supervision is that each step taken towards a better architecture is costly. This constraint is independent of the search method used. Whether it's employing reinforcement learning \cite{google_reinf,mit_reinf} or evolutionary algorithms \cite{Such2017a}, for each change the system must evaluate candidates and each evaluation means training a full network on a usually complex task. The smaller the changes, the more candidates need to be evaluated. 
The space of possible options is too large to allow searching or evolving a full architecture from basic building blocks like matrix additions or multiplications. Shortcuts are thus necessary. 

Neural evolution can be seen as a combination of two problems - defining a neural module search space and creating a policy to create that space. The question of finding the right policy has received almost all the community's attention \cite{deeparch,Such2017a,google_reinf}, with the search space receiving almost none. A notable exception is a recent work on \cite{Schrimpf2017}  that explicitly states that different domains require different operators, that are subsequently combined to form neural architectures.

We propose constructing the search space by using the known architectures for similar tasks.
Expert supervision can guide the search and lower the network creation cost. In our view, however, this supervision need not be a laborious task linked to the task at hand. Instead, it can come from repositories of computation graphs that have been published for the tasks similar to this one before. This removes the need for handcrafted constraints built into the evolution task itself. It also leverages a previously unused resource - the network structures published by previous researchers. 
\begin{figure*}
\begin{center}
\includegraphics[width=\textwidth]{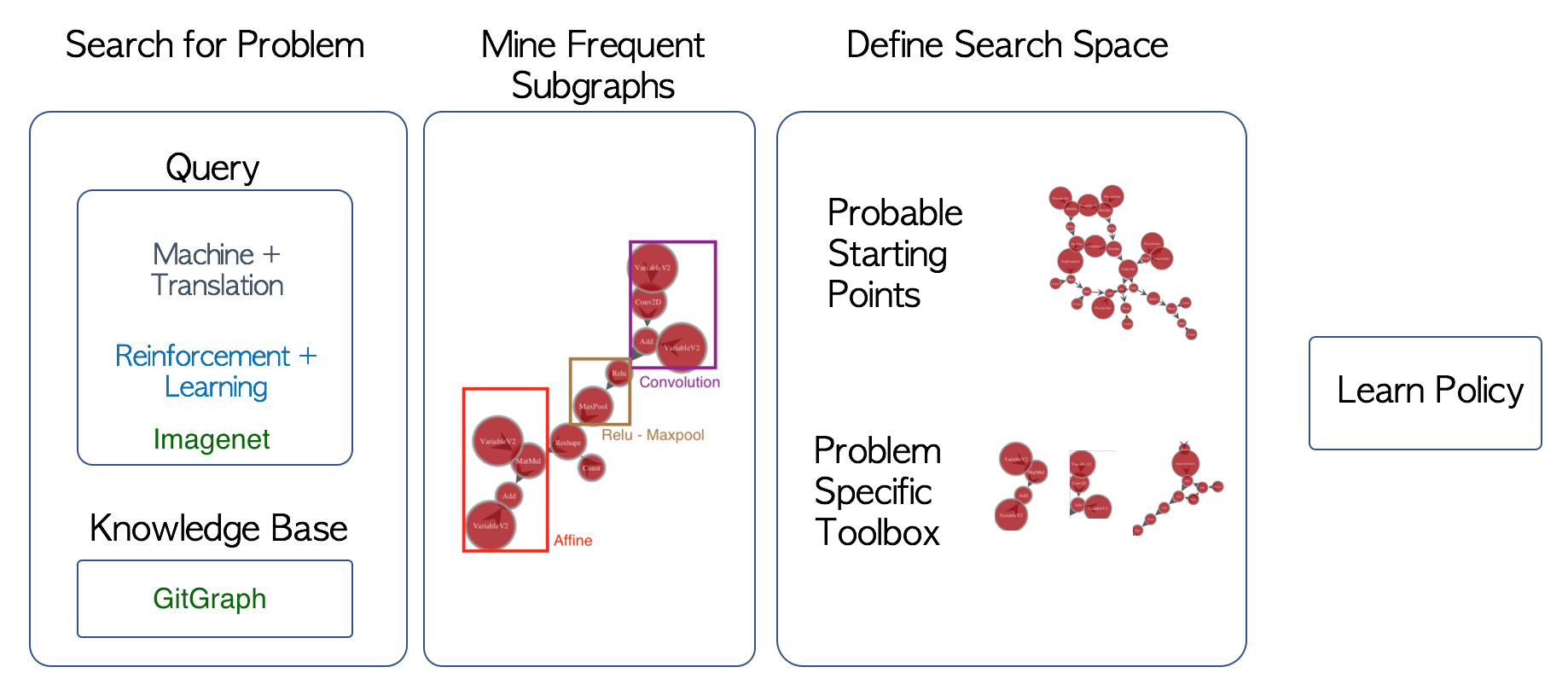}
\caption{Automated Architecture Search Space Definition}
\label{fig-intro}
\end{center}
\end{figure*}

As shown in figure \ref{fig-intro}, we split the task of search space definition into three parts:
\begin{itemize}
	\item {Search} for architectures that solve similar problems. This step yields a collection of graphs.
	\item {Common Subgraph Mining}. Extract the neural modules and combinations of modules that are common between the found architectures. As shown in the corresponding task in Figure \ref{fig-intro}, we can directly mine combinations like \textit{convolution + Max pool + affine}.
	\item {Defining the Search Space} by specifying which modules are large, frequent and unique enough to be useful. These subgraphs then become a toolbox of task-oriented modules. 
\end{itemize}
The resulting task specific neural module toolbox becomes the starting point to evolve new architectures.

 A great \textbf{example}, present in \ref{subfreq_example}, regards the common task of treating images inputs. We found that there is a subgraph (also shown in Figure \ref{conv_example}) that appears at least once in 30\% or more of the distinct computational graphs for image-related tasks. This subgraph is actually a combination of known modules. It is a convolution operation followed by a Relu activation and Maxpooling, the result is reshaped and eventually an affine transformation is performed (weight multiplication and bias addition). This chain of operations is "typical" when dealing with images.

 The advantage of finding the computational graphs and the common components for a task is twofold. Graphs that resemble many others for the same task have a high likelihood of being a good starting point in searching for a new architecture. In addition, common components can be transformed into modules that can be used as changes when searching for an architecture. We thus go into a supervised learning approach, where we use the knowledge of humans who manually devised the architectures previously to create new ones.

 In this paper we introduce GitGraph \footnote{\url{https://www.mycloud.ch/s/S00E8129370EFE75830040072AD8203611E4F9971E1}} - a dataset of TensorFlow \cite{tensorflow} computational graphs, alongside with the description of tasks they are useful for.
 In addition to publishing the dataset, we make three contributions:
 \begin{itemize}
 	\item We show that GitGraph can be used to search for a problem and define the neural search space. We show that there are enough distinct computational graphs for common problems to make connections between them and extract common components.
 	\item We propose a method to identify frequent neural subgraphs.
 	\item We show that by using common subgraphs as modules, we can reduce the complexity of the resulting architectures by up to 70\%.
 \end{itemize}

\input{related_work}

\input{crawling_section}
\input{reduction}

\input{conclusion}

%% file: related_work.tex
\section{Related Work}

\subsection{Architecture Search}

A standard way of searching for a neural architecture is to build one from nothing.\cite{google_reinf} use reinforcement learning build convolution stacks for image-related problems and recurrent networks for text-related ones. 
In \cite{google_reinf} , they encoded the architecture of a neural network as a string and used an RNN (the controller) to sample architectures. This RNN is trained using REINFORCE \cite{Williams:1992:reinforce} in order to maximize the expected reward (accuracy on validation dataset) of the architectures generated. Their method was applied in order to generate a CNN architecture and a reccurent cell by creating by hand one appropriate search space for each task.
A known problem is the cost of the search, determined mostly by the number of candidates that need to be evaluated. \cite{google_reinf} limit the number of possibilities by imposing a rigid structure (e.g. a stack of convolutions). In addition, they do not fully train the candidate networks. With these constraints, they showed that state of the art results could be obtained with learnt networks, but with a computational cost 10.000 times higher than training a single network.

MetaQNN \cite{mit_reinf} also generates CNN architectures based on reinforcement learning.
The layer selection process is modeled as a Markov Decision Process MDP , where each is state is defined as tuple of relevant layer parameters such as the type of the layer (Convolution , Pooling , Fully connected, Global Average Pooling, SoftMax) associated with some parameters dependant on the type (e.g. number of neurons in a fully connected layer, stride for convolution), the optimal architecture is found using Q-learning with an epsilon-greedy strategy. Finally they sampled from the optimal policy which is not deterministic (stopping at epsilon=0.1) 5 models and ensemble them to make the predictions.
\cite{mit_reinf} rely on the search space when doing comparison with other networks : better performance on networks that used only the same components as they have in their search space.

\cite{Such2017a}, following in the footsteps of \cite{Salimans2017}, replace reinforcement learning with evolution strategies. The latter are shown to be a viable alternative to gradient descent in neural architecture creation. Neuroevolution passes from learning weights of a predefined architecture to learning the links between modules and the weights attached to them jointly. However, it still relies on the existence of modules that can be joined.
The reinforcement or genetic evolution of architectures have the advantage of coverage - if there is an architecture that is superior for a task but unbeknown to the research community at this point, there is hope that it will be found.

A distinct advantage of evolving using borrowed modules is their improbable nature. As shown by NEAT \cite{Stanley:2002:ENN:638553.638554}, there is a tendency for new architectures to go extinct from an evolving population before they could realize their potential. We believe this effect can be countered by adding whole blocks to the net, instead of their individual components.
In addition, neuroevolution has moved from direct to indirect gene encodings, because of the growing size of the networks. As shown in \cite{Stanley:2007:CPP}, the number of genes can be much lower than the number of connections and neurons in the brain. A compositional pattern producing network (CPPN) compresses a pattern with regularities and symmetries into a relatively small set of genes. By constructing a toolbox of task-oriented neural modules, we also dissociate the number of mutations needed from the number of actual connections in the resulting ANN. 

Instead of building networks from basic building blocks like additions and multiplications, the Neural Architect \cite{deeparch} searches a space large, prebuilt modules to achieve the same outcome. The downside of this approach is that the toolbox is universal, with the same Relu, conv or affine layers used irrespective of the task. The addition of a module that is useful to the task at hand can only be done manually. Moreover, a data scientist or researcher's knowledge is needed to define which modules are useful and how they can be combined.
The Neural Architect is much faster than the methods above, because of trying a small number of candidates to get to a competitive final result. 

A work that prioritizes the search space over the search policy is the recently introduced search \cite{Schrimpf2017} based on domain specific languages.
Like the Neural Architect, Salesforce \cite{Schrimpf2017} introduced a novel way of searching for neural architectures relying on human supervision. The main improvement of their approach is the creation of a domain specific language (DSL). This contrasts strongly to the standard toolbox that the Neural Architect uses and allows the creation of architectures with modules that are especially created for that task. They study the case of recurrent nets and show that a very simple domain specific search space  leads to good architectures. For instance, when defining a DSL for recurrent networks, it will contain 4 unary operators, 2 binary operators, and a single ternary operator. This very specific choice is given by the researchers' prior knowledge of the structure of GRUs \cite{gru_original} and LSTMs \cite{lstm_original}. 
The intuition is that, once the search space is created, finding the right search policy is feasible - the right quantity and combination of the modules can be found with ease. 
We thus focus on the creation of the initial search space, with the goal of automating this initial step that is, for now, a human prerogative.


\subsection{Frequent Subgraph Mining}

To find the frequent neural modules for a chosen problem, we employ methods typically used for subgraph mining.
Given a graph dataset $D = {G_0, G_1,….G_n}$, $support(g)$ denotes the number of graphs in D in which g is a subgraph. The problem of frequent subgraph mining is to find any subgraph $g$ s.t. $support(g) >= minSup$.

A well-known method is graph based Substructure pattern mining, 
gSpan \cite{gspan}. It finds all the frequent subgraphs without candidate generation and false positive pruning. Relying on Depth-First search , it introduces two novel techniques DFS lexical order and minimum DFS Code which makes the frequent subgraph mining task solved efficiently.

Compared to gSpan \cite{gspan}, CloseGraph \cite{Yan:2003:CMC:956750.956784}  aims to mine \textbf{closed} frequent subgraphs. A graph g is closed in a database if there exists no proper super graph of g that has the same support as g. Mining closed subgraph helps to get rid off redudant subgraphs. The search space is pruned by introducing two novel concepts: equivalent occurence and early termination on top of gSpan\'s concepts. This method is also highly effective it has been shown by the authors that it outperforms gSpan \cite{gspan} by a factor of 4 to 10 when the frequent subgraphs are large.

\subsection{Code Corpora}

We apply the subgraph mining methods on a corpus of computational graphs published on GitHub \footnote{www.github.com} - GitGraph. While this is the first corpus of its kind, GitGraph is conceptually similar to existing code corpora from two points of view. Firstly - just like code corpora, we are scraping a corpus from a publicly available resource and thus we have to verify its properties like the amount of duplicate contents. Secondly, the graphs themselves are derived from Tensorflow code and thus, if a generic compiler becomes available, a computational graph corpus could be obtained from the original Tensorflow code. This would reduce the graph corpus building to a code corpus building problem.

We have a keen interest in spotting duplicated graphs, as this may affect the statistics extracted about the frequency of subgraphs.
In \cite{Schwarz:2012:OCC:2337223.2337398} they show that a large amount of code is duplicated on several projects coming from the same open-source software eco-system. The importance of this phenoma yield to different techniques that had been developed to detect code duplication, These methods can be token-based such as \cite{DBLP:journals/corr/SainiSKL16} or rely on abstract syntax trees \cite{clone_detection} or hash-based \cite{Schwarz:2012:OCC:2337223.2337398}.

%% file: crawling_section.tex

\section{The GitGraph Corpus}

\subsection{Corpus Creation}

\begin{figure*}
\begin{center}

\subfloat[With duplicated graphs]{
\input{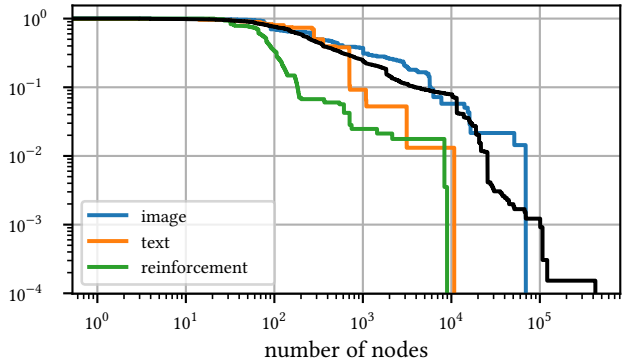}
}
\subfloat[Without duplicated graphs]{
\input{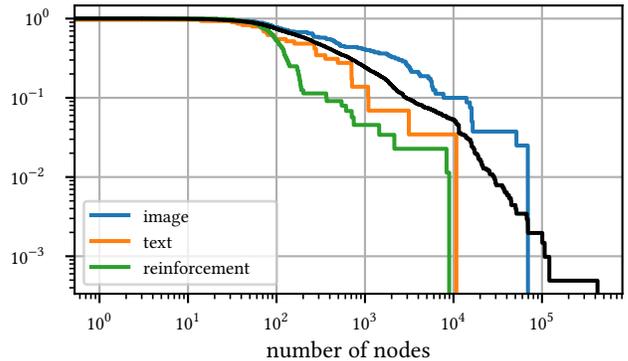}
}
\caption{Distribution of nodes in the graphs per task}
\label{fig-dedup-task}
\end{center}
\end{figure*}

The first  contribution of this paper is the creation of a searchable database of computational graphs along with a description of the problem they solve. We checked out Github repositories of neural networks written in Tensorflow \footnote{www.tensorflow.org}. An observation led to a quick progress in the corpus creation - creating automated compiling routines is not necessary in a first instance. We can instead obtain the computational graphs from the checkpoints \(.ckpt files\) that the authors did not include in their .gitignore . 
We then stored the graphs, alongside with the descriptions in the \textit{readme} files and the github descriptions in a non relational database, in this case MongoDB. We use the search functionality of the database to retrieve the graphs that are linked to specific problems or techniques, like \textit{reinforcement learning}.
In its current version, GitGraph contains 6863 graphs in total, coming from 1449 repositories, for an average of 4.73 graphs per repository.

A node in Tensorflow graph contains the operation performed, from a set of standard operation such as addition, matrix multiplication or convolution. It may also contain additional information depending on the type of operation such as certain hyperparameters values.
The contents of the Tensorflow checkpoints cannot are not compatible with the graph libraries used afterwards \cite{Yan:2003:CMC:956750.956784,gspan}. A preprocessing phase is needed, where the graph definitions are extracted from these files and cleaned.
We convert the checkpoint contents to Graph-tool \footnote{\url{https://graph-tool.skewed.de}} graphs. Graph-tool is a python library for graph manpilation where the core algorithms and data structures are written in C++ for fast computations.

The distribution of nodes is shown by the black "all" curve in Figure \ref{fig-dedup-task} a. We can observe that most architectures contain between $10^2$ and $10^3$ nodes, with the smallest 20\% having less than $10^2$ and the largest 20\% more than $10^3$ nodes.

\subsubsection{Deduplication and Limitations}
Many of the graphs are duplicates, due to multiple checkpoints for the same model and forked repositories. To counter the effect that these duplicated graphs will have on the subsequent analysis, we perform a deduplication step.
We only remove exact graph duplicates. To assess equality between two nodes, we only factor in the type of operation, not any possible additional information like hyperparameter values.

A fully duplicated architecture distorts the results when mining frequent subgraphs. From a different point of view, it can be interesting, since we believe people will tend to duplicate good architectures. We do not explore this in the current work. 
Even after the deduplication phase, it is still possible for two distinct graphs to the same behavior. For example, stacking two times an Affine Layer \(matrix multiplication + addition of bias\), without an activation will result in the same behavior as only one affine layer. We do not include this type of duplication in the analysis.

The deduplication leads to a subset of 2033 unique graphs from the original 6863.
We see the result of the deduplication in figure \ref{fig-dedup-task} b.
Since the number of nodes varies wildly, we opt for a complementary cumulative distribution function (CCDF) curve. The values on the curves show the proportions of the graphs for which the number of nodes exceeds a given threshold, represented by the value on the X axis. 
We can observe from the difference between Figures \ref{fig-dedup-task} a and b that, for the black curve, corresponding to all the graphs, the shape of the CCDF curves stays almost unchanged when the graphs are deduplicated. 

\subsection{Subgraph Mining Scope Definition}

\textbf{Defining the scope} of the problem. If, for instance, someone is interested in \textit{machine translation for Swiss German}, we may not find any prior researchers who tackled that specific problem. A reasonable assumption is that neural architectures made for \textit{machine translation} or, in the best case for \textit{machine translation German} are similar to the given task.

To attract the interest of a varied audience, we focus on three tasks, from three different domains of machine learning: \textit{image processing}, \textit{text processing} and \textit{reinforcement learning}. GitGraph contains:
\begin{itemize}
\item For image : 139 graphs with duplicates , 80 without.
\item For text : 77 graphs with duplicates, 29 without.
\item For reinforcement : 283 with duplicates, 88 without.
\end{itemize}

We visualize the distribution of nodes in graphs linked to the three chosen tasks before and after the deduplication step in figure \ref{fig-dedup-task} a. and b.
The most important observation that can be drawn from figure \ref{fig-dedup-task} a and b is that the shapes are remarkably similar. Removing the duplicates had no major impact on the node distribution of the graphs.

Unsurprisingly, the largest graphs are the ones that deal with image processing. Text processing comes second, with the smallest number of nodes being found in reinforcement learning tasks.
An important element that is visible \ref{fig-dedup-task} b. in the \textit{text} task is that there are graphs with 0 nodes. This is more visible after the deduplication because they represent a higher proportion of the total number of graphs.

We observe an interesting behaviour - for image oriented architectures, after deduplication we have a higher proportion of large (more than $10^3$ nodes). This shows that for every day tasks the large architectures like Resnet may prove too much and researchers favor smaller ones.

\subsection{Graph Cleaning}

After defining the datasets of unique graphs for each task, we perform a sequence of preprocessing steps in order to focus only on the core neural architecture. 
We remove the auxiliary discernible components in order to make the graph lighter in terms of nodes. We remove nodes that are not useful for the solving the central problem, but are instead used for connected tasks. We thus:

\begin{itemize}
\item remove all nodes created by the optimizer including all the subsequent gradient computation nodes. 
\item remove all nodes concerning saving/restoring variables as well as summarization (visualisation on the tensorboard).
\item remove the nodes used to initialize a variable
\item reduce multiple nodes pertaining to a variables definition to a single one. We remove assign and identity nodes, and forward the edges directly to the variable node.
\end{itemize}

By doing so, we ensure that the subgraphs that are common to the graphs resulting from the reduction phase are useful in computing the core elements of the task. In the absence of reduction, a common problem is that the common subgraphs are actually outside the core. Instead of finding elements that can be then recombined into a better atchitecture for the same task, we would mix core with non-core modules, making the automated learning process slower and more cumbersome instead of more streamlined.

%% file: reduction.tex

\section{Frequent Subgraphs}

The second contribution of this work is a method of mining reusable neural modules. We focus on subgraphs that are repeated in the published graphs in GitGraph.
We define a common subgraph, for a certain task, as one that appears in the graphs made for that task more than a given threshold.
As in the deduplication phase, we only employ the type of operation within the node when computing the node equalities.

We define a subgraph that is common at $\tau\%$ for task \textit{T} as one that appears in a \textit{minimum} of $\tau\%$  of the graphs for task \textit{T}. The higher the maximum threshold of a subgraph, the more likely it is that subgraph is actually relevant for the given task.

\begin{figure*}
\begin{center}
\subfloat[Before reduction]{
\input{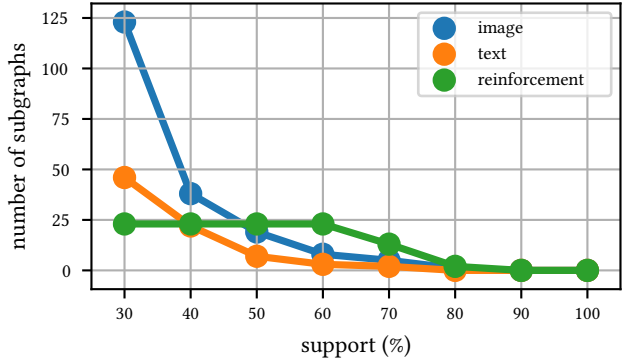}
\label{nbfreqsub-before-reduc}
}
\subfloat[After reduction]{
\input{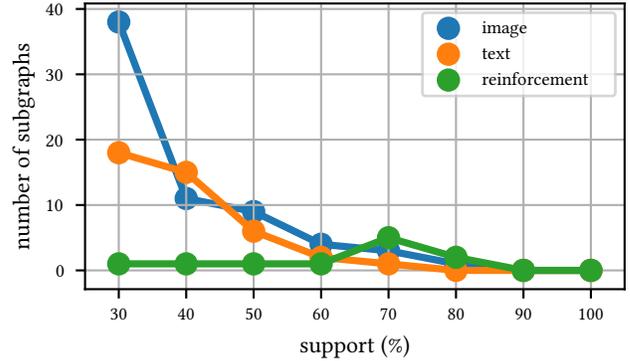}
\label{nbfreqsub-after-reduc}
}
\caption{Number of distinct frequent subgraphs given support}
\end{center}
\end{figure*}

In Figure \ref{nbfreqsub-before-reduc} we portray the number of distinct frequent subgraphs, for each support $\tau$ and each example task. For low $\tau$ values the number of frequent subgraphs is very high and uninformative. We thus plot the subgraph count for $\tau$ values above 30\%.
While for image and text processing we notice the expected steady decrease in the number of common subgraphs with the increase of $\tau$, for reinforcement we notice that there are exactly 25 common subgraphs for any $\tau$ below 70\%. This high similarity between various distinct reinforcement architectures is explained by multiple forking of the same repository. 

\subsection {Subgraph Mining and Matching}
\label{sec-mining}

The biggest common subgraphs are more informative than smaller ones contained within them. The main goal of GitGraph is to find neural components that reduce the complexity of the space when creating neural evolution or search policies. We thus eliminate the small common subgraphs that are contained within bigger ones that are still common given the same value of $\tau$.

CloseGraph is a GSpan \cite{gspan} option that ensures that the subgraphs we mine are distinct and not mere variations of the same one. The functioning of CloseGraph is the following. Let A-B-C be the notation corresponding to a subgraph where node A is connected to node B and node B is connected to node C. If, within a set of graphs, the subgraph A-B-C is present $k_{ABC}$ times and graph A-B-C-D is present $k_{ABCD}$ times, with $k_ABC=k_{ABCD}$, then A-B-C is always present only in A-B-C-D. If a subgraph is only present inside a bigger subgraph, then only the biggest subgraph is returned. 
If $k_{ABC}>k_{ABCD}$, then A-B-C is also present outside of A-B-C-D $k_{ABC-ABCD}$ times. If $k_{ABC-ABCD}>\tau$ then we also consider A-B-C a subgraph on its own. 
In order to this, once we have all the frequent subgraph fetched by gSpan, we counted in how many graphs they appear (unique count)  without considering them if they appear in a bigger subgraph returned by gSpan. The unique count $k_{ABC}-k{superset ABC}$ is then compared to the threshold to determine if the subgraph should be included in the returned set.

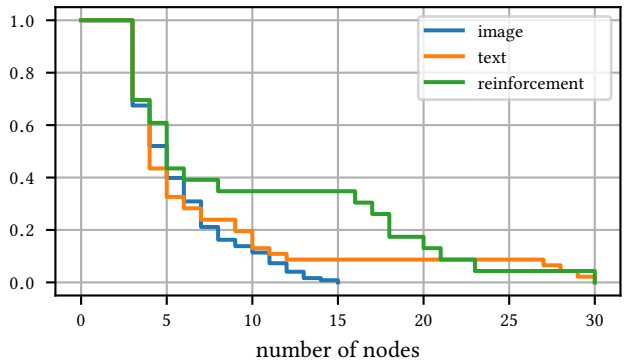
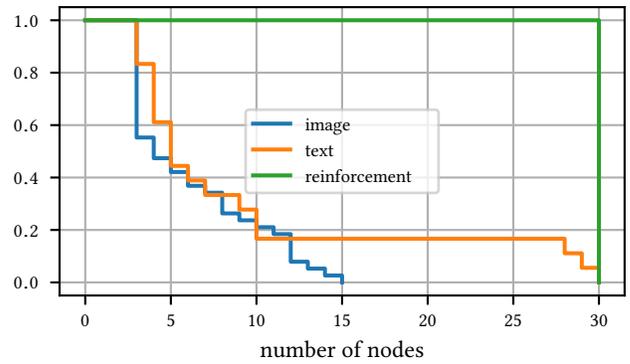
\begin{figure*}
\begin{center}
\subfloat[Before reduction]{
\input{figures/paper/sub_nodes_before_reduc_ccdf.pgf}
\label{sizefreqsub-before-reduction}
}
\subfloat[After reduction]{
\input{figures/paper/sub_nodes_ccdf.pgf}
\label{sizefreqsub-after-reduction}
}
\end{center}
\caption{Size of frequent subgraphs (in no of nodes) (min support = 30\%)}
\end{figure*}

In figure \ref{nbfreqsub-after-reduc} we portray the number of distinct frequent subgraph after the subgraph reduction step
for each support $\tau$ ranging from 30 to 100\%. For image and text processing, the figure shows an expected decrease in the number of subgraphs - from 123 to 38 for images and from 46 to 18 for text.

In the special case of reinforcement - with $\tau=30\%$ we obtain 23 frequent subgraphs but only 1 remains after the subgraph reduction - that is reused in almost all of the others.
A counterintuitive aspect that occurs at $\tau=70\%$ is that the number of subgraphs increases from 1 to 5. 
This means that the only common subgraph for $\tau<=60\%$ actually occurs in less than $\tau=70\%$ of all the reinforcement graphs.
However, $5$ parts of it are common for $\tau=70\%$ of the graphs, which explains the unexpected jump in the number of common subgraphs, with the increase of $\tau$.
These observations are thus indicative of a low level of architectural changes in reinforcement-oriented graphs.

\subsection {Subgraph Size}

The frequent subgraphs are meaningful, large chains, containing tens of nodes. If they are replaced with single nodes, they can lead to a significant reduction of the complexity of the network. We thus do not show common subgraphs smaller than $3$ nodes.

In figure \ref{sizefreqsub-before-reduction} and \ref{sizefreqsub-after-reduction} we show the number of nodes in the frequent subgraphs, before and after the subgraph reduction respectively. Each CCDF curve shows the percentage of distinct frequent subgraphs mined for this task with the size exceeding the threshold on the abscissa.

The differences between the pairs of CCDF curves portray the effect of the subgraph reduction. From the reinforcement curve in Figure \ref{sizefreqsub-after-reduction} we notice that the size of the common reinforcement subgraph discussed in the previous section is 30 nodes. This subgraph is shown in Figure \ref{fig-example-reinforcement}. Without the reduction phase, this subgraph would not have stood out so evidently.

For image and text analyses, roughly half of the frequent subgraphs have a size of between 3 and 5 nodes. The other half is between 5 to 15 for image and 5 to 30 for text. 
We are primarily interested in the larger graphs, since a simple operation in Tensorflow can spawn multiple nodes. For example, adding two constants create 3 nodes: one node for each constant + one node for the addition operator.
For \textit{text}, we notice that roughly 20\% of the reduced common subgraphs are actually very long - having more than 25 nodes. A manual analysis of these nodes shows that they correspond to commonly used recurrent units, like the LSTM in Figure \ref{fig-example-LSTM}.

\subsection {Graph Compositionality and Complexity Reduction}

We show that the subgraphs mined with the methods presented in section \ref{sec-mining} appear a large number of times in the original graphs. This insight is proven by the large complexity reduction possible through subgraph mining. This is an important element in building a small yet effective architecture search space.

\begin{figure*}
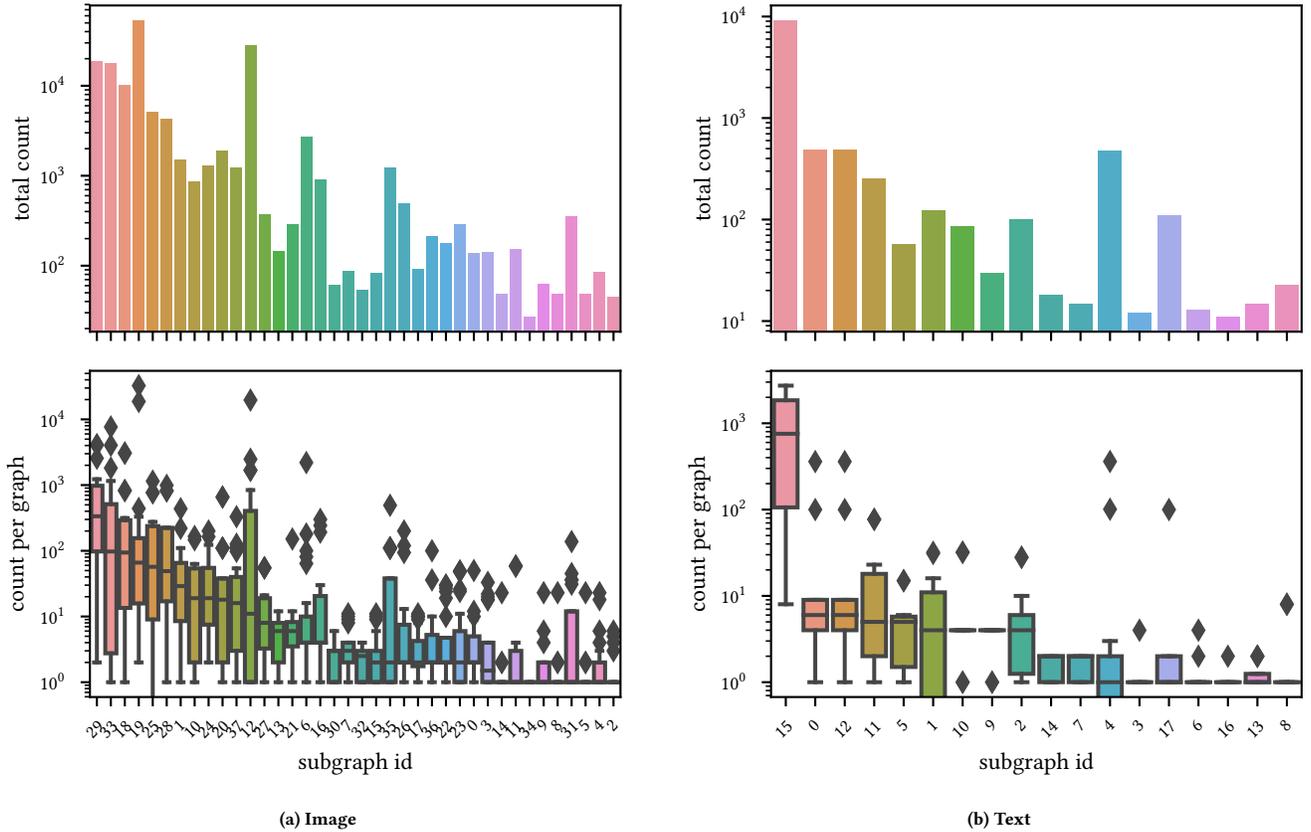

\begin{center}
\subfloat[Image]{
\input{figures/image/multiple_count_woutlier_log.pgf}
}
\subfloat[Text]{
\input{figures/nlp_translation_sentiment_ner/multiple_count_woutlier_log.pgf}
}
\caption{Count presence (multiple time) for each frequent subgraphs after reduction}
\label{fig_count_multiple}
\end{center}
\end{figure*}

In Figure \ref{fig_count_multiple} we show the fine-grained picture of individual subgraphs. The low number of subgraphs in the considered problems, after reduction, allows us to track each one individually. We opt for this visualization technique to emphasize the large number of occurrences some of these subgraphs have. 
The top figure indicates for each subgraph how many time they appears in total, sum of all the occurences of this frequent subgraphs.
The bottom figure shows for each subgraph the number of time it appears per graph. As we can see some frequent subgraphs appears multiple times in the same graphs with a high variability. Others, like subgraphs id 13 for \textit{image} or 14 for \textit{text}, appear the same number of time in each graph.

Some examples are telling of the reusability potential of these graphs as individualized modules. For the image-oriented task, subgraph 12 containing 6 nodes has 28149 occurrences in total and in the 26 distinct graphs in which this module appear the median number of occurence per graph is 11. 
For text, the subraph 15 with 7 nodes has a total number of occurences of 9183 and in the 9 distinct graphs in which this graph occurs the median number of occurence per graph is 756 . 
For reinforcement learning, there is a single subgraph. It appears exactly once in each of the 59 dinstict graphs it occurs in. However its size is so great (30 nodes) that it actually eclipses the changes that each graph adds to this common component. 
This naturally leads to the question of what gains are achievable if the subgraphs are treated as individual nodes. We address this in the following section.

Many subgraphs have occurrence outliers in Figure \ref{fig_count_multiple}. These portray the importance of using the found subgraphs as modules in the architecture evolution or search. If the cost of reusing the subgraph is low, it can be done a high number of times, thus mimicking the human creators of existing architectures.

Alternatively, from the lack of variability in other subgraph occurrences,
we discern the idea of layer and we get architectural insights. For instance, if a subgraph is a convolution, we can determine, from its occurrence count, the usual number of convolutions needed to solve the problem.
We only show plots for image and text, as for reinforcement there is a single subgraph, discussed previously and shown in \ref{fig-example-reinforcement}, that appears exactly once in each of the 59 dinstict graphs it occurs in.

\subsubsection {Complexity Reduction}

We inquire whether turning the found subgraphs into individual nodes helps in reducing the complexity of the neural search space. Currently, the GitGraph repository does not yet allow the creation of architectures using the frequent submodules as building blocks, which we leave for future work. We thus investigate the possible gains through a proxy - the ratio of the number of nodes that are in frequent subgraphs, with respect to the number of nodes found outside them. We use the term \textit{complexity reduction} to capture the ratio.

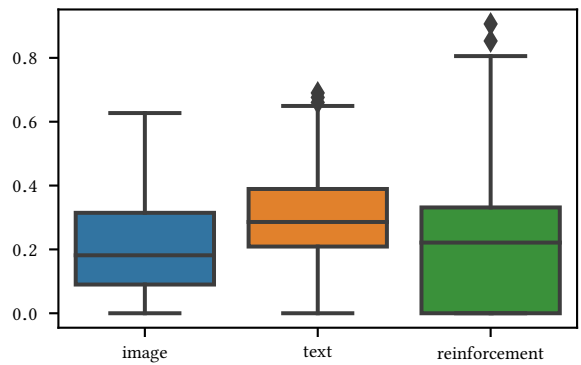
\begin{figure}
\input{figures/paper/stat_reduction_aggreg_all.pgf}
\caption{Distribution of node reduction on the whole data}
\label{fig_complex_reduc_all}
\end{figure}

To reduce a graph, we replace each of its frequent subgraphs by one node. We then compute the difference between the original graph node count and that of the reduced graph. Finally, we normalize by the original count to obtain a ratio of nodes that belong to the frequent components.
Figure \ref{fig_complex_reduc_all} plots the complexity reduction for the three studied tasks.
The median reduction for the three tasks ranges from 20 to 30\%, with the highest value being recorded for text.

This approach also allows us to identify the graphs where the reduction is 0\% or close to it. In some cases, these can contain new concepts. In others, this can be a filter to identify mistakes - for instance empty graphs or ones that have been included in the analyses because of a failure of the TF-IDF based search method.

\subsection {Generalization Capacity}

A relevant critique of the analysis in the section above is that it is descriptive and does not show the capacity of the subgraphs found to generalize. The fact that a subgraph is found in 20\% or more of the data in one dataset does not, in itself, guarantee that for a different set of graphs created for the same purpose will share these subgraphs.

\begin{figure*}
	\begin{center}
	\subfloat[Image]{
	\input{figures/image/train_test_stats.pgf}
	}
	\subfloat[Text]{
	\input{figures/nlp_translation_sentiment_ner/train_test_stats.pgf}
	}
	\subfloat[Reinforcement]{
	\input{figures/reinforcement/train_test_stats.pgf}
	}

	\caption{Distribution of node reduction on five sets (train/test split)}
	\label{fig_complex_reduc_train_test}
	\end{center}
\end{figure*}
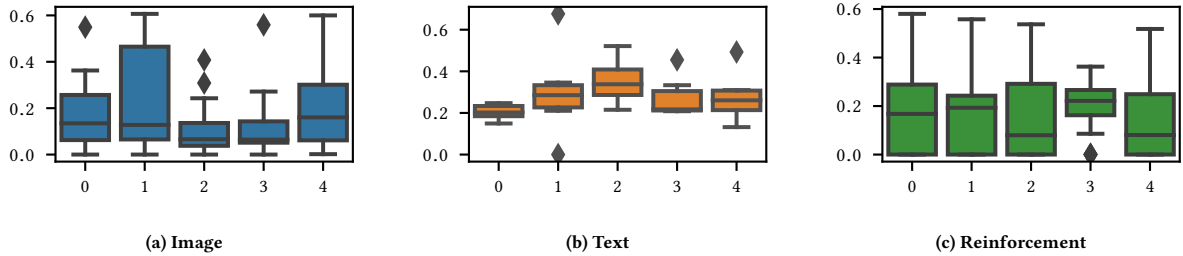

We thus create a new batch of tests to test the hypothesis that the subgraphs found are general and lined to the task itself. For each task, we create five different experiments in which we randomly split the data into a \"training\" set and a \"test\" set, with the goal of seeing whether the subgraphs mined from the graphs within the training set will occur in the graphs within the test set. In each experiment, we held 80\% of the graphs as training and the remaining 20\% for testing. We performed the frequent subgraph mining, followed by the addtional reduction only on the training dataset and we computed the node reduction using these subgraphs on the test dataset.
We report the results for each experiment individually in Figure \ref{fig_complex_reduc_train_test}.
The median and variance change very little from the previous experiment. This result upholds the one in the previous section and shows the complexity reduction is achievable on unseen graphs, for the same task.

%% file: figures/paper/sub_nodes_before_reduc_ccdf.pgf
\begingroup%
\makeatletter%
\begin{pgfpicture}%
\pgfpathrectangle{\pgfpointorigin}{\pgfqpoint{3.502802in}{2.164851in}}%
\pgfusepath{use as bounding box, clip}%
\begin{pgfscope}%
\pgfsetbuttcap%
\pgfsetmiterjoin%
\pgfsetlinewidth{0.000000pt}%
\definecolor{currentstroke}{rgb}{1.000000,1.000000,1.000000}%
\pgfsetstrokecolor{currentstroke}%
\pgfsetstrokeopacity{0.000000}%
\pgfsetdash{}{0pt}%
\pgfpathmoveto{\pgfqpoint{0.000000in}{0.000000in}}%
\pgfpathlineto{\pgfqpoint{3.502802in}{0.000000in}}%
\pgfpathlineto{\pgfqpoint{3.502802in}{2.164851in}}%
\pgfpathlineto{\pgfqpoint{0.000000in}{2.164851in}}%
\pgfpathclose%
\pgfusepath{}%
\end{pgfscope}%
\begin{pgfscope}%
\pgfsetbuttcap%
\pgfsetmiterjoin%
\definecolor{currentfill}{rgb}{1.000000,1.000000,1.000000}%
\pgfsetfillcolor{currentfill}%
\pgfsetlinewidth{0.000000pt}%
\definecolor{currentstroke}{rgb}{0.000000,0.000000,0.000000}%
\pgfsetstrokecolor{currentstroke}%
\pgfsetstrokeopacity{0.000000}%
\pgfsetdash{}{0pt}%
\pgfpathmoveto{\pgfqpoint{0.375397in}{0.484570in}}%
\pgfpathlineto{\pgfqpoint{3.332802in}{0.484570in}}%
\pgfpathlineto{\pgfqpoint{3.332802in}{1.994851in}}%
\pgfpathlineto{\pgfqpoint{0.375397in}{1.994851in}}%
\pgfpathclose%
\pgfusepath{fill}%
\end{pgfscope}%
\begin{pgfscope}%
\pgfpathrectangle{\pgfqpoint{0.375397in}{0.484570in}}{\pgfqpoint{2.957405in}{1.510280in}} %
\pgfusepath{clip}%
\pgfsetrectcap%
\pgfsetroundjoin%
\pgfsetlinewidth{0.803000pt}%
\definecolor{currentstroke}{rgb}{0.690196,0.690196,0.690196}%
\pgfsetstrokecolor{currentstroke}%
\pgfsetdash{}{0pt}%
\pgfpathmoveto{\pgfqpoint{0.509825in}{0.484570in}}%
\pgfpathlineto{\pgfqpoint{0.509825in}{1.994851in}}%
\pgfusepath{stroke}%
\end{pgfscope}%
\begin{pgfscope}%
\pgfsetbuttcap%
\pgfsetroundjoin%
\definecolor{currentfill}{rgb}{0.000000,0.000000,0.000000}%
\pgfsetfillcolor{currentfill}%
\pgfsetlinewidth{0.803000pt}%
\definecolor{currentstroke}{rgb}{0.000000,0.000000,0.000000}%
\pgfsetstrokecolor{currentstroke}%
\pgfsetdash{}{0pt}%
\pgfsys@defobject{currentmarker}{\pgfqpoint{0.000000in}{-0.048611in}}{\pgfqpoint{0.000000in}{0.000000in}}{%
\pgfpathmoveto{\pgfqpoint{0.000000in}{0.000000in}}%
\pgfpathlineto{\pgfqpoint{0.000000in}{-0.048611in}}%
\pgfusepath{stroke,fill}%
}%
\begin{pgfscope}%
\pgfsys@transformshift{0.509825in}{0.484570in}%
\pgfsys@useobject{currentmarker}{}%
\end{pgfscope}%
\end{pgfscope}%
\begin{pgfscope}%
\pgftext[x=0.509825in,y=0.387348in,,top]{\rmfamily\fontsize{7.000000}{8.400000}\selectfont \(\displaystyle 0\)}%
\end{pgfscope}%
\begin{pgfscope}%
\pgfpathrectangle{\pgfqpoint{0.375397in}{0.484570in}}{\pgfqpoint{2.957405in}{1.510280in}} %
\pgfusepath{clip}%
\pgfsetrectcap%
\pgfsetroundjoin%
\pgfsetlinewidth{0.803000pt}%
\definecolor{currentstroke}{rgb}{0.690196,0.690196,0.690196}%
\pgfsetstrokecolor{currentstroke}%
\pgfsetdash{}{0pt}%
\pgfpathmoveto{\pgfqpoint{0.957916in}{0.484570in}}%
\pgfpathlineto{\pgfqpoint{0.957916in}{1.994851in}}%
\pgfusepath{stroke}%
\end{pgfscope}%
\begin{pgfscope}%
\pgfsetbuttcap%
\pgfsetroundjoin%
\definecolor{currentfill}{rgb}{0.000000,0.000000,0.000000}%
\pgfsetfillcolor{currentfill}%
\pgfsetlinewidth{0.803000pt}%
\definecolor{currentstroke}{rgb}{0.000000,0.000000,0.000000}%
\pgfsetstrokecolor{currentstroke}%
\pgfsetdash{}{0pt}%
\pgfsys@defobject{currentmarker}{\pgfqpoint{0.000000in}{-0.048611in}}{\pgfqpoint{0.000000in}{0.000000in}}{%
\pgfpathmoveto{\pgfqpoint{0.000000in}{0.000000in}}%
\pgfpathlineto{\pgfqpoint{0.000000in}{-0.048611in}}%
\pgfusepath{stroke,fill}%
}%
\begin{pgfscope}%
\pgfsys@transformshift{0.957916in}{0.484570in}%
\pgfsys@useobject{currentmarker}{}%
\end{pgfscope}%
\end{pgfscope}%
\begin{pgfscope}%
\pgftext[x=0.957916in,y=0.387348in,,top]{\rmfamily\fontsize{7.000000}{8.400000}\selectfont \(\displaystyle 5\)}%
\end{pgfscope}%
\begin{pgfscope}%
\pgfpathrectangle{\pgfqpoint{0.375397in}{0.484570in}}{\pgfqpoint{2.957405in}{1.510280in}} %
\pgfusepath{clip}%
\pgfsetrectcap%
\pgfsetroundjoin%
\pgfsetlinewidth{0.803000pt}%
\definecolor{currentstroke}{rgb}{0.690196,0.690196,0.690196}%
\pgfsetstrokecolor{currentstroke}%
\pgfsetdash{}{0pt}%
\pgfpathmoveto{\pgfqpoint{1.406008in}{0.484570in}}%
\pgfpathlineto{\pgfqpoint{1.406008in}{1.994851in}}%
\pgfusepath{stroke}%
\end{pgfscope}%
\begin{pgfscope}%
\pgfsetbuttcap%
\pgfsetroundjoin%
\definecolor{currentfill}{rgb}{0.000000,0.000000,0.000000}%
\pgfsetfillcolor{currentfill}%
\pgfsetlinewidth{0.803000pt}%
\definecolor{currentstroke}{rgb}{0.000000,0.000000,0.000000}%
\pgfsetstrokecolor{currentstroke}%
\pgfsetdash{}{0pt}%
\pgfsys@defobject{currentmarker}{\pgfqpoint{0.000000in}{-0.048611in}}{\pgfqpoint{0.000000in}{0.000000in}}{%
\pgfpathmoveto{\pgfqpoint{0.000000in}{0.000000in}}%
\pgfpathlineto{\pgfqpoint{0.000000in}{-0.048611in}}%
\pgfusepath{stroke,fill}%
}%
\begin{pgfscope}%
\pgfsys@transformshift{1.406008in}{0.484570in}%
\pgfsys@useobject{currentmarker}{}%
\end{pgfscope}%
\end{pgfscope}%
\begin{pgfscope}%
\pgftext[x=1.406008in,y=0.387348in,,top]{\rmfamily\fontsize{7.000000}{8.400000}\selectfont \(\displaystyle 10\)}%
\end{pgfscope}%
\begin{pgfscope}%
\pgfpathrectangle{\pgfqpoint{0.375397in}{0.484570in}}{\pgfqpoint{2.957405in}{1.510280in}} %
\pgfusepath{clip}%
\pgfsetrectcap%
\pgfsetroundjoin%
\pgfsetlinewidth{0.803000pt}%
\definecolor{currentstroke}{rgb}{0.690196,0.690196,0.690196}%
\pgfsetstrokecolor{currentstroke}%
\pgfsetdash{}{0pt}%
\pgfpathmoveto{\pgfqpoint{1.854100in}{0.484570in}}%
\pgfpathlineto{\pgfqpoint{1.854100in}{1.994851in}}%
\pgfusepath{stroke}%
\end{pgfscope}%
\begin{pgfscope}%
\pgfsetbuttcap%
\pgfsetroundjoin%
\definecolor{currentfill}{rgb}{0.000000,0.000000,0.000000}%
\pgfsetfillcolor{currentfill}%
\pgfsetlinewidth{0.803000pt}%
\definecolor{currentstroke}{rgb}{0.000000,0.000000,0.000000}%
\pgfsetstrokecolor{currentstroke}%
\pgfsetdash{}{0pt}%
\pgfsys@defobject{currentmarker}{\pgfqpoint{0.000000in}{-0.048611in}}{\pgfqpoint{0.000000in}{0.000000in}}{%
\pgfpathmoveto{\pgfqpoint{0.000000in}{0.000000in}}%
\pgfpathlineto{\pgfqpoint{0.000000in}{-0.048611in}}%
\pgfusepath{stroke,fill}%
}%
\begin{pgfscope}%
\pgfsys@transformshift{1.854100in}{0.484570in}%
\pgfsys@useobject{currentmarker}{}%
\end{pgfscope}%
\end{pgfscope}%
\begin{pgfscope}%
\pgftext[x=1.854100in,y=0.387348in,,top]{\rmfamily\fontsize{7.000000}{8.400000}\selectfont \(\displaystyle 15\)}%
\end{pgfscope}%
\begin{pgfscope}%
\pgfpathrectangle{\pgfqpoint{0.375397in}{0.484570in}}{\pgfqpoint{2.957405in}{1.510280in}} %
\pgfusepath{clip}%
\pgfsetrectcap%
\pgfsetroundjoin%
\pgfsetlinewidth{0.803000pt}%
\definecolor{currentstroke}{rgb}{0.690196,0.690196,0.690196}%
\pgfsetstrokecolor{currentstroke}%
\pgfsetdash{}{0pt}%
\pgfpathmoveto{\pgfqpoint{2.302191in}{0.484570in}}%
\pgfpathlineto{\pgfqpoint{2.302191in}{1.994851in}}%
\pgfusepath{stroke}%
\end{pgfscope}%
\begin{pgfscope}%
\pgfsetbuttcap%
\pgfsetroundjoin%
\definecolor{currentfill}{rgb}{0.000000,0.000000,0.000000}%
\pgfsetfillcolor{currentfill}%
\pgfsetlinewidth{0.803000pt}%
\definecolor{currentstroke}{rgb}{0.000000,0.000000,0.000000}%
\pgfsetstrokecolor{currentstroke}%
\pgfsetdash{}{0pt}%
\pgfsys@defobject{currentmarker}{\pgfqpoint{0.000000in}{-0.048611in}}{\pgfqpoint{0.000000in}{0.000000in}}{%
\pgfpathmoveto{\pgfqpoint{0.000000in}{0.000000in}}%
\pgfpathlineto{\pgfqpoint{0.000000in}{-0.048611in}}%
\pgfusepath{stroke,fill}%
}%
\begin{pgfscope}%
\pgfsys@transformshift{2.302191in}{0.484570in}%
\pgfsys@useobject{currentmarker}{}%
\end{pgfscope}%
\end{pgfscope}%
\begin{pgfscope}%
\pgftext[x=2.302191in,y=0.387348in,,top]{\rmfamily\fontsize{7.000000}{8.400000}\selectfont \(\displaystyle 20\)}%
\end{pgfscope}%
\begin{pgfscope}%
\pgfpathrectangle{\pgfqpoint{0.375397in}{0.484570in}}{\pgfqpoint{2.957405in}{1.510280in}} %
\pgfusepath{clip}%
\pgfsetrectcap%
\pgfsetroundjoin%
\pgfsetlinewidth{0.803000pt}%
\definecolor{currentstroke}{rgb}{0.690196,0.690196,0.690196}%
\pgfsetstrokecolor{currentstroke}%
\pgfsetdash{}{0pt}%
\pgfpathmoveto{\pgfqpoint{2.750283in}{0.484570in}}%
\pgfpathlineto{\pgfqpoint{2.750283in}{1.994851in}}%
\pgfusepath{stroke}%
\end{pgfscope}%
\begin{pgfscope}%
\pgfsetbuttcap%
\pgfsetroundjoin%
\definecolor{currentfill}{rgb}{0.000000,0.000000,0.000000}%
\pgfsetfillcolor{currentfill}%
\pgfsetlinewidth{0.803000pt}%
\definecolor{currentstroke}{rgb}{0.000000,0.000000,0.000000}%
\pgfsetstrokecolor{currentstroke}%
\pgfsetdash{}{0pt}%
\pgfsys@defobject{currentmarker}{\pgfqpoint{0.000000in}{-0.048611in}}{\pgfqpoint{0.000000in}{0.000000in}}{%
\pgfpathmoveto{\pgfqpoint{0.000000in}{0.000000in}}%
\pgfpathlineto{\pgfqpoint{0.000000in}{-0.048611in}}%
\pgfusepath{stroke,fill}%
}%
\begin{pgfscope}%
\pgfsys@transformshift{2.750283in}{0.484570in}%
\pgfsys@useobject{currentmarker}{}%
\end{pgfscope}%
\end{pgfscope}%
\begin{pgfscope}%
\pgftext[x=2.750283in,y=0.387348in,,top]{\rmfamily\fontsize{7.000000}{8.400000}\selectfont \(\displaystyle 25\)}%
\end{pgfscope}%
\begin{pgfscope}%
\pgfpathrectangle{\pgfqpoint{0.375397in}{0.484570in}}{\pgfqpoint{2.957405in}{1.510280in}} %
\pgfusepath{clip}%
\pgfsetrectcap%
\pgfsetroundjoin%
\pgfsetlinewidth{0.803000pt}%
\definecolor{currentstroke}{rgb}{0.690196,0.690196,0.690196}%
\pgfsetstrokecolor{currentstroke}%
\pgfsetdash{}{0pt}%
\pgfpathmoveto{\pgfqpoint{3.198375in}{0.484570in}}%
\pgfpathlineto{\pgfqpoint{3.198375in}{1.994851in}}%
\pgfusepath{stroke}%
\end{pgfscope}%
\begin{pgfscope}%
\pgfsetbuttcap%
\pgfsetroundjoin%
\definecolor{currentfill}{rgb}{0.000000,0.000000,0.000000}%
\pgfsetfillcolor{currentfill}%
\pgfsetlinewidth{0.803000pt}%
\definecolor{currentstroke}{rgb}{0.000000,0.000000,0.000000}%
\pgfsetstrokecolor{currentstroke}%
\pgfsetdash{}{0pt}%
\pgfsys@defobject{currentmarker}{\pgfqpoint{0.000000in}{-0.048611in}}{\pgfqpoint{0.000000in}{0.000000in}}{%
\pgfpathmoveto{\pgfqpoint{0.000000in}{0.000000in}}%
\pgfpathlineto{\pgfqpoint{0.000000in}{-0.048611in}}%
\pgfusepath{stroke,fill}%
}%
\begin{pgfscope}%
\pgfsys@transformshift{3.198375in}{0.484570in}%
\pgfsys@useobject{currentmarker}{}%
\end{pgfscope}%
\end{pgfscope}%
\begin{pgfscope}%
\pgftext[x=3.198375in,y=0.387348in,,top]{\rmfamily\fontsize{7.000000}{8.400000}\selectfont \(\displaystyle 30\)}%
\end{pgfscope}%
\begin{pgfscope}%
\pgftext[x=1.854100in,y=0.245934in,,top]{\rmfamily\fontsize{9.000000}{10.800000}\selectfont number of nodes}%
\end{pgfscope}%
\begin{pgfscope}%
\pgfpathrectangle{\pgfqpoint{0.375397in}{0.484570in}}{\pgfqpoint{2.957405in}{1.510280in}} %
\pgfusepath{clip}%
\pgfsetrectcap%
\pgfsetroundjoin%
\pgfsetlinewidth{0.803000pt}%
\definecolor{currentstroke}{rgb}{0.690196,0.690196,0.690196}%
\pgfsetstrokecolor{currentstroke}%
\pgfsetdash{}{0pt}%
\pgfpathmoveto{\pgfqpoint{0.375397in}{0.553219in}}%
\pgfpathlineto{\pgfqpoint{3.332802in}{0.553219in}}%
\pgfusepath{stroke}%
\end{pgfscope}%
\begin{pgfscope}%
\pgfsetbuttcap%
\pgfsetroundjoin%
\definecolor{currentfill}{rgb}{0.000000,0.000000,0.000000}%
\pgfsetfillcolor{currentfill}%
\pgfsetlinewidth{0.803000pt}%
\definecolor{currentstroke}{rgb}{0.000000,0.000000,0.000000}%
\pgfsetstrokecolor{currentstroke}%
\pgfsetdash{}{0pt}%
\pgfsys@defobject{currentmarker}{\pgfqpoint{-0.048611in}{0.000000in}}{\pgfqpoint{0.000000in}{0.000000in}}{%
\pgfpathmoveto{\pgfqpoint{0.000000in}{0.000000in}}%
\pgfpathlineto{\pgfqpoint{-0.048611in}{0.000000in}}%
\pgfusepath{stroke,fill}%
}%
\begin{pgfscope}%
\pgfsys@transformshift{0.375397in}{0.553219in}%
\pgfsys@useobject{currentmarker}{}%
\end{pgfscope}%
\end{pgfscope}%
\begin{pgfscope}%
\pgftext[x=0.134463in,y=0.519740in,left,base]{\rmfamily\fontsize{7.000000}{8.400000}\selectfont \(\displaystyle 0.0\)}%
\end{pgfscope}%
\begin{pgfscope}%
\pgfpathrectangle{\pgfqpoint{0.375397in}{0.484570in}}{\pgfqpoint{2.957405in}{1.510280in}} %
\pgfusepath{clip}%
\pgfsetrectcap%
\pgfsetroundjoin%
\pgfsetlinewidth{0.803000pt}%
\definecolor{currentstroke}{rgb}{0.690196,0.690196,0.690196}%
\pgfsetstrokecolor{currentstroke}%
\pgfsetdash{}{0pt}%
\pgfpathmoveto{\pgfqpoint{0.375397in}{0.827816in}}%
\pgfpathlineto{\pgfqpoint{3.332802in}{0.827816in}}%
\pgfusepath{stroke}%
\end{pgfscope}%
\begin{pgfscope}%
\pgfsetbuttcap%
\pgfsetroundjoin%
\definecolor{currentfill}{rgb}{0.000000,0.000000,0.000000}%
\pgfsetfillcolor{currentfill}%
\pgfsetlinewidth{0.803000pt}%
\definecolor{currentstroke}{rgb}{0.000000,0.000000,0.000000}%
\pgfsetstrokecolor{currentstroke}%
\pgfsetdash{}{0pt}%
\pgfsys@defobject{currentmarker}{\pgfqpoint{-0.048611in}{0.000000in}}{\pgfqpoint{0.000000in}{0.000000in}}{%
\pgfpathmoveto{\pgfqpoint{0.000000in}{0.000000in}}%
\pgfpathlineto{\pgfqpoint{-0.048611in}{0.000000in}}%
\pgfusepath{stroke,fill}%
}%
\begin{pgfscope}%
\pgfsys@transformshift{0.375397in}{0.827816in}%
\pgfsys@useobject{currentmarker}{}%
\end{pgfscope}%
\end{pgfscope}%
\begin{pgfscope}%
\pgftext[x=0.134463in,y=0.794336in,left,base]{\rmfamily\fontsize{7.000000}{8.400000}\selectfont \(\displaystyle 0.2\)}%
\end{pgfscope}%
\begin{pgfscope}%
\pgfpathrectangle{\pgfqpoint{0.375397in}{0.484570in}}{\pgfqpoint{2.957405in}{1.510280in}} %
\pgfusepath{clip}%
\pgfsetrectcap%
\pgfsetroundjoin%
\pgfsetlinewidth{0.803000pt}%
\definecolor{currentstroke}{rgb}{0.690196,0.690196,0.690196}%
\pgfsetstrokecolor{currentstroke}%
\pgfsetdash{}{0pt}%
\pgfpathmoveto{\pgfqpoint{0.375397in}{1.102412in}}%
\pgfpathlineto{\pgfqpoint{3.332802in}{1.102412in}}%
\pgfusepath{stroke}%
\end{pgfscope}%
\begin{pgfscope}%
\pgfsetbuttcap%
\pgfsetroundjoin%
\definecolor{currentfill}{rgb}{0.000000,0.000000,0.000000}%
\pgfsetfillcolor{currentfill}%
\pgfsetlinewidth{0.803000pt}%
\definecolor{currentstroke}{rgb}{0.000000,0.000000,0.000000}%
\pgfsetstrokecolor{currentstroke}%
\pgfsetdash{}{0pt}%
\pgfsys@defobject{currentmarker}{\pgfqpoint{-0.048611in}{0.000000in}}{\pgfqpoint{0.000000in}{0.000000in}}{%
\pgfpathmoveto{\pgfqpoint{0.000000in}{0.000000in}}%
\pgfpathlineto{\pgfqpoint{-0.048611in}{0.000000in}}%
\pgfusepath{stroke,fill}%
}%
\begin{pgfscope}%
\pgfsys@transformshift{0.375397in}{1.102412in}%
\pgfsys@useobject{currentmarker}{}%
\end{pgfscope}%
\end{pgfscope}%
\begin{pgfscope}%
\pgftext[x=0.134463in,y=1.068933in,left,base]{\rmfamily\fontsize{7.000000}{8.400000}\selectfont \(\displaystyle 0.4\)}%
\end{pgfscope}%
\begin{pgfscope}%
\pgfpathrectangle{\pgfqpoint{0.375397in}{0.484570in}}{\pgfqpoint{2.957405in}{1.510280in}} %
\pgfusepath{clip}%
\pgfsetrectcap%
\pgfsetroundjoin%
\pgfsetlinewidth{0.803000pt}%
\definecolor{currentstroke}{rgb}{0.690196,0.690196,0.690196}%
\pgfsetstrokecolor{currentstroke}%
\pgfsetdash{}{0pt}%
\pgfpathmoveto{\pgfqpoint{0.375397in}{1.377009in}}%
\pgfpathlineto{\pgfqpoint{3.332802in}{1.377009in}}%
\pgfusepath{stroke}%
\end{pgfscope}%
\begin{pgfscope}%
\pgfsetbuttcap%
\pgfsetroundjoin%
\definecolor{currentfill}{rgb}{0.000000,0.000000,0.000000}%
\pgfsetfillcolor{currentfill}%
\pgfsetlinewidth{0.803000pt}%
\definecolor{currentstroke}{rgb}{0.000000,0.000000,0.000000}%
\pgfsetstrokecolor{currentstroke}%
\pgfsetdash{}{0pt}%
\pgfsys@defobject{currentmarker}{\pgfqpoint{-0.048611in}{0.000000in}}{\pgfqpoint{0.000000in}{0.000000in}}{%
\pgfpathmoveto{\pgfqpoint{0.000000in}{0.000000in}}%
\pgfpathlineto{\pgfqpoint{-0.048611in}{0.000000in}}%
\pgfusepath{stroke,fill}%
}%
\begin{pgfscope}%
\pgfsys@transformshift{0.375397in}{1.377009in}%
\pgfsys@useobject{currentmarker}{}%
\end{pgfscope}%
\end{pgfscope}%
\begin{pgfscope}%
\pgftext[x=0.134463in,y=1.343529in,left,base]{\rmfamily\fontsize{7.000000}{8.400000}\selectfont \(\displaystyle 0.6\)}%
\end{pgfscope}%
\begin{pgfscope}%
\pgfpathrectangle{\pgfqpoint{0.375397in}{0.484570in}}{\pgfqpoint{2.957405in}{1.510280in}} %
\pgfusepath{clip}%
\pgfsetrectcap%
\pgfsetroundjoin%
\pgfsetlinewidth{0.803000pt}%
\definecolor{currentstroke}{rgb}{0.690196,0.690196,0.690196}%
\pgfsetstrokecolor{currentstroke}%
\pgfsetdash{}{0pt}%
\pgfpathmoveto{\pgfqpoint{0.375397in}{1.651605in}}%
\pgfpathlineto{\pgfqpoint{3.332802in}{1.651605in}}%
\pgfusepath{stroke}%
\end{pgfscope}%
\begin{pgfscope}%
\pgfsetbuttcap%
\pgfsetroundjoin%
\definecolor{currentfill}{rgb}{0.000000,0.000000,0.000000}%
\pgfsetfillcolor{currentfill}%
\pgfsetlinewidth{0.803000pt}%
\definecolor{currentstroke}{rgb}{0.000000,0.000000,0.000000}%
\pgfsetstrokecolor{currentstroke}%
\pgfsetdash{}{0pt}%
\pgfsys@defobject{currentmarker}{\pgfqpoint{-0.048611in}{0.000000in}}{\pgfqpoint{0.000000in}{0.000000in}}{%
\pgfpathmoveto{\pgfqpoint{0.000000in}{0.000000in}}%
\pgfpathlineto{\pgfqpoint{-0.048611in}{0.000000in}}%
\pgfusepath{stroke,fill}%
}%
\begin{pgfscope}%
\pgfsys@transformshift{0.375397in}{1.651605in}%
\pgfsys@useobject{currentmarker}{}%
\end{pgfscope}%
\end{pgfscope}%
\begin{pgfscope}%
\pgftext[x=0.134463in,y=1.618126in,left,base]{\rmfamily\fontsize{7.000000}{8.400000}\selectfont \(\displaystyle 0.8\)}%
\end{pgfscope}%
\begin{pgfscope}%
\pgfpathrectangle{\pgfqpoint{0.375397in}{0.484570in}}{\pgfqpoint{2.957405in}{1.510280in}} %
\pgfusepath{clip}%
\pgfsetrectcap%
\pgfsetroundjoin%
\pgfsetlinewidth{0.803000pt}%
\definecolor{currentstroke}{rgb}{0.690196,0.690196,0.690196}%
\pgfsetstrokecolor{currentstroke}%
\pgfsetdash{}{0pt}%
\pgfpathmoveto{\pgfqpoint{0.375397in}{1.926202in}}%
\pgfpathlineto{\pgfqpoint{3.332802in}{1.926202in}}%
\pgfusepath{stroke}%
\end{pgfscope}%
\begin{pgfscope}%
\pgfsetbuttcap%
\pgfsetroundjoin%
\definecolor{currentfill}{rgb}{0.000000,0.000000,0.000000}%
\pgfsetfillcolor{currentfill}%
\pgfsetlinewidth{0.803000pt}%
\definecolor{currentstroke}{rgb}{0.000000,0.000000,0.000000}%
\pgfsetstrokecolor{currentstroke}%
\pgfsetdash{}{0pt}%
\pgfsys@defobject{currentmarker}{\pgfqpoint{-0.048611in}{0.000000in}}{\pgfqpoint{0.000000in}{0.000000in}}{%
\pgfpathmoveto{\pgfqpoint{0.000000in}{0.000000in}}%
\pgfpathlineto{\pgfqpoint{-0.048611in}{0.000000in}}%
\pgfusepath{stroke,fill}%
}%
\begin{pgfscope}%
\pgfsys@transformshift{0.375397in}{1.926202in}%
\pgfsys@useobject{currentmarker}{}%
\end{pgfscope}%
\end{pgfscope}%
\begin{pgfscope}%
\pgftext[x=0.134463in,y=1.892722in,left,base]{\rmfamily\fontsize{7.000000}{8.400000}\selectfont \(\displaystyle 1.0\)}%
\end{pgfscope}%
\begin{pgfscope}%
\pgfpathrectangle{\pgfqpoint{0.375397in}{0.484570in}}{\pgfqpoint{2.957405in}{1.510280in}} %
\pgfusepath{clip}%
\pgfsetrectcap%
\pgfsetroundjoin%
\pgfsetlinewidth{1.505625pt}%
\definecolor{currentstroke}{rgb}{0.121569,0.466667,0.705882}%
\pgfsetstrokecolor{currentstroke}%
\pgfsetdash{}{0pt}%
\pgfpathmoveto{\pgfqpoint{0.509825in}{1.926202in}}%
\pgfpathlineto{\pgfqpoint{0.599443in}{1.926202in}}%
\pgfpathlineto{\pgfqpoint{0.599443in}{1.926202in}}%
\pgfpathlineto{\pgfqpoint{0.689061in}{1.926202in}}%
\pgfpathlineto{\pgfqpoint{0.689061in}{1.926202in}}%
\pgfpathlineto{\pgfqpoint{0.778680in}{1.926202in}}%
\pgfpathlineto{\pgfqpoint{0.778680in}{1.479703in}}%
\pgfpathlineto{\pgfqpoint{0.868298in}{1.479703in}}%
\pgfpathlineto{\pgfqpoint{0.868298in}{1.267617in}}%
\pgfpathlineto{\pgfqpoint{0.957916in}{1.267617in}}%
\pgfpathlineto{\pgfqpoint{0.957916in}{1.100180in}}%
\pgfpathlineto{\pgfqpoint{1.047535in}{1.100180in}}%
\pgfpathlineto{\pgfqpoint{1.047535in}{0.977393in}}%
\pgfpathlineto{\pgfqpoint{1.137153in}{0.977393in}}%
\pgfpathlineto{\pgfqpoint{1.137153in}{0.843443in}}%
\pgfpathlineto{\pgfqpoint{1.226771in}{0.843443in}}%
\pgfpathlineto{\pgfqpoint{1.226771in}{0.776469in}}%
\pgfpathlineto{\pgfqpoint{1.316390in}{0.776469in}}%
\pgfpathlineto{\pgfqpoint{1.316390in}{0.742981in}}%
\pgfpathlineto{\pgfqpoint{1.406008in}{0.742981in}}%
\pgfpathlineto{\pgfqpoint{1.406008in}{0.709494in}}%
\pgfpathlineto{\pgfqpoint{1.495626in}{0.709494in}}%
\pgfpathlineto{\pgfqpoint{1.495626in}{0.653682in}}%
\pgfpathlineto{\pgfqpoint{1.585245in}{0.653682in}}%
\pgfpathlineto{\pgfqpoint{1.585245in}{0.609032in}}%
\pgfpathlineto{\pgfqpoint{1.674863in}{0.609032in}}%
\pgfpathlineto{\pgfqpoint{1.674863in}{0.575544in}}%
\pgfpathlineto{\pgfqpoint{1.764481in}{0.575544in}}%
\pgfpathlineto{\pgfqpoint{1.764481in}{0.564382in}}%
\pgfpathlineto{\pgfqpoint{1.854100in}{0.564382in}}%
\pgfpathlineto{\pgfqpoint{1.854100in}{0.553219in}}%
\pgfusepath{stroke}%
\end{pgfscope}%
\begin{pgfscope}%
\pgfpathrectangle{\pgfqpoint{0.375397in}{0.484570in}}{\pgfqpoint{2.957405in}{1.510280in}} %
\pgfusepath{clip}%
\pgfsetrectcap%
\pgfsetroundjoin%
\pgfsetlinewidth{1.505625pt}%
\definecolor{currentstroke}{rgb}{1.000000,0.498039,0.054902}%
\pgfsetstrokecolor{currentstroke}%
\pgfsetdash{}{0pt}%
\pgfpathmoveto{\pgfqpoint{0.509825in}{1.926202in}}%
\pgfpathlineto{\pgfqpoint{0.599443in}{1.926202in}}%
\pgfpathlineto{\pgfqpoint{0.599443in}{1.926202in}}%
\pgfpathlineto{\pgfqpoint{0.689061in}{1.926202in}}%
\pgfpathlineto{\pgfqpoint{0.689061in}{1.926202in}}%
\pgfpathlineto{\pgfqpoint{0.778680in}{1.926202in}}%
\pgfpathlineto{\pgfqpoint{0.778680in}{1.508337in}}%
\pgfpathlineto{\pgfqpoint{0.868298in}{1.508337in}}%
\pgfpathlineto{\pgfqpoint{0.868298in}{1.150168in}}%
\pgfpathlineto{\pgfqpoint{0.957916in}{1.150168in}}%
\pgfpathlineto{\pgfqpoint{0.957916in}{1.000931in}}%
\pgfpathlineto{\pgfqpoint{1.047535in}{1.000931in}}%
\pgfpathlineto{\pgfqpoint{1.047535in}{0.941236in}}%
\pgfpathlineto{\pgfqpoint{1.137153in}{0.941236in}}%
\pgfpathlineto{\pgfqpoint{1.137153in}{0.881541in}}%
\pgfpathlineto{\pgfqpoint{1.226771in}{0.881541in}}%
\pgfpathlineto{\pgfqpoint{1.226771in}{0.881541in}}%
\pgfpathlineto{\pgfqpoint{1.316390in}{0.881541in}}%
\pgfpathlineto{\pgfqpoint{1.316390in}{0.821846in}}%
\pgfpathlineto{\pgfqpoint{1.406008in}{0.821846in}}%
\pgfpathlineto{\pgfqpoint{1.406008in}{0.732304in}}%
\pgfpathlineto{\pgfqpoint{1.495626in}{0.732304in}}%
\pgfpathlineto{\pgfqpoint{1.495626in}{0.702457in}}%
\pgfpathlineto{\pgfqpoint{1.585245in}{0.702457in}}%
\pgfpathlineto{\pgfqpoint{1.585245in}{0.672609in}}%
\pgfpathlineto{\pgfqpoint{1.674863in}{0.672609in}}%
\pgfpathlineto{\pgfqpoint{1.674863in}{0.672609in}}%
\pgfpathlineto{\pgfqpoint{1.764481in}{0.672609in}}%
\pgfpathlineto{\pgfqpoint{1.764481in}{0.672609in}}%
\pgfpathlineto{\pgfqpoint{1.854100in}{0.672609in}}%
\pgfpathlineto{\pgfqpoint{1.854100in}{0.672609in}}%
\pgfpathlineto{\pgfqpoint{1.943718in}{0.672609in}}%
\pgfpathlineto{\pgfqpoint{1.943718in}{0.672609in}}%
\pgfpathlineto{\pgfqpoint{2.033336in}{0.672609in}}%
\pgfpathlineto{\pgfqpoint{2.033336in}{0.672609in}}%
\pgfpathlineto{\pgfqpoint{2.122955in}{0.672609in}}%
\pgfpathlineto{\pgfqpoint{2.122955in}{0.672609in}}%
\pgfpathlineto{\pgfqpoint{2.212573in}{0.672609in}}%
\pgfpathlineto{\pgfqpoint{2.212573in}{0.672609in}}%
\pgfpathlineto{\pgfqpoint{2.302191in}{0.672609in}}%
\pgfpathlineto{\pgfqpoint{2.302191in}{0.672609in}}%
\pgfpathlineto{\pgfqpoint{2.391810in}{0.672609in}}%
\pgfpathlineto{\pgfqpoint{2.391810in}{0.672609in}}%
\pgfpathlineto{\pgfqpoint{2.481428in}{0.672609in}}%
\pgfpathlineto{\pgfqpoint{2.481428in}{0.672609in}}%
\pgfpathlineto{\pgfqpoint{2.571046in}{0.672609in}}%
\pgfpathlineto{\pgfqpoint{2.571046in}{0.672609in}}%
\pgfpathlineto{\pgfqpoint{2.660665in}{0.672609in}}%
\pgfpathlineto{\pgfqpoint{2.660665in}{0.672609in}}%
\pgfpathlineto{\pgfqpoint{2.750283in}{0.672609in}}%
\pgfpathlineto{\pgfqpoint{2.750283in}{0.672609in}}%
\pgfpathlineto{\pgfqpoint{2.839901in}{0.672609in}}%
\pgfpathlineto{\pgfqpoint{2.839901in}{0.672609in}}%
\pgfpathlineto{\pgfqpoint{2.929520in}{0.672609in}}%
\pgfpathlineto{\pgfqpoint{2.929520in}{0.642762in}}%
\pgfpathlineto{\pgfqpoint{3.019138in}{0.642762in}}%
\pgfpathlineto{\pgfqpoint{3.019138in}{0.612914in}}%
\pgfpathlineto{\pgfqpoint{3.108756in}{0.612914in}}%
\pgfpathlineto{\pgfqpoint{3.108756in}{0.583067in}}%
\pgfpathlineto{\pgfqpoint{3.198375in}{0.583067in}}%
\pgfpathlineto{\pgfqpoint{3.198375in}{0.553219in}}%
\pgfusepath{stroke}%
\end{pgfscope}%
\begin{pgfscope}%
\pgfpathrectangle{\pgfqpoint{0.375397in}{0.484570in}}{\pgfqpoint{2.957405in}{1.510280in}} %
\pgfusepath{clip}%
\pgfsetrectcap%
\pgfsetroundjoin%
\pgfsetlinewidth{1.505625pt}%
\definecolor{currentstroke}{rgb}{0.172549,0.627451,0.172549}%
\pgfsetstrokecolor{currentstroke}%
\pgfsetdash{}{0pt}%
\pgfpathmoveto{\pgfqpoint{0.509825in}{1.926202in}}%
\pgfpathlineto{\pgfqpoint{0.599443in}{1.926202in}}%
\pgfpathlineto{\pgfqpoint{0.599443in}{1.926202in}}%
\pgfpathlineto{\pgfqpoint{0.689061in}{1.926202in}}%
\pgfpathlineto{\pgfqpoint{0.689061in}{1.926202in}}%
\pgfpathlineto{\pgfqpoint{0.778680in}{1.926202in}}%
\pgfpathlineto{\pgfqpoint{0.778680in}{1.508337in}}%
\pgfpathlineto{\pgfqpoint{0.868298in}{1.508337in}}%
\pgfpathlineto{\pgfqpoint{0.868298in}{1.388948in}}%
\pgfpathlineto{\pgfqpoint{0.957916in}{1.388948in}}%
\pgfpathlineto{\pgfqpoint{0.957916in}{1.150168in}}%
\pgfpathlineto{\pgfqpoint{1.047535in}{1.150168in}}%
\pgfpathlineto{\pgfqpoint{1.047535in}{1.090473in}}%
\pgfpathlineto{\pgfqpoint{1.137153in}{1.090473in}}%
\pgfpathlineto{\pgfqpoint{1.137153in}{1.090473in}}%
\pgfpathlineto{\pgfqpoint{1.226771in}{1.090473in}}%
\pgfpathlineto{\pgfqpoint{1.226771in}{1.030778in}}%
\pgfpathlineto{\pgfqpoint{1.316390in}{1.030778in}}%
\pgfpathlineto{\pgfqpoint{1.316390in}{1.030778in}}%
\pgfpathlineto{\pgfqpoint{1.406008in}{1.030778in}}%
\pgfpathlineto{\pgfqpoint{1.406008in}{1.030778in}}%
\pgfpathlineto{\pgfqpoint{1.495626in}{1.030778in}}%
\pgfpathlineto{\pgfqpoint{1.495626in}{1.030778in}}%
\pgfpathlineto{\pgfqpoint{1.585245in}{1.030778in}}%
\pgfpathlineto{\pgfqpoint{1.585245in}{1.030778in}}%
\pgfpathlineto{\pgfqpoint{1.674863in}{1.030778in}}%
\pgfpathlineto{\pgfqpoint{1.674863in}{1.030778in}}%
\pgfpathlineto{\pgfqpoint{1.764481in}{1.030778in}}%
\pgfpathlineto{\pgfqpoint{1.764481in}{1.030778in}}%
\pgfpathlineto{\pgfqpoint{1.854100in}{1.030778in}}%
\pgfpathlineto{\pgfqpoint{1.854100in}{1.030778in}}%
\pgfpathlineto{\pgfqpoint{1.943718in}{1.030778in}}%
\pgfpathlineto{\pgfqpoint{1.943718in}{0.971084in}}%
\pgfpathlineto{\pgfqpoint{2.033336in}{0.971084in}}%
\pgfpathlineto{\pgfqpoint{2.033336in}{0.911389in}}%
\pgfpathlineto{\pgfqpoint{2.122955in}{0.911389in}}%
\pgfpathlineto{\pgfqpoint{2.122955in}{0.791999in}}%
\pgfpathlineto{\pgfqpoint{2.212573in}{0.791999in}}%
\pgfpathlineto{\pgfqpoint{2.212573in}{0.791999in}}%
\pgfpathlineto{\pgfqpoint{2.302191in}{0.791999in}}%
\pgfpathlineto{\pgfqpoint{2.302191in}{0.732304in}}%
\pgfpathlineto{\pgfqpoint{2.391810in}{0.732304in}}%
\pgfpathlineto{\pgfqpoint{2.391810in}{0.672609in}}%
\pgfpathlineto{\pgfqpoint{2.481428in}{0.672609in}}%
\pgfpathlineto{\pgfqpoint{2.481428in}{0.672609in}}%
\pgfpathlineto{\pgfqpoint{2.571046in}{0.672609in}}%
\pgfpathlineto{\pgfqpoint{2.571046in}{0.612914in}}%
\pgfpathlineto{\pgfqpoint{2.660665in}{0.612914in}}%
\pgfpathlineto{\pgfqpoint{2.660665in}{0.612914in}}%
\pgfpathlineto{\pgfqpoint{2.750283in}{0.612914in}}%
\pgfpathlineto{\pgfqpoint{2.750283in}{0.612914in}}%
\pgfpathlineto{\pgfqpoint{2.839901in}{0.612914in}}%
\pgfpathlineto{\pgfqpoint{2.839901in}{0.612914in}}%
\pgfpathlineto{\pgfqpoint{2.929520in}{0.612914in}}%
\pgfpathlineto{\pgfqpoint{2.929520in}{0.612914in}}%
\pgfpathlineto{\pgfqpoint{3.019138in}{0.612914in}}%
\pgfpathlineto{\pgfqpoint{3.019138in}{0.612914in}}%
\pgfpathlineto{\pgfqpoint{3.108756in}{0.612914in}}%
\pgfpathlineto{\pgfqpoint{3.108756in}{0.612914in}}%
\pgfpathlineto{\pgfqpoint{3.198375in}{0.612914in}}%
\pgfpathlineto{\pgfqpoint{3.198375in}{0.553219in}}%
\pgfusepath{stroke}%
\end{pgfscope}%
\begin{pgfscope}%
\pgfsetrectcap%
\pgfsetmiterjoin%
\pgfsetlinewidth{0.803000pt}%
\definecolor{currentstroke}{rgb}{0.000000,0.000000,0.000000}%
\pgfsetstrokecolor{currentstroke}%
\pgfsetdash{}{0pt}%
\pgfpathmoveto{\pgfqpoint{0.375397in}{0.484570in}}%
\pgfpathlineto{\pgfqpoint{0.375397in}{1.994851in}}%
\pgfusepath{stroke}%
\end{pgfscope}%
\begin{pgfscope}%
\pgfsetrectcap%
\pgfsetmiterjoin%
\pgfsetlinewidth{0.803000pt}%
\definecolor{currentstroke}{rgb}{0.000000,0.000000,0.000000}%
\pgfsetstrokecolor{currentstroke}%
\pgfsetdash{}{0pt}%
\pgfpathmoveto{\pgfqpoint{3.332802in}{0.484570in}}%
\pgfpathlineto{\pgfqpoint{3.332802in}{1.994851in}}%
\pgfusepath{stroke}%
\end{pgfscope}%
\begin{pgfscope}%
\pgfsetrectcap%
\pgfsetmiterjoin%
\pgfsetlinewidth{0.803000pt}%
\definecolor{currentstroke}{rgb}{0.000000,0.000000,0.000000}%
\pgfsetstrokecolor{currentstroke}%
\pgfsetdash{}{0pt}%
\pgfpathmoveto{\pgfqpoint{0.375397in}{0.484570in}}%
\pgfpathlineto{\pgfqpoint{3.332802in}{0.484570in}}%
\pgfusepath{stroke}%
\end{pgfscope}%
\begin{pgfscope}%
\pgfsetrectcap%
\pgfsetmiterjoin%
\pgfsetlinewidth{0.803000pt}%
\definecolor{currentstroke}{rgb}{0.000000,0.000000,0.000000}%
\pgfsetstrokecolor{currentstroke}%
\pgfsetdash{}{0pt}%
\pgfpathmoveto{\pgfqpoint{0.375397in}{1.994851in}}%
\pgfpathlineto{\pgfqpoint{3.332802in}{1.994851in}}%
\pgfusepath{stroke}%
\end{pgfscope}%
\begin{pgfscope}%
\pgfsetbuttcap%
\pgfsetmiterjoin%
\definecolor{currentfill}{rgb}{1.000000,1.000000,1.000000}%
\pgfsetfillcolor{currentfill}%
\pgfsetfillopacity{0.800000}%
\pgfsetlinewidth{1.003750pt}%
\definecolor{currentstroke}{rgb}{0.800000,0.800000,0.800000}%
\pgfsetstrokecolor{currentstroke}%
\pgfsetstrokeopacity{0.800000}%
\pgfsetdash{}{0pt}%
\pgfpathmoveto{\pgfqpoint{2.293881in}{1.510374in}}%
\pgfpathlineto{\pgfqpoint{3.264746in}{1.510374in}}%
\pgfpathquadraticcurveto{\pgfqpoint{3.284191in}{1.510374in}}{\pgfqpoint{3.284191in}{1.529818in}}%
\pgfpathlineto{\pgfqpoint{3.284191in}{1.926795in}}%
\pgfpathquadraticcurveto{\pgfqpoint{3.284191in}{1.946240in}}{\pgfqpoint{3.264746in}{1.946240in}}%
\pgfpathlineto{\pgfqpoint{2.293881in}{1.946240in}}%
\pgfpathquadraticcurveto{\pgfqpoint{2.274437in}{1.946240in}}{\pgfqpoint{2.274437in}{1.926795in}}%
\pgfpathlineto{\pgfqpoint{2.274437in}{1.529818in}}%
\pgfpathquadraticcurveto{\pgfqpoint{2.274437in}{1.510374in}}{\pgfqpoint{2.293881in}{1.510374in}}%
\pgfpathclose%
\pgfusepath{stroke,fill}%
\end{pgfscope}%
\begin{pgfscope}%
\pgfsetrectcap%
\pgfsetroundjoin%
\pgfsetlinewidth{1.505625pt}%
\definecolor{currentstroke}{rgb}{0.121569,0.466667,0.705882}%
\pgfsetstrokecolor{currentstroke}%
\pgfsetdash{}{0pt}%
\pgfpathmoveto{\pgfqpoint{2.313326in}{1.873323in}}%
\pgfpathlineto{\pgfqpoint{2.507770in}{1.873323in}}%
\pgfusepath{stroke}%
\end{pgfscope}%
\begin{pgfscope}%
\pgftext[x=2.585548in,y=1.839295in,left,base]{\rmfamily\fontsize{7.000000}{8.400000}\selectfont image}%
\end{pgfscope}%
\begin{pgfscope}%
\pgfsetrectcap%
\pgfsetroundjoin%
\pgfsetlinewidth{1.505625pt}%
\definecolor{currentstroke}{rgb}{1.000000,0.498039,0.054902}%
\pgfsetstrokecolor{currentstroke}%
\pgfsetdash{}{0pt}%
\pgfpathmoveto{\pgfqpoint{2.313326in}{1.737757in}}%
\pgfpathlineto{\pgfqpoint{2.507770in}{1.737757in}}%
\pgfusepath{stroke}%
\end{pgfscope}%
\begin{pgfscope}%
\pgftext[x=2.585548in,y=1.703729in,left,base]{\rmfamily\fontsize{7.000000}{8.400000}\selectfont text}%
\end{pgfscope}%
\begin{pgfscope}%
\pgfsetrectcap%
\pgfsetroundjoin%
\pgfsetlinewidth{1.505625pt}%
\definecolor{currentstroke}{rgb}{0.172549,0.627451,0.172549}%
\pgfsetstrokecolor{currentstroke}%
\pgfsetdash{}{0pt}%
\pgfpathmoveto{\pgfqpoint{2.313326in}{1.602190in}}%
\pgfpathlineto{\pgfqpoint{2.507770in}{1.602190in}}%
\pgfusepath{stroke}%
\end{pgfscope}%
\begin{pgfscope}%
\pgftext[x=2.585548in,y=1.568162in,left,base]{\rmfamily\fontsize{7.000000}{8.400000}\selectfont reinforcement}%
\end{pgfscope}%
\end{pgfpicture}%
\makeatother%
\endgroup%

%% file: figures/paper/sub_nodes_ccdf.pgf
\begingroup%
\makeatletter%
\begin{pgfpicture}%
\pgfpathrectangle{\pgfpointorigin}{\pgfqpoint{3.502802in}{2.164851in}}%
\pgfusepath{use as bounding box, clip}%
\begin{pgfscope}%
\pgfsetbuttcap%
\pgfsetmiterjoin%
\pgfsetlinewidth{0.000000pt}%
\definecolor{currentstroke}{rgb}{1.000000,1.000000,1.000000}%
\pgfsetstrokecolor{currentstroke}%
\pgfsetstrokeopacity{0.000000}%
\pgfsetdash{}{0pt}%
\pgfpathmoveto{\pgfqpoint{0.000000in}{0.000000in}}%
\pgfpathlineto{\pgfqpoint{3.502802in}{0.000000in}}%
\pgfpathlineto{\pgfqpoint{3.502802in}{2.164851in}}%
\pgfpathlineto{\pgfqpoint{0.000000in}{2.164851in}}%
\pgfpathclose%
\pgfusepath{}%
\end{pgfscope}%
\begin{pgfscope}%
\pgfsetbuttcap%
\pgfsetmiterjoin%
\definecolor{currentfill}{rgb}{1.000000,1.000000,1.000000}%
\pgfsetfillcolor{currentfill}%
\pgfsetlinewidth{0.000000pt}%
\definecolor{currentstroke}{rgb}{0.000000,0.000000,0.000000}%
\pgfsetstrokecolor{currentstroke}%
\pgfsetstrokeopacity{0.000000}%
\pgfsetdash{}{0pt}%
\pgfpathmoveto{\pgfqpoint{0.375397in}{0.484570in}}%
\pgfpathlineto{\pgfqpoint{3.332802in}{0.484570in}}%
\pgfpathlineto{\pgfqpoint{3.332802in}{1.994851in}}%
\pgfpathlineto{\pgfqpoint{0.375397in}{1.994851in}}%
\pgfpathclose%
\pgfusepath{fill}%
\end{pgfscope}%
\begin{pgfscope}%
\pgfpathrectangle{\pgfqpoint{0.375397in}{0.484570in}}{\pgfqpoint{2.957405in}{1.510280in}} %
\pgfusepath{clip}%
\pgfsetrectcap%
\pgfsetroundjoin%
\pgfsetlinewidth{0.803000pt}%
\definecolor{currentstroke}{rgb}{0.690196,0.690196,0.690196}%
\pgfsetstrokecolor{currentstroke}%
\pgfsetdash{}{0pt}%
\pgfpathmoveto{\pgfqpoint{0.509825in}{0.484570in}}%
\pgfpathlineto{\pgfqpoint{0.509825in}{1.994851in}}%
\pgfusepath{stroke}%
\end{pgfscope}%
\begin{pgfscope}%
\pgfsetbuttcap%
\pgfsetroundjoin%
\definecolor{currentfill}{rgb}{0.000000,0.000000,0.000000}%
\pgfsetfillcolor{currentfill}%
\pgfsetlinewidth{0.803000pt}%
\definecolor{currentstroke}{rgb}{0.000000,0.000000,0.000000}%
\pgfsetstrokecolor{currentstroke}%
\pgfsetdash{}{0pt}%
\pgfsys@defobject{currentmarker}{\pgfqpoint{0.000000in}{-0.048611in}}{\pgfqpoint{0.000000in}{0.000000in}}{%
\pgfpathmoveto{\pgfqpoint{0.000000in}{0.000000in}}%
\pgfpathlineto{\pgfqpoint{0.000000in}{-0.048611in}}%
\pgfusepath{stroke,fill}%
}%
\begin{pgfscope}%
\pgfsys@transformshift{0.509825in}{0.484570in}%
\pgfsys@useobject{currentmarker}{}%
\end{pgfscope}%
\end{pgfscope}%
\begin{pgfscope}%
\pgftext[x=0.509825in,y=0.387348in,,top]{\rmfamily\fontsize{7.000000}{8.400000}\selectfont \(\displaystyle 0\)}%
\end{pgfscope}%
\begin{pgfscope}%
\pgfpathrectangle{\pgfqpoint{0.375397in}{0.484570in}}{\pgfqpoint{2.957405in}{1.510280in}} %
\pgfusepath{clip}%
\pgfsetrectcap%
\pgfsetroundjoin%
\pgfsetlinewidth{0.803000pt}%
\definecolor{currentstroke}{rgb}{0.690196,0.690196,0.690196}%
\pgfsetstrokecolor{currentstroke}%
\pgfsetdash{}{0pt}%
\pgfpathmoveto{\pgfqpoint{0.957916in}{0.484570in}}%
\pgfpathlineto{\pgfqpoint{0.957916in}{1.994851in}}%
\pgfusepath{stroke}%
\end{pgfscope}%
\begin{pgfscope}%
\pgfsetbuttcap%
\pgfsetroundjoin%
\definecolor{currentfill}{rgb}{0.000000,0.000000,0.000000}%
\pgfsetfillcolor{currentfill}%
\pgfsetlinewidth{0.803000pt}%
\definecolor{currentstroke}{rgb}{0.000000,0.000000,0.000000}%
\pgfsetstrokecolor{currentstroke}%
\pgfsetdash{}{0pt}%
\pgfsys@defobject{currentmarker}{\pgfqpoint{0.000000in}{-0.048611in}}{\pgfqpoint{0.000000in}{0.000000in}}{%
\pgfpathmoveto{\pgfqpoint{0.000000in}{0.000000in}}%
\pgfpathlineto{\pgfqpoint{0.000000in}{-0.048611in}}%
\pgfusepath{stroke,fill}%
}%
\begin{pgfscope}%
\pgfsys@transformshift{0.957916in}{0.484570in}%
\pgfsys@useobject{currentmarker}{}%
\end{pgfscope}%
\end{pgfscope}%
\begin{pgfscope}%
\pgftext[x=0.957916in,y=0.387348in,,top]{\rmfamily\fontsize{7.000000}{8.400000}\selectfont \(\displaystyle 5\)}%
\end{pgfscope}%
\begin{pgfscope}%
\pgfpathrectangle{\pgfqpoint{0.375397in}{0.484570in}}{\pgfqpoint{2.957405in}{1.510280in}} %
\pgfusepath{clip}%
\pgfsetrectcap%
\pgfsetroundjoin%
\pgfsetlinewidth{0.803000pt}%
\definecolor{currentstroke}{rgb}{0.690196,0.690196,0.690196}%
\pgfsetstrokecolor{currentstroke}%
\pgfsetdash{}{0pt}%
\pgfpathmoveto{\pgfqpoint{1.406008in}{0.484570in}}%
\pgfpathlineto{\pgfqpoint{1.406008in}{1.994851in}}%
\pgfusepath{stroke}%
\end{pgfscope}%
\begin{pgfscope}%
\pgfsetbuttcap%
\pgfsetroundjoin%
\definecolor{currentfill}{rgb}{0.000000,0.000000,0.000000}%
\pgfsetfillcolor{currentfill}%
\pgfsetlinewidth{0.803000pt}%
\definecolor{currentstroke}{rgb}{0.000000,0.000000,0.000000}%
\pgfsetstrokecolor{currentstroke}%
\pgfsetdash{}{0pt}%
\pgfsys@defobject{currentmarker}{\pgfqpoint{0.000000in}{-0.048611in}}{\pgfqpoint{0.000000in}{0.000000in}}{%
\pgfpathmoveto{\pgfqpoint{0.000000in}{0.000000in}}%
\pgfpathlineto{\pgfqpoint{0.000000in}{-0.048611in}}%
\pgfusepath{stroke,fill}%
}%
\begin{pgfscope}%
\pgfsys@transformshift{1.406008in}{0.484570in}%
\pgfsys@useobject{currentmarker}{}%
\end{pgfscope}%
\end{pgfscope}%
\begin{pgfscope}%
\pgftext[x=1.406008in,y=0.387348in,,top]{\rmfamily\fontsize{7.000000}{8.400000}\selectfont \(\displaystyle 10\)}%
\end{pgfscope}%
\begin{pgfscope}%
\pgfpathrectangle{\pgfqpoint{0.375397in}{0.484570in}}{\pgfqpoint{2.957405in}{1.510280in}} %
\pgfusepath{clip}%
\pgfsetrectcap%
\pgfsetroundjoin%
\pgfsetlinewidth{0.803000pt}%
\definecolor{currentstroke}{rgb}{0.690196,0.690196,0.690196}%
\pgfsetstrokecolor{currentstroke}%
\pgfsetdash{}{0pt}%
\pgfpathmoveto{\pgfqpoint{1.854100in}{0.484570in}}%
\pgfpathlineto{\pgfqpoint{1.854100in}{1.994851in}}%
\pgfusepath{stroke}%
\end{pgfscope}%
\begin{pgfscope}%
\pgfsetbuttcap%
\pgfsetroundjoin%
\definecolor{currentfill}{rgb}{0.000000,0.000000,0.000000}%
\pgfsetfillcolor{currentfill}%
\pgfsetlinewidth{0.803000pt}%
\definecolor{currentstroke}{rgb}{0.000000,0.000000,0.000000}%
\pgfsetstrokecolor{currentstroke}%
\pgfsetdash{}{0pt}%
\pgfsys@defobject{currentmarker}{\pgfqpoint{0.000000in}{-0.048611in}}{\pgfqpoint{0.000000in}{0.000000in}}{%
\pgfpathmoveto{\pgfqpoint{0.000000in}{0.000000in}}%
\pgfpathlineto{\pgfqpoint{0.000000in}{-0.048611in}}%
\pgfusepath{stroke,fill}%
}%
\begin{pgfscope}%
\pgfsys@transformshift{1.854100in}{0.484570in}%
\pgfsys@useobject{currentmarker}{}%
\end{pgfscope}%
\end{pgfscope}%
\begin{pgfscope}%
\pgftext[x=1.854100in,y=0.387348in,,top]{\rmfamily\fontsize{7.000000}{8.400000}\selectfont \(\displaystyle 15\)}%
\end{pgfscope}%
\begin{pgfscope}%
\pgfpathrectangle{\pgfqpoint{0.375397in}{0.484570in}}{\pgfqpoint{2.957405in}{1.510280in}} %
\pgfusepath{clip}%
\pgfsetrectcap%
\pgfsetroundjoin%
\pgfsetlinewidth{0.803000pt}%
\definecolor{currentstroke}{rgb}{0.690196,0.690196,0.690196}%
\pgfsetstrokecolor{currentstroke}%
\pgfsetdash{}{0pt}%
\pgfpathmoveto{\pgfqpoint{2.302191in}{0.484570in}}%
\pgfpathlineto{\pgfqpoint{2.302191in}{1.994851in}}%
\pgfusepath{stroke}%
\end{pgfscope}%
\begin{pgfscope}%
\pgfsetbuttcap%
\pgfsetroundjoin%
\definecolor{currentfill}{rgb}{0.000000,0.000000,0.000000}%
\pgfsetfillcolor{currentfill}%
\pgfsetlinewidth{0.803000pt}%
\definecolor{currentstroke}{rgb}{0.000000,0.000000,0.000000}%
\pgfsetstrokecolor{currentstroke}%
\pgfsetdash{}{0pt}%
\pgfsys@defobject{currentmarker}{\pgfqpoint{0.000000in}{-0.048611in}}{\pgfqpoint{0.000000in}{0.000000in}}{%
\pgfpathmoveto{\pgfqpoint{0.000000in}{0.000000in}}%
\pgfpathlineto{\pgfqpoint{0.000000in}{-0.048611in}}%
\pgfusepath{stroke,fill}%
}%
\begin{pgfscope}%
\pgfsys@transformshift{2.302191in}{0.484570in}%
\pgfsys@useobject{currentmarker}{}%
\end{pgfscope}%
\end{pgfscope}%
\begin{pgfscope}%
\pgftext[x=2.302191in,y=0.387348in,,top]{\rmfamily\fontsize{7.000000}{8.400000}\selectfont \(\displaystyle 20\)}%
\end{pgfscope}%
\begin{pgfscope}%
\pgfpathrectangle{\pgfqpoint{0.375397in}{0.484570in}}{\pgfqpoint{2.957405in}{1.510280in}} %
\pgfusepath{clip}%
\pgfsetrectcap%
\pgfsetroundjoin%
\pgfsetlinewidth{0.803000pt}%
\definecolor{currentstroke}{rgb}{0.690196,0.690196,0.690196}%
\pgfsetstrokecolor{currentstroke}%
\pgfsetdash{}{0pt}%
\pgfpathmoveto{\pgfqpoint{2.750283in}{0.484570in}}%
\pgfpathlineto{\pgfqpoint{2.750283in}{1.994851in}}%
\pgfusepath{stroke}%
\end{pgfscope}%
\begin{pgfscope}%
\pgfsetbuttcap%
\pgfsetroundjoin%
\definecolor{currentfill}{rgb}{0.000000,0.000000,0.000000}%
\pgfsetfillcolor{currentfill}%
\pgfsetlinewidth{0.803000pt}%
\definecolor{currentstroke}{rgb}{0.000000,0.000000,0.000000}%
\pgfsetstrokecolor{currentstroke}%
\pgfsetdash{}{0pt}%
\pgfsys@defobject{currentmarker}{\pgfqpoint{0.000000in}{-0.048611in}}{\pgfqpoint{0.000000in}{0.000000in}}{%
\pgfpathmoveto{\pgfqpoint{0.000000in}{0.000000in}}%
\pgfpathlineto{\pgfqpoint{0.000000in}{-0.048611in}}%
\pgfusepath{stroke,fill}%
}%
\begin{pgfscope}%
\pgfsys@transformshift{2.750283in}{0.484570in}%
\pgfsys@useobject{currentmarker}{}%
\end{pgfscope}%
\end{pgfscope}%
\begin{pgfscope}%
\pgftext[x=2.750283in,y=0.387348in,,top]{\rmfamily\fontsize{7.000000}{8.400000}\selectfont \(\displaystyle 25\)}%
\end{pgfscope}%
\begin{pgfscope}%
\pgfpathrectangle{\pgfqpoint{0.375397in}{0.484570in}}{\pgfqpoint{2.957405in}{1.510280in}} %
\pgfusepath{clip}%
\pgfsetrectcap%
\pgfsetroundjoin%
\pgfsetlinewidth{0.803000pt}%
\definecolor{currentstroke}{rgb}{0.690196,0.690196,0.690196}%
\pgfsetstrokecolor{currentstroke}%
\pgfsetdash{}{0pt}%
\pgfpathmoveto{\pgfqpoint{3.198375in}{0.484570in}}%
\pgfpathlineto{\pgfqpoint{3.198375in}{1.994851in}}%
\pgfusepath{stroke}%
\end{pgfscope}%
\begin{pgfscope}%
\pgfsetbuttcap%
\pgfsetroundjoin%
\definecolor{currentfill}{rgb}{0.000000,0.000000,0.000000}%
\pgfsetfillcolor{currentfill}%
\pgfsetlinewidth{0.803000pt}%
\definecolor{currentstroke}{rgb}{0.000000,0.000000,0.000000}%
\pgfsetstrokecolor{currentstroke}%
\pgfsetdash{}{0pt}%
\pgfsys@defobject{currentmarker}{\pgfqpoint{0.000000in}{-0.048611in}}{\pgfqpoint{0.000000in}{0.000000in}}{%
\pgfpathmoveto{\pgfqpoint{0.000000in}{0.000000in}}%
\pgfpathlineto{\pgfqpoint{0.000000in}{-0.048611in}}%
\pgfusepath{stroke,fill}%
}%
\begin{pgfscope}%
\pgfsys@transformshift{3.198375in}{0.484570in}%
\pgfsys@useobject{currentmarker}{}%
\end{pgfscope}%
\end{pgfscope}%
\begin{pgfscope}%
\pgftext[x=3.198375in,y=0.387348in,,top]{\rmfamily\fontsize{7.000000}{8.400000}\selectfont \(\displaystyle 30\)}%
\end{pgfscope}%
\begin{pgfscope}%
\pgftext[x=1.854100in,y=0.245934in,,top]{\rmfamily\fontsize{9.000000}{10.800000}\selectfont number of nodes}%
\end{pgfscope}%
\begin{pgfscope}%
\pgfpathrectangle{\pgfqpoint{0.375397in}{0.484570in}}{\pgfqpoint{2.957405in}{1.510280in}} %
\pgfusepath{clip}%
\pgfsetrectcap%
\pgfsetroundjoin%
\pgfsetlinewidth{0.803000pt}%
\definecolor{currentstroke}{rgb}{0.690196,0.690196,0.690196}%
\pgfsetstrokecolor{currentstroke}%
\pgfsetdash{}{0pt}%
\pgfpathmoveto{\pgfqpoint{0.375397in}{0.553219in}}%
\pgfpathlineto{\pgfqpoint{3.332802in}{0.553219in}}%
\pgfusepath{stroke}%
\end{pgfscope}%
\begin{pgfscope}%
\pgfsetbuttcap%
\pgfsetroundjoin%
\definecolor{currentfill}{rgb}{0.000000,0.000000,0.000000}%
\pgfsetfillcolor{currentfill}%
\pgfsetlinewidth{0.803000pt}%
\definecolor{currentstroke}{rgb}{0.000000,0.000000,0.000000}%
\pgfsetstrokecolor{currentstroke}%
\pgfsetdash{}{0pt}%
\pgfsys@defobject{currentmarker}{\pgfqpoint{-0.048611in}{0.000000in}}{\pgfqpoint{0.000000in}{0.000000in}}{%
\pgfpathmoveto{\pgfqpoint{0.000000in}{0.000000in}}%
\pgfpathlineto{\pgfqpoint{-0.048611in}{0.000000in}}%
\pgfusepath{stroke,fill}%
}%
\begin{pgfscope}%
\pgfsys@transformshift{0.375397in}{0.553219in}%
\pgfsys@useobject{currentmarker}{}%
\end{pgfscope}%
\end{pgfscope}%
\begin{pgfscope}%
\pgftext[x=0.134463in,y=0.519740in,left,base]{\rmfamily\fontsize{7.000000}{8.400000}\selectfont \(\displaystyle 0.0\)}%
\end{pgfscope}%
\begin{pgfscope}%
\pgfpathrectangle{\pgfqpoint{0.375397in}{0.484570in}}{\pgfqpoint{2.957405in}{1.510280in}} %
\pgfusepath{clip}%
\pgfsetrectcap%
\pgfsetroundjoin%
\pgfsetlinewidth{0.803000pt}%
\definecolor{currentstroke}{rgb}{0.690196,0.690196,0.690196}%
\pgfsetstrokecolor{currentstroke}%
\pgfsetdash{}{0pt}%
\pgfpathmoveto{\pgfqpoint{0.375397in}{0.827816in}}%
\pgfpathlineto{\pgfqpoint{3.332802in}{0.827816in}}%
\pgfusepath{stroke}%
\end{pgfscope}%
\begin{pgfscope}%
\pgfsetbuttcap%
\pgfsetroundjoin%
\definecolor{currentfill}{rgb}{0.000000,0.000000,0.000000}%
\pgfsetfillcolor{currentfill}%
\pgfsetlinewidth{0.803000pt}%
\definecolor{currentstroke}{rgb}{0.000000,0.000000,0.000000}%
\pgfsetstrokecolor{currentstroke}%
\pgfsetdash{}{0pt}%
\pgfsys@defobject{currentmarker}{\pgfqpoint{-0.048611in}{0.000000in}}{\pgfqpoint{0.000000in}{0.000000in}}{%
\pgfpathmoveto{\pgfqpoint{0.000000in}{0.000000in}}%
\pgfpathlineto{\pgfqpoint{-0.048611in}{0.000000in}}%
\pgfusepath{stroke,fill}%
}%
\begin{pgfscope}%
\pgfsys@transformshift{0.375397in}{0.827816in}%
\pgfsys@useobject{currentmarker}{}%
\end{pgfscope}%
\end{pgfscope}%
\begin{pgfscope}%
\pgftext[x=0.134463in,y=0.794336in,left,base]{\rmfamily\fontsize{7.000000}{8.400000}\selectfont \(\displaystyle 0.2\)}%
\end{pgfscope}%
\begin{pgfscope}%
\pgfpathrectangle{\pgfqpoint{0.375397in}{0.484570in}}{\pgfqpoint{2.957405in}{1.510280in}} %
\pgfusepath{clip}%
\pgfsetrectcap%
\pgfsetroundjoin%
\pgfsetlinewidth{0.803000pt}%
\definecolor{currentstroke}{rgb}{0.690196,0.690196,0.690196}%
\pgfsetstrokecolor{currentstroke}%
\pgfsetdash{}{0pt}%
\pgfpathmoveto{\pgfqpoint{0.375397in}{1.102412in}}%
\pgfpathlineto{\pgfqpoint{3.332802in}{1.102412in}}%
\pgfusepath{stroke}%
\end{pgfscope}%
\begin{pgfscope}%
\pgfsetbuttcap%
\pgfsetroundjoin%
\definecolor{currentfill}{rgb}{0.000000,0.000000,0.000000}%
\pgfsetfillcolor{currentfill}%
\pgfsetlinewidth{0.803000pt}%
\definecolor{currentstroke}{rgb}{0.000000,0.000000,0.000000}%
\pgfsetstrokecolor{currentstroke}%
\pgfsetdash{}{0pt}%
\pgfsys@defobject{currentmarker}{\pgfqpoint{-0.048611in}{0.000000in}}{\pgfqpoint{0.000000in}{0.000000in}}{%
\pgfpathmoveto{\pgfqpoint{0.000000in}{0.000000in}}%
\pgfpathlineto{\pgfqpoint{-0.048611in}{0.000000in}}%
\pgfusepath{stroke,fill}%
}%
\begin{pgfscope}%
\pgfsys@transformshift{0.375397in}{1.102412in}%
\pgfsys@useobject{currentmarker}{}%
\end{pgfscope}%
\end{pgfscope}%
\begin{pgfscope}%
\pgftext[x=0.134463in,y=1.068933in,left,base]{\rmfamily\fontsize{7.000000}{8.400000}\selectfont \(\displaystyle 0.4\)}%
\end{pgfscope}%
\begin{pgfscope}%
\pgfpathrectangle{\pgfqpoint{0.375397in}{0.484570in}}{\pgfqpoint{2.957405in}{1.510280in}} %
\pgfusepath{clip}%
\pgfsetrectcap%
\pgfsetroundjoin%
\pgfsetlinewidth{0.803000pt}%
\definecolor{currentstroke}{rgb}{0.690196,0.690196,0.690196}%
\pgfsetstrokecolor{currentstroke}%
\pgfsetdash{}{0pt}%
\pgfpathmoveto{\pgfqpoint{0.375397in}{1.377009in}}%
\pgfpathlineto{\pgfqpoint{3.332802in}{1.377009in}}%
\pgfusepath{stroke}%
\end{pgfscope}%
\begin{pgfscope}%
\pgfsetbuttcap%
\pgfsetroundjoin%
\definecolor{currentfill}{rgb}{0.000000,0.000000,0.000000}%
\pgfsetfillcolor{currentfill}%
\pgfsetlinewidth{0.803000pt}%
\definecolor{currentstroke}{rgb}{0.000000,0.000000,0.000000}%
\pgfsetstrokecolor{currentstroke}%
\pgfsetdash{}{0pt}%
\pgfsys@defobject{currentmarker}{\pgfqpoint{-0.048611in}{0.000000in}}{\pgfqpoint{0.000000in}{0.000000in}}{%
\pgfpathmoveto{\pgfqpoint{0.000000in}{0.000000in}}%
\pgfpathlineto{\pgfqpoint{-0.048611in}{0.000000in}}%
\pgfusepath{stroke,fill}%
}%
\begin{pgfscope}%
\pgfsys@transformshift{0.375397in}{1.377009in}%
\pgfsys@useobject{currentmarker}{}%
\end{pgfscope}%
\end{pgfscope}%
\begin{pgfscope}%
\pgftext[x=0.134463in,y=1.343529in,left,base]{\rmfamily\fontsize{7.000000}{8.400000}\selectfont \(\displaystyle 0.6\)}%
\end{pgfscope}%
\begin{pgfscope}%
\pgfpathrectangle{\pgfqpoint{0.375397in}{0.484570in}}{\pgfqpoint{2.957405in}{1.510280in}} %
\pgfusepath{clip}%
\pgfsetrectcap%
\pgfsetroundjoin%
\pgfsetlinewidth{0.803000pt}%
\definecolor{currentstroke}{rgb}{0.690196,0.690196,0.690196}%
\pgfsetstrokecolor{currentstroke}%
\pgfsetdash{}{0pt}%
\pgfpathmoveto{\pgfqpoint{0.375397in}{1.651605in}}%
\pgfpathlineto{\pgfqpoint{3.332802in}{1.651605in}}%
\pgfusepath{stroke}%
\end{pgfscope}%
\begin{pgfscope}%
\pgfsetbuttcap%
\pgfsetroundjoin%
\definecolor{currentfill}{rgb}{0.000000,0.000000,0.000000}%
\pgfsetfillcolor{currentfill}%
\pgfsetlinewidth{0.803000pt}%
\definecolor{currentstroke}{rgb}{0.000000,0.000000,0.000000}%
\pgfsetstrokecolor{currentstroke}%
\pgfsetdash{}{0pt}%
\pgfsys@defobject{currentmarker}{\pgfqpoint{-0.048611in}{0.000000in}}{\pgfqpoint{0.000000in}{0.000000in}}{%
\pgfpathmoveto{\pgfqpoint{0.000000in}{0.000000in}}%
\pgfpathlineto{\pgfqpoint{-0.048611in}{0.000000in}}%
\pgfusepath{stroke,fill}%
}%
\begin{pgfscope}%
\pgfsys@transformshift{0.375397in}{1.651605in}%
\pgfsys@useobject{currentmarker}{}%
\end{pgfscope}%
\end{pgfscope}%
\begin{pgfscope}%
\pgftext[x=0.134463in,y=1.618126in,left,base]{\rmfamily\fontsize{7.000000}{8.400000}\selectfont \(\displaystyle 0.8\)}%
\end{pgfscope}%
\begin{pgfscope}%
\pgfpathrectangle{\pgfqpoint{0.375397in}{0.484570in}}{\pgfqpoint{2.957405in}{1.510280in}} %
\pgfusepath{clip}%
\pgfsetrectcap%
\pgfsetroundjoin%
\pgfsetlinewidth{0.803000pt}%
\definecolor{currentstroke}{rgb}{0.690196,0.690196,0.690196}%
\pgfsetstrokecolor{currentstroke}%
\pgfsetdash{}{0pt}%
\pgfpathmoveto{\pgfqpoint{0.375397in}{1.926202in}}%
\pgfpathlineto{\pgfqpoint{3.332802in}{1.926202in}}%
\pgfusepath{stroke}%
\end{pgfscope}%
\begin{pgfscope}%
\pgfsetbuttcap%
\pgfsetroundjoin%
\definecolor{currentfill}{rgb}{0.000000,0.000000,0.000000}%
\pgfsetfillcolor{currentfill}%
\pgfsetlinewidth{0.803000pt}%
\definecolor{currentstroke}{rgb}{0.000000,0.000000,0.000000}%
\pgfsetstrokecolor{currentstroke}%
\pgfsetdash{}{0pt}%
\pgfsys@defobject{currentmarker}{\pgfqpoint{-0.048611in}{0.000000in}}{\pgfqpoint{0.000000in}{0.000000in}}{%
\pgfpathmoveto{\pgfqpoint{0.000000in}{0.000000in}}%
\pgfpathlineto{\pgfqpoint{-0.048611in}{0.000000in}}%
\pgfusepath{stroke,fill}%
}%
\begin{pgfscope}%
\pgfsys@transformshift{0.375397in}{1.926202in}%
\pgfsys@useobject{currentmarker}{}%
\end{pgfscope}%
\end{pgfscope}%
\begin{pgfscope}%
\pgftext[x=0.134463in,y=1.892722in,left,base]{\rmfamily\fontsize{7.000000}{8.400000}\selectfont \(\displaystyle 1.0\)}%
\end{pgfscope}%
\begin{pgfscope}%
\pgfpathrectangle{\pgfqpoint{0.375397in}{0.484570in}}{\pgfqpoint{2.957405in}{1.510280in}} %
\pgfusepath{clip}%
\pgfsetrectcap%
\pgfsetroundjoin%
\pgfsetlinewidth{1.505625pt}%
\definecolor{currentstroke}{rgb}{0.121569,0.466667,0.705882}%
\pgfsetstrokecolor{currentstroke}%
\pgfsetdash{}{0pt}%
\pgfpathmoveto{\pgfqpoint{0.509825in}{1.926202in}}%
\pgfpathlineto{\pgfqpoint{0.599443in}{1.926202in}}%
\pgfpathlineto{\pgfqpoint{0.599443in}{1.926202in}}%
\pgfpathlineto{\pgfqpoint{0.689061in}{1.926202in}}%
\pgfpathlineto{\pgfqpoint{0.689061in}{1.926202in}}%
\pgfpathlineto{\pgfqpoint{0.778680in}{1.926202in}}%
\pgfpathlineto{\pgfqpoint{0.778680in}{1.311973in}}%
\pgfpathlineto{\pgfqpoint{0.868298in}{1.311973in}}%
\pgfpathlineto{\pgfqpoint{0.868298in}{1.203579in}}%
\pgfpathlineto{\pgfqpoint{0.957916in}{1.203579in}}%
\pgfpathlineto{\pgfqpoint{0.957916in}{1.131317in}}%
\pgfpathlineto{\pgfqpoint{1.047535in}{1.131317in}}%
\pgfpathlineto{\pgfqpoint{1.047535in}{1.059055in}}%
\pgfpathlineto{\pgfqpoint{1.137153in}{1.059055in}}%
\pgfpathlineto{\pgfqpoint{1.137153in}{1.022924in}}%
\pgfpathlineto{\pgfqpoint{1.226771in}{1.022924in}}%
\pgfpathlineto{\pgfqpoint{1.226771in}{0.914531in}}%
\pgfpathlineto{\pgfqpoint{1.316390in}{0.914531in}}%
\pgfpathlineto{\pgfqpoint{1.316390in}{0.878399in}}%
\pgfpathlineto{\pgfqpoint{1.406008in}{0.878399in}}%
\pgfpathlineto{\pgfqpoint{1.406008in}{0.842268in}}%
\pgfpathlineto{\pgfqpoint{1.495626in}{0.842268in}}%
\pgfpathlineto{\pgfqpoint{1.495626in}{0.806137in}}%
\pgfpathlineto{\pgfqpoint{1.585245in}{0.806137in}}%
\pgfpathlineto{\pgfqpoint{1.585245in}{0.661613in}}%
\pgfpathlineto{\pgfqpoint{1.674863in}{0.661613in}}%
\pgfpathlineto{\pgfqpoint{1.674863in}{0.625482in}}%
\pgfpathlineto{\pgfqpoint{1.764481in}{0.625482in}}%
\pgfpathlineto{\pgfqpoint{1.764481in}{0.589351in}}%
\pgfpathlineto{\pgfqpoint{1.854100in}{0.589351in}}%
\pgfpathlineto{\pgfqpoint{1.854100in}{0.553219in}}%
\pgfusepath{stroke}%
\end{pgfscope}%
\begin{pgfscope}%
\pgfpathrectangle{\pgfqpoint{0.375397in}{0.484570in}}{\pgfqpoint{2.957405in}{1.510280in}} %
\pgfusepath{clip}%
\pgfsetrectcap%
\pgfsetroundjoin%
\pgfsetlinewidth{1.505625pt}%
\definecolor{currentstroke}{rgb}{1.000000,0.498039,0.054902}%
\pgfsetstrokecolor{currentstroke}%
\pgfsetdash{}{0pt}%
\pgfpathmoveto{\pgfqpoint{0.509825in}{1.926202in}}%
\pgfpathlineto{\pgfqpoint{0.599443in}{1.926202in}}%
\pgfpathlineto{\pgfqpoint{0.599443in}{1.926202in}}%
\pgfpathlineto{\pgfqpoint{0.689061in}{1.926202in}}%
\pgfpathlineto{\pgfqpoint{0.689061in}{1.926202in}}%
\pgfpathlineto{\pgfqpoint{0.778680in}{1.926202in}}%
\pgfpathlineto{\pgfqpoint{0.778680in}{1.697371in}}%
\pgfpathlineto{\pgfqpoint{0.868298in}{1.697371in}}%
\pgfpathlineto{\pgfqpoint{0.868298in}{1.392264in}}%
\pgfpathlineto{\pgfqpoint{0.957916in}{1.392264in}}%
\pgfpathlineto{\pgfqpoint{0.957916in}{1.163434in}}%
\pgfpathlineto{\pgfqpoint{1.047535in}{1.163434in}}%
\pgfpathlineto{\pgfqpoint{1.047535in}{1.087157in}}%
\pgfpathlineto{\pgfqpoint{1.137153in}{1.087157in}}%
\pgfpathlineto{\pgfqpoint{1.137153in}{1.010880in}}%
\pgfpathlineto{\pgfqpoint{1.226771in}{1.010880in}}%
\pgfpathlineto{\pgfqpoint{1.226771in}{1.010880in}}%
\pgfpathlineto{\pgfqpoint{1.316390in}{1.010880in}}%
\pgfpathlineto{\pgfqpoint{1.316390in}{0.934603in}}%
\pgfpathlineto{\pgfqpoint{1.406008in}{0.934603in}}%
\pgfpathlineto{\pgfqpoint{1.406008in}{0.782050in}}%
\pgfpathlineto{\pgfqpoint{1.495626in}{0.782050in}}%
\pgfpathlineto{\pgfqpoint{1.495626in}{0.782050in}}%
\pgfpathlineto{\pgfqpoint{1.585245in}{0.782050in}}%
\pgfpathlineto{\pgfqpoint{1.585245in}{0.782050in}}%
\pgfpathlineto{\pgfqpoint{1.674863in}{0.782050in}}%
\pgfpathlineto{\pgfqpoint{1.674863in}{0.782050in}}%
\pgfpathlineto{\pgfqpoint{1.764481in}{0.782050in}}%
\pgfpathlineto{\pgfqpoint{1.764481in}{0.782050in}}%
\pgfpathlineto{\pgfqpoint{1.854100in}{0.782050in}}%
\pgfpathlineto{\pgfqpoint{1.854100in}{0.782050in}}%
\pgfpathlineto{\pgfqpoint{1.943718in}{0.782050in}}%
\pgfpathlineto{\pgfqpoint{1.943718in}{0.782050in}}%
\pgfpathlineto{\pgfqpoint{2.033336in}{0.782050in}}%
\pgfpathlineto{\pgfqpoint{2.033336in}{0.782050in}}%
\pgfpathlineto{\pgfqpoint{2.122955in}{0.782050in}}%
\pgfpathlineto{\pgfqpoint{2.122955in}{0.782050in}}%
\pgfpathlineto{\pgfqpoint{2.212573in}{0.782050in}}%
\pgfpathlineto{\pgfqpoint{2.212573in}{0.782050in}}%
\pgfpathlineto{\pgfqpoint{2.302191in}{0.782050in}}%
\pgfpathlineto{\pgfqpoint{2.302191in}{0.782050in}}%
\pgfpathlineto{\pgfqpoint{2.391810in}{0.782050in}}%
\pgfpathlineto{\pgfqpoint{2.391810in}{0.782050in}}%
\pgfpathlineto{\pgfqpoint{2.481428in}{0.782050in}}%
\pgfpathlineto{\pgfqpoint{2.481428in}{0.782050in}}%
\pgfpathlineto{\pgfqpoint{2.571046in}{0.782050in}}%
\pgfpathlineto{\pgfqpoint{2.571046in}{0.782050in}}%
\pgfpathlineto{\pgfqpoint{2.660665in}{0.782050in}}%
\pgfpathlineto{\pgfqpoint{2.660665in}{0.782050in}}%
\pgfpathlineto{\pgfqpoint{2.750283in}{0.782050in}}%
\pgfpathlineto{\pgfqpoint{2.750283in}{0.782050in}}%
\pgfpathlineto{\pgfqpoint{2.839901in}{0.782050in}}%
\pgfpathlineto{\pgfqpoint{2.839901in}{0.782050in}}%
\pgfpathlineto{\pgfqpoint{2.929520in}{0.782050in}}%
\pgfpathlineto{\pgfqpoint{2.929520in}{0.782050in}}%
\pgfpathlineto{\pgfqpoint{3.019138in}{0.782050in}}%
\pgfpathlineto{\pgfqpoint{3.019138in}{0.705773in}}%
\pgfpathlineto{\pgfqpoint{3.108756in}{0.705773in}}%
\pgfpathlineto{\pgfqpoint{3.108756in}{0.629496in}}%
\pgfpathlineto{\pgfqpoint{3.198375in}{0.629496in}}%
\pgfpathlineto{\pgfqpoint{3.198375in}{0.553219in}}%
\pgfusepath{stroke}%
\end{pgfscope}%
\begin{pgfscope}%
\pgfpathrectangle{\pgfqpoint{0.375397in}{0.484570in}}{\pgfqpoint{2.957405in}{1.510280in}} %
\pgfusepath{clip}%
\pgfsetrectcap%
\pgfsetroundjoin%
\pgfsetlinewidth{1.505625pt}%
\definecolor{currentstroke}{rgb}{0.172549,0.627451,0.172549}%
\pgfsetstrokecolor{currentstroke}%
\pgfsetdash{}{0pt}%
\pgfpathmoveto{\pgfqpoint{0.509825in}{1.926202in}}%
\pgfpathlineto{\pgfqpoint{0.599443in}{1.926202in}}%
\pgfpathlineto{\pgfqpoint{0.599443in}{1.926202in}}%
\pgfpathlineto{\pgfqpoint{0.689061in}{1.926202in}}%
\pgfpathlineto{\pgfqpoint{0.689061in}{1.926202in}}%
\pgfpathlineto{\pgfqpoint{0.778680in}{1.926202in}}%
\pgfpathlineto{\pgfqpoint{0.778680in}{1.926202in}}%
\pgfpathlineto{\pgfqpoint{0.868298in}{1.926202in}}%
\pgfpathlineto{\pgfqpoint{0.868298in}{1.926202in}}%
\pgfpathlineto{\pgfqpoint{0.957916in}{1.926202in}}%
\pgfpathlineto{\pgfqpoint{0.957916in}{1.926202in}}%
\pgfpathlineto{\pgfqpoint{1.047535in}{1.926202in}}%
\pgfpathlineto{\pgfqpoint{1.047535in}{1.926202in}}%
\pgfpathlineto{\pgfqpoint{1.137153in}{1.926202in}}%
\pgfpathlineto{\pgfqpoint{1.137153in}{1.926202in}}%
\pgfpathlineto{\pgfqpoint{1.226771in}{1.926202in}}%
\pgfpathlineto{\pgfqpoint{1.226771in}{1.926202in}}%
\pgfpathlineto{\pgfqpoint{1.316390in}{1.926202in}}%
\pgfpathlineto{\pgfqpoint{1.316390in}{1.926202in}}%
\pgfpathlineto{\pgfqpoint{1.406008in}{1.926202in}}%
\pgfpathlineto{\pgfqpoint{1.406008in}{1.926202in}}%
\pgfpathlineto{\pgfqpoint{1.495626in}{1.926202in}}%
\pgfpathlineto{\pgfqpoint{1.495626in}{1.926202in}}%
\pgfpathlineto{\pgfqpoint{1.585245in}{1.926202in}}%
\pgfpathlineto{\pgfqpoint{1.585245in}{1.926202in}}%
\pgfpathlineto{\pgfqpoint{1.674863in}{1.926202in}}%
\pgfpathlineto{\pgfqpoint{1.674863in}{1.926202in}}%
\pgfpathlineto{\pgfqpoint{1.764481in}{1.926202in}}%
\pgfpathlineto{\pgfqpoint{1.764481in}{1.926202in}}%
\pgfpathlineto{\pgfqpoint{1.854100in}{1.926202in}}%
\pgfpathlineto{\pgfqpoint{1.854100in}{1.926202in}}%
\pgfpathlineto{\pgfqpoint{1.943718in}{1.926202in}}%
\pgfpathlineto{\pgfqpoint{1.943718in}{1.926202in}}%
\pgfpathlineto{\pgfqpoint{2.033336in}{1.926202in}}%
\pgfpathlineto{\pgfqpoint{2.033336in}{1.926202in}}%
\pgfpathlineto{\pgfqpoint{2.122955in}{1.926202in}}%
\pgfpathlineto{\pgfqpoint{2.122955in}{1.926202in}}%
\pgfpathlineto{\pgfqpoint{2.212573in}{1.926202in}}%
\pgfpathlineto{\pgfqpoint{2.212573in}{1.926202in}}%
\pgfpathlineto{\pgfqpoint{2.302191in}{1.926202in}}%
\pgfpathlineto{\pgfqpoint{2.302191in}{1.926202in}}%
\pgfpathlineto{\pgfqpoint{2.391810in}{1.926202in}}%
\pgfpathlineto{\pgfqpoint{2.391810in}{1.926202in}}%
\pgfpathlineto{\pgfqpoint{2.481428in}{1.926202in}}%
\pgfpathlineto{\pgfqpoint{2.481428in}{1.926202in}}%
\pgfpathlineto{\pgfqpoint{2.571046in}{1.926202in}}%
\pgfpathlineto{\pgfqpoint{2.571046in}{1.926202in}}%
\pgfpathlineto{\pgfqpoint{2.660665in}{1.926202in}}%
\pgfpathlineto{\pgfqpoint{2.660665in}{1.926202in}}%
\pgfpathlineto{\pgfqpoint{2.750283in}{1.926202in}}%
\pgfpathlineto{\pgfqpoint{2.750283in}{1.926202in}}%
\pgfpathlineto{\pgfqpoint{2.839901in}{1.926202in}}%
\pgfpathlineto{\pgfqpoint{2.839901in}{1.926202in}}%
\pgfpathlineto{\pgfqpoint{2.929520in}{1.926202in}}%
\pgfpathlineto{\pgfqpoint{2.929520in}{1.926202in}}%
\pgfpathlineto{\pgfqpoint{3.019138in}{1.926202in}}%
\pgfpathlineto{\pgfqpoint{3.019138in}{1.926202in}}%
\pgfpathlineto{\pgfqpoint{3.108756in}{1.926202in}}%
\pgfpathlineto{\pgfqpoint{3.108756in}{1.926202in}}%
\pgfpathlineto{\pgfqpoint{3.198375in}{1.926202in}}%
\pgfpathlineto{\pgfqpoint{3.198375in}{0.553219in}}%
\pgfusepath{stroke}%
\end{pgfscope}%
\begin{pgfscope}%
\pgfsetrectcap%
\pgfsetmiterjoin%
\pgfsetlinewidth{0.803000pt}%
\definecolor{currentstroke}{rgb}{0.000000,0.000000,0.000000}%
\pgfsetstrokecolor{currentstroke}%
\pgfsetdash{}{0pt}%
\pgfpathmoveto{\pgfqpoint{0.375397in}{0.484570in}}%
\pgfpathlineto{\pgfqpoint{0.375397in}{1.994851in}}%
\pgfusepath{stroke}%
\end{pgfscope}%
\begin{pgfscope}%
\pgfsetrectcap%
\pgfsetmiterjoin%
\pgfsetlinewidth{0.803000pt}%
\definecolor{currentstroke}{rgb}{0.000000,0.000000,0.000000}%
\pgfsetstrokecolor{currentstroke}%
\pgfsetdash{}{0pt}%
\pgfpathmoveto{\pgfqpoint{3.332802in}{0.484570in}}%
\pgfpathlineto{\pgfqpoint{3.332802in}{1.994851in}}%
\pgfusepath{stroke}%
\end{pgfscope}%
\begin{pgfscope}%
\pgfsetrectcap%
\pgfsetmiterjoin%
\pgfsetlinewidth{0.803000pt}%
\definecolor{currentstroke}{rgb}{0.000000,0.000000,0.000000}%
\pgfsetstrokecolor{currentstroke}%
\pgfsetdash{}{0pt}%
\pgfpathmoveto{\pgfqpoint{0.375397in}{0.484570in}}%
\pgfpathlineto{\pgfqpoint{3.332802in}{0.484570in}}%
\pgfusepath{stroke}%
\end{pgfscope}%
\begin{pgfscope}%
\pgfsetrectcap%
\pgfsetmiterjoin%
\pgfsetlinewidth{0.803000pt}%
\definecolor{currentstroke}{rgb}{0.000000,0.000000,0.000000}%
\pgfsetstrokecolor{currentstroke}%
\pgfsetdash{}{0pt}%
\pgfpathmoveto{\pgfqpoint{0.375397in}{1.994851in}}%
\pgfpathlineto{\pgfqpoint{3.332802in}{1.994851in}}%
\pgfusepath{stroke}%
\end{pgfscope}%
\begin{pgfscope}%
\pgfsetbuttcap%
\pgfsetmiterjoin%
\definecolor{currentfill}{rgb}{1.000000,1.000000,1.000000}%
\pgfsetfillcolor{currentfill}%
\pgfsetfillopacity{0.800000}%
\pgfsetlinewidth{1.003750pt}%
\definecolor{currentstroke}{rgb}{0.800000,0.800000,0.800000}%
\pgfsetstrokecolor{currentstroke}%
\pgfsetstrokeopacity{0.800000}%
\pgfsetdash{}{0pt}%
\pgfpathmoveto{\pgfqpoint{1.368667in}{1.021778in}}%
\pgfpathlineto{\pgfqpoint{2.339532in}{1.021778in}}%
\pgfpathquadraticcurveto{\pgfqpoint{2.358977in}{1.021778in}}{\pgfqpoint{2.358977in}{1.041222in}}%
\pgfpathlineto{\pgfqpoint{2.358977in}{1.438199in}}%
\pgfpathquadraticcurveto{\pgfqpoint{2.358977in}{1.457643in}}{\pgfqpoint{2.339532in}{1.457643in}}%
\pgfpathlineto{\pgfqpoint{1.368667in}{1.457643in}}%
\pgfpathquadraticcurveto{\pgfqpoint{1.349223in}{1.457643in}}{\pgfqpoint{1.349223in}{1.438199in}}%
\pgfpathlineto{\pgfqpoint{1.349223in}{1.041222in}}%
\pgfpathquadraticcurveto{\pgfqpoint{1.349223in}{1.021778in}}{\pgfqpoint{1.368667in}{1.021778in}}%
\pgfpathclose%
\pgfusepath{stroke,fill}%
\end{pgfscope}%
\begin{pgfscope}%
\pgfsetrectcap%
\pgfsetroundjoin%
\pgfsetlinewidth{1.505625pt}%
\definecolor{currentstroke}{rgb}{0.121569,0.466667,0.705882}%
\pgfsetstrokecolor{currentstroke}%
\pgfsetdash{}{0pt}%
\pgfpathmoveto{\pgfqpoint{1.388111in}{1.384727in}}%
\pgfpathlineto{\pgfqpoint{1.582556in}{1.384727in}}%
\pgfusepath{stroke}%
\end{pgfscope}%
\begin{pgfscope}%
\pgftext[x=1.660334in,y=1.350699in,left,base]{\rmfamily\fontsize{7.000000}{8.400000}\selectfont image}%
\end{pgfscope}%
\begin{pgfscope}%
\pgfsetrectcap%
\pgfsetroundjoin%
\pgfsetlinewidth{1.505625pt}%
\definecolor{currentstroke}{rgb}{1.000000,0.498039,0.054902}%
\pgfsetstrokecolor{currentstroke}%
\pgfsetdash{}{0pt}%
\pgfpathmoveto{\pgfqpoint{1.388111in}{1.249160in}}%
\pgfpathlineto{\pgfqpoint{1.582556in}{1.249160in}}%
\pgfusepath{stroke}%
\end{pgfscope}%
\begin{pgfscope}%
\pgftext[x=1.660334in,y=1.215133in,left,base]{\rmfamily\fontsize{7.000000}{8.400000}\selectfont text}%
\end{pgfscope}%
\begin{pgfscope}%
\pgfsetrectcap%
\pgfsetroundjoin%
\pgfsetlinewidth{1.505625pt}%
\definecolor{currentstroke}{rgb}{0.172549,0.627451,0.172549}%
\pgfsetstrokecolor{currentstroke}%
\pgfsetdash{}{0pt}%
\pgfpathmoveto{\pgfqpoint{1.388111in}{1.113594in}}%
\pgfpathlineto{\pgfqpoint{1.582556in}{1.113594in}}%
\pgfusepath{stroke}%
\end{pgfscope}%
\begin{pgfscope}%
\pgftext[x=1.660334in,y=1.079566in,left,base]{\rmfamily\fontsize{7.000000}{8.400000}\selectfont reinforcement}%
\end{pgfscope}%
\end{pgfpicture}%
\makeatother%
\endgroup%

%% file: figures/paper/stat_reduction_aggreg_all.pgf
\begingroup%
\makeatletter%
\begin{pgfpicture}%
\pgfpathrectangle{\pgfpointorigin}{\pgfqpoint{3.502802in}{2.164851in}}%
\pgfusepath{use as bounding box, clip}%
\begin{pgfscope}%
\pgfsetbuttcap%
\pgfsetmiterjoin%
\pgfsetlinewidth{0.000000pt}%
\definecolor{currentstroke}{rgb}{1.000000,1.000000,1.000000}%
\pgfsetstrokecolor{currentstroke}%
\pgfsetstrokeopacity{0.000000}%
\pgfsetdash{}{0pt}%
\pgfpathmoveto{\pgfqpoint{0.000000in}{0.000000in}}%
\pgfpathlineto{\pgfqpoint{3.502802in}{0.000000in}}%
\pgfpathlineto{\pgfqpoint{3.502802in}{2.164851in}}%
\pgfpathlineto{\pgfqpoint{0.000000in}{2.164851in}}%
\pgfpathclose%
\pgfusepath{}%
\end{pgfscope}%
\begin{pgfscope}%
\pgfsetbuttcap%
\pgfsetmiterjoin%
\definecolor{currentfill}{rgb}{1.000000,1.000000,1.000000}%
\pgfsetfillcolor{currentfill}%
\pgfsetlinewidth{0.000000pt}%
\definecolor{currentstroke}{rgb}{0.000000,0.000000,0.000000}%
\pgfsetstrokecolor{currentstroke}%
\pgfsetstrokeopacity{0.000000}%
\pgfsetdash{}{0pt}%
\pgfpathmoveto{\pgfqpoint{0.437850in}{0.238134in}}%
\pgfpathlineto{\pgfqpoint{3.152522in}{0.238134in}}%
\pgfpathlineto{\pgfqpoint{3.152522in}{1.905069in}}%
\pgfpathlineto{\pgfqpoint{0.437850in}{1.905069in}}%
\pgfpathclose%
\pgfusepath{fill}%
\end{pgfscope}%
\begin{pgfscope}%
\pgfpathrectangle{\pgfqpoint{0.437850in}{0.238134in}}{\pgfqpoint{2.714672in}{1.666935in}} %
\pgfusepath{clip}%
\pgfsetbuttcap%
\pgfsetmiterjoin%
\definecolor{currentfill}{rgb}{0.194608,0.453431,0.632843}%
\pgfsetfillcolor{currentfill}%
\pgfsetlinewidth{1.505625pt}%
\definecolor{currentstroke}{rgb}{0.239216,0.239216,0.239216}%
\pgfsetstrokecolor{currentstroke}%
\pgfsetdash{}{0pt}%
\pgfpathmoveto{\pgfqpoint{0.528339in}{0.464309in}}%
\pgfpathlineto{\pgfqpoint{1.252252in}{0.464309in}}%
\pgfpathlineto{\pgfqpoint{1.252252in}{0.839969in}}%
\pgfpathlineto{\pgfqpoint{0.528339in}{0.839969in}}%
\pgfpathlineto{\pgfqpoint{0.528339in}{0.464309in}}%
\pgfpathclose%
\pgfusepath{stroke,fill}%
\end{pgfscope}%
\begin{pgfscope}%
\pgfpathrectangle{\pgfqpoint{0.437850in}{0.238134in}}{\pgfqpoint{2.714672in}{1.666935in}} %
\pgfusepath{clip}%
\pgfsetbuttcap%
\pgfsetmiterjoin%
\definecolor{currentfill}{rgb}{0.881863,0.505392,0.173039}%
\pgfsetfillcolor{currentfill}%
\pgfsetlinewidth{1.505625pt}%
\definecolor{currentstroke}{rgb}{0.239216,0.239216,0.239216}%
\pgfsetstrokecolor{currentstroke}%
\pgfsetdash{}{0pt}%
\pgfpathmoveto{\pgfqpoint{1.433230in}{0.663310in}}%
\pgfpathlineto{\pgfqpoint{2.157142in}{0.663310in}}%
\pgfpathlineto{\pgfqpoint{2.157142in}{0.964852in}}%
\pgfpathlineto{\pgfqpoint{1.433230in}{0.964852in}}%
\pgfpathlineto{\pgfqpoint{1.433230in}{0.663310in}}%
\pgfpathclose%
\pgfusepath{stroke,fill}%
\end{pgfscope}%
\begin{pgfscope}%
\pgfpathrectangle{\pgfqpoint{0.437850in}{0.238134in}}{\pgfqpoint{2.714672in}{1.666935in}} %
\pgfusepath{clip}%
\pgfsetbuttcap%
\pgfsetmiterjoin%
\definecolor{currentfill}{rgb}{0.229412,0.570588,0.229412}%
\pgfsetfillcolor{currentfill}%
\pgfsetlinewidth{1.505625pt}%
\definecolor{currentstroke}{rgb}{0.239216,0.239216,0.239216}%
\pgfsetstrokecolor{currentstroke}%
\pgfsetdash{}{0pt}%
\pgfpathmoveto{\pgfqpoint{2.338120in}{0.313903in}}%
\pgfpathlineto{\pgfqpoint{3.062033in}{0.313903in}}%
\pgfpathlineto{\pgfqpoint{3.062033in}{0.868160in}}%
\pgfpathlineto{\pgfqpoint{2.338120in}{0.868160in}}%
\pgfpathlineto{\pgfqpoint{2.338120in}{0.313903in}}%
\pgfpathclose%
\pgfusepath{stroke,fill}%
\end{pgfscope}%
\begin{pgfscope}%
\pgfsetbuttcap%
\pgfsetroundjoin%
\definecolor{currentfill}{rgb}{0.000000,0.000000,0.000000}%
\pgfsetfillcolor{currentfill}%
\pgfsetlinewidth{0.803000pt}%
\definecolor{currentstroke}{rgb}{0.000000,0.000000,0.000000}%
\pgfsetstrokecolor{currentstroke}%
\pgfsetdash{}{0pt}%
\pgfsys@defobject{currentmarker}{\pgfqpoint{0.000000in}{-0.048611in}}{\pgfqpoint{0.000000in}{0.000000in}}{%
\pgfpathmoveto{\pgfqpoint{0.000000in}{0.000000in}}%
\pgfpathlineto{\pgfqpoint{0.000000in}{-0.048611in}}%
\pgfusepath{stroke,fill}%
}%
\begin{pgfscope}%
\pgfsys@transformshift{0.890296in}{0.238134in}%
\pgfsys@useobject{currentmarker}{}%
\end{pgfscope}%
\end{pgfscope}%
\begin{pgfscope}%
\pgftext[x=0.890296in,y=0.140911in,,top]{\rmfamily\fontsize{7.000000}{8.400000}\selectfont image}%
\end{pgfscope}%
\begin{pgfscope}%
\pgfsetbuttcap%
\pgfsetroundjoin%
\definecolor{currentfill}{rgb}{0.000000,0.000000,0.000000}%
\pgfsetfillcolor{currentfill}%
\pgfsetlinewidth{0.803000pt}%
\definecolor{currentstroke}{rgb}{0.000000,0.000000,0.000000}%
\pgfsetstrokecolor{currentstroke}%
\pgfsetdash{}{0pt}%
\pgfsys@defobject{currentmarker}{\pgfqpoint{0.000000in}{-0.048611in}}{\pgfqpoint{0.000000in}{0.000000in}}{%
\pgfpathmoveto{\pgfqpoint{0.000000in}{0.000000in}}%
\pgfpathlineto{\pgfqpoint{0.000000in}{-0.048611in}}%
\pgfusepath{stroke,fill}%
}%
\begin{pgfscope}%
\pgfsys@transformshift{1.795186in}{0.238134in}%
\pgfsys@useobject{currentmarker}{}%
\end{pgfscope}%
\end{pgfscope}%
\begin{pgfscope}%
\pgftext[x=1.795186in,y=0.140911in,,top]{\rmfamily\fontsize{7.000000}{8.400000}\selectfont text}%
\end{pgfscope}%
\begin{pgfscope}%
\pgfsetbuttcap%
\pgfsetroundjoin%
\definecolor{currentfill}{rgb}{0.000000,0.000000,0.000000}%
\pgfsetfillcolor{currentfill}%
\pgfsetlinewidth{0.803000pt}%
\definecolor{currentstroke}{rgb}{0.000000,0.000000,0.000000}%
\pgfsetstrokecolor{currentstroke}%
\pgfsetdash{}{0pt}%
\pgfsys@defobject{currentmarker}{\pgfqpoint{0.000000in}{-0.048611in}}{\pgfqpoint{0.000000in}{0.000000in}}{%
\pgfpathmoveto{\pgfqpoint{0.000000in}{0.000000in}}%
\pgfpathlineto{\pgfqpoint{0.000000in}{-0.048611in}}%
\pgfusepath{stroke,fill}%
}%
\begin{pgfscope}%
\pgfsys@transformshift{2.700077in}{0.238134in}%
\pgfsys@useobject{currentmarker}{}%
\end{pgfscope}%
\end{pgfscope}%
\begin{pgfscope}%
\pgftext[x=2.700077in,y=0.140911in,,top]{\rmfamily\fontsize{7.000000}{8.400000}\selectfont reinforcement}%
\end{pgfscope}%
\begin{pgfscope}%
\pgfsetbuttcap%
\pgfsetroundjoin%
\definecolor{currentfill}{rgb}{0.000000,0.000000,0.000000}%
\pgfsetfillcolor{currentfill}%
\pgfsetlinewidth{0.803000pt}%
\definecolor{currentstroke}{rgb}{0.000000,0.000000,0.000000}%
\pgfsetstrokecolor{currentstroke}%
\pgfsetdash{}{0pt}%
\pgfsys@defobject{currentmarker}{\pgfqpoint{-0.048611in}{0.000000in}}{\pgfqpoint{0.000000in}{0.000000in}}{%
\pgfpathmoveto{\pgfqpoint{0.000000in}{0.000000in}}%
\pgfpathlineto{\pgfqpoint{-0.048611in}{0.000000in}}%
\pgfusepath{stroke,fill}%
}%
\begin{pgfscope}%
\pgfsys@transformshift{0.437850in}{0.313903in}%
\pgfsys@useobject{currentmarker}{}%
\end{pgfscope}%
\end{pgfscope}%
\begin{pgfscope}%
\pgftext[x=0.196916in,y=0.280424in,left,base]{\rmfamily\fontsize{7.000000}{8.400000}\selectfont \(\displaystyle 0.0\)}%
\end{pgfscope}%
\begin{pgfscope}%
\pgfsetbuttcap%
\pgfsetroundjoin%
\definecolor{currentfill}{rgb}{0.000000,0.000000,0.000000}%
\pgfsetfillcolor{currentfill}%
\pgfsetlinewidth{0.803000pt}%
\definecolor{currentstroke}{rgb}{0.000000,0.000000,0.000000}%
\pgfsetstrokecolor{currentstroke}%
\pgfsetdash{}{0pt}%
\pgfsys@defobject{currentmarker}{\pgfqpoint{-0.048611in}{0.000000in}}{\pgfqpoint{0.000000in}{0.000000in}}{%
\pgfpathmoveto{\pgfqpoint{0.000000in}{0.000000in}}%
\pgfpathlineto{\pgfqpoint{-0.048611in}{0.000000in}}%
\pgfusepath{stroke,fill}%
}%
\begin{pgfscope}%
\pgfsys@transformshift{0.437850in}{0.648335in}%
\pgfsys@useobject{currentmarker}{}%
\end{pgfscope}%
\end{pgfscope}%
\begin{pgfscope}%
\pgftext[x=0.196916in,y=0.614856in,left,base]{\rmfamily\fontsize{7.000000}{8.400000}\selectfont \(\displaystyle 0.2\)}%
\end{pgfscope}%
\begin{pgfscope}%
\pgfsetbuttcap%
\pgfsetroundjoin%
\definecolor{currentfill}{rgb}{0.000000,0.000000,0.000000}%
\pgfsetfillcolor{currentfill}%
\pgfsetlinewidth{0.803000pt}%
\definecolor{currentstroke}{rgb}{0.000000,0.000000,0.000000}%
\pgfsetstrokecolor{currentstroke}%
\pgfsetdash{}{0pt}%
\pgfsys@defobject{currentmarker}{\pgfqpoint{-0.048611in}{0.000000in}}{\pgfqpoint{0.000000in}{0.000000in}}{%
\pgfpathmoveto{\pgfqpoint{0.000000in}{0.000000in}}%
\pgfpathlineto{\pgfqpoint{-0.048611in}{0.000000in}}%
\pgfusepath{stroke,fill}%
}%
\begin{pgfscope}%
\pgfsys@transformshift{0.437850in}{0.982768in}%
\pgfsys@useobject{currentmarker}{}%
\end{pgfscope}%
\end{pgfscope}%
\begin{pgfscope}%
\pgftext[x=0.196916in,y=0.949288in,left,base]{\rmfamily\fontsize{7.000000}{8.400000}\selectfont \(\displaystyle 0.4\)}%
\end{pgfscope}%
\begin{pgfscope}%
\pgfsetbuttcap%
\pgfsetroundjoin%
\definecolor{currentfill}{rgb}{0.000000,0.000000,0.000000}%
\pgfsetfillcolor{currentfill}%
\pgfsetlinewidth{0.803000pt}%
\definecolor{currentstroke}{rgb}{0.000000,0.000000,0.000000}%
\pgfsetstrokecolor{currentstroke}%
\pgfsetdash{}{0pt}%
\pgfsys@defobject{currentmarker}{\pgfqpoint{-0.048611in}{0.000000in}}{\pgfqpoint{0.000000in}{0.000000in}}{%
\pgfpathmoveto{\pgfqpoint{0.000000in}{0.000000in}}%
\pgfpathlineto{\pgfqpoint{-0.048611in}{0.000000in}}%
\pgfusepath{stroke,fill}%
}%
\begin{pgfscope}%
\pgfsys@transformshift{0.437850in}{1.317200in}%
\pgfsys@useobject{currentmarker}{}%
\end{pgfscope}%
\end{pgfscope}%
\begin{pgfscope}%
\pgftext[x=0.196916in,y=1.283720in,left,base]{\rmfamily\fontsize{7.000000}{8.400000}\selectfont \(\displaystyle 0.6\)}%
\end{pgfscope}%
\begin{pgfscope}%
\pgfsetbuttcap%
\pgfsetroundjoin%
\definecolor{currentfill}{rgb}{0.000000,0.000000,0.000000}%
\pgfsetfillcolor{currentfill}%
\pgfsetlinewidth{0.803000pt}%
\definecolor{currentstroke}{rgb}{0.000000,0.000000,0.000000}%
\pgfsetstrokecolor{currentstroke}%
\pgfsetdash{}{0pt}%
\pgfsys@defobject{currentmarker}{\pgfqpoint{-0.048611in}{0.000000in}}{\pgfqpoint{0.000000in}{0.000000in}}{%
\pgfpathmoveto{\pgfqpoint{0.000000in}{0.000000in}}%
\pgfpathlineto{\pgfqpoint{-0.048611in}{0.000000in}}%
\pgfusepath{stroke,fill}%
}%
\begin{pgfscope}%
\pgfsys@transformshift{0.437850in}{1.651632in}%
\pgfsys@useobject{currentmarker}{}%
\end{pgfscope}%
\end{pgfscope}%
\begin{pgfscope}%
\pgftext[x=0.196916in,y=1.618152in,left,base]{\rmfamily\fontsize{7.000000}{8.400000}\selectfont \(\displaystyle 0.8\)}%
\end{pgfscope}%
\begin{pgfscope}%
\pgfpathrectangle{\pgfqpoint{0.437850in}{0.238134in}}{\pgfqpoint{2.714672in}{1.666935in}} %
\pgfusepath{clip}%
\pgfsetrectcap%
\pgfsetroundjoin%
\pgfsetlinewidth{1.505625pt}%
\definecolor{currentstroke}{rgb}{0.239216,0.239216,0.239216}%
\pgfsetstrokecolor{currentstroke}%
\pgfsetdash{}{0pt}%
\pgfpathmoveto{\pgfqpoint{0.890296in}{0.464309in}}%
\pgfpathlineto{\pgfqpoint{0.890296in}{0.313903in}}%
\pgfusepath{stroke}%
\end{pgfscope}%
\begin{pgfscope}%
\pgfpathrectangle{\pgfqpoint{0.437850in}{0.238134in}}{\pgfqpoint{2.714672in}{1.666935in}} %
\pgfusepath{clip}%
\pgfsetrectcap%
\pgfsetroundjoin%
\pgfsetlinewidth{1.505625pt}%
\definecolor{currentstroke}{rgb}{0.239216,0.239216,0.239216}%
\pgfsetstrokecolor{currentstroke}%
\pgfsetdash{}{0pt}%
\pgfpathmoveto{\pgfqpoint{0.890296in}{0.839969in}}%
\pgfpathlineto{\pgfqpoint{0.890296in}{1.362546in}}%
\pgfusepath{stroke}%
\end{pgfscope}%
\begin{pgfscope}%
\pgfpathrectangle{\pgfqpoint{0.437850in}{0.238134in}}{\pgfqpoint{2.714672in}{1.666935in}} %
\pgfusepath{clip}%
\pgfsetrectcap%
\pgfsetroundjoin%
\pgfsetlinewidth{1.505625pt}%
\definecolor{currentstroke}{rgb}{0.239216,0.239216,0.239216}%
\pgfsetstrokecolor{currentstroke}%
\pgfsetdash{}{0pt}%
\pgfpathmoveto{\pgfqpoint{0.709317in}{0.313903in}}%
\pgfpathlineto{\pgfqpoint{1.071274in}{0.313903in}}%
\pgfusepath{stroke}%
\end{pgfscope}%
\begin{pgfscope}%
\pgfpathrectangle{\pgfqpoint{0.437850in}{0.238134in}}{\pgfqpoint{2.714672in}{1.666935in}} %
\pgfusepath{clip}%
\pgfsetrectcap%
\pgfsetroundjoin%
\pgfsetlinewidth{1.505625pt}%
\definecolor{currentstroke}{rgb}{0.239216,0.239216,0.239216}%
\pgfsetstrokecolor{currentstroke}%
\pgfsetdash{}{0pt}%
\pgfpathmoveto{\pgfqpoint{0.709317in}{1.362546in}}%
\pgfpathlineto{\pgfqpoint{1.071274in}{1.362546in}}%
\pgfusepath{stroke}%
\end{pgfscope}%
\begin{pgfscope}%
\pgfpathrectangle{\pgfqpoint{0.437850in}{0.238134in}}{\pgfqpoint{2.714672in}{1.666935in}} %
\pgfusepath{clip}%
\pgfsetrectcap%
\pgfsetroundjoin%
\pgfsetlinewidth{1.505625pt}%
\definecolor{currentstroke}{rgb}{0.239216,0.239216,0.239216}%
\pgfsetstrokecolor{currentstroke}%
\pgfsetdash{}{0pt}%
\pgfpathmoveto{\pgfqpoint{1.795186in}{0.663310in}}%
\pgfpathlineto{\pgfqpoint{1.795186in}{0.313903in}}%
\pgfusepath{stroke}%
\end{pgfscope}%
\begin{pgfscope}%
\pgfpathrectangle{\pgfqpoint{0.437850in}{0.238134in}}{\pgfqpoint{2.714672in}{1.666935in}} %
\pgfusepath{clip}%
\pgfsetrectcap%
\pgfsetroundjoin%
\pgfsetlinewidth{1.505625pt}%
\definecolor{currentstroke}{rgb}{0.239216,0.239216,0.239216}%
\pgfsetstrokecolor{currentstroke}%
\pgfsetdash{}{0pt}%
\pgfpathmoveto{\pgfqpoint{1.795186in}{0.964852in}}%
\pgfpathlineto{\pgfqpoint{1.795186in}{1.399722in}}%
\pgfusepath{stroke}%
\end{pgfscope}%
\begin{pgfscope}%
\pgfpathrectangle{\pgfqpoint{0.437850in}{0.238134in}}{\pgfqpoint{2.714672in}{1.666935in}} %
\pgfusepath{clip}%
\pgfsetrectcap%
\pgfsetroundjoin%
\pgfsetlinewidth{1.505625pt}%
\definecolor{currentstroke}{rgb}{0.239216,0.239216,0.239216}%
\pgfsetstrokecolor{currentstroke}%
\pgfsetdash{}{0pt}%
\pgfpathmoveto{\pgfqpoint{1.614208in}{0.313903in}}%
\pgfpathlineto{\pgfqpoint{1.976164in}{0.313903in}}%
\pgfusepath{stroke}%
\end{pgfscope}%
\begin{pgfscope}%
\pgfpathrectangle{\pgfqpoint{0.437850in}{0.238134in}}{\pgfqpoint{2.714672in}{1.666935in}} %
\pgfusepath{clip}%
\pgfsetrectcap%
\pgfsetroundjoin%
\pgfsetlinewidth{1.505625pt}%
\definecolor{currentstroke}{rgb}{0.239216,0.239216,0.239216}%
\pgfsetstrokecolor{currentstroke}%
\pgfsetdash{}{0pt}%
\pgfpathmoveto{\pgfqpoint{1.614208in}{1.399722in}}%
\pgfpathlineto{\pgfqpoint{1.976164in}{1.399722in}}%
\pgfusepath{stroke}%
\end{pgfscope}%
\begin{pgfscope}%
\pgfpathrectangle{\pgfqpoint{0.437850in}{0.238134in}}{\pgfqpoint{2.714672in}{1.666935in}} %
\pgfusepath{clip}%
\pgfsetbuttcap%
\pgfsetmiterjoin%
\definecolor{currentfill}{rgb}{0.239216,0.239216,0.239216}%
\pgfsetfillcolor{currentfill}%
\pgfsetlinewidth{1.003750pt}%
\definecolor{currentstroke}{rgb}{0.239216,0.239216,0.239216}%
\pgfsetstrokecolor{currentstroke}%
\pgfsetdash{}{0pt}%
\pgfsys@defobject{currentmarker}{\pgfqpoint{-0.029463in}{-0.049105in}}{\pgfqpoint{0.029463in}{0.049105in}}{%
\pgfpathmoveto{\pgfqpoint{-0.000000in}{-0.049105in}}%
\pgfpathlineto{\pgfqpoint{0.029463in}{0.000000in}}%
\pgfpathlineto{\pgfqpoint{0.000000in}{0.049105in}}%
\pgfpathlineto{\pgfqpoint{-0.029463in}{0.000000in}}%
\pgfpathclose%
\pgfusepath{stroke,fill}%
}%
\begin{pgfscope}%
\pgfsys@transformshift{1.795186in}{1.467930in}%
\pgfsys@useobject{currentmarker}{}%
\end{pgfscope}%
\begin{pgfscope}%
\pgfsys@transformshift{1.795186in}{1.444536in}%
\pgfsys@useobject{currentmarker}{}%
\end{pgfscope}%
\begin{pgfscope}%
\pgfsys@transformshift{1.795186in}{1.418160in}%
\pgfsys@useobject{currentmarker}{}%
\end{pgfscope}%
\end{pgfscope}%
\begin{pgfscope}%
\pgfpathrectangle{\pgfqpoint{0.437850in}{0.238134in}}{\pgfqpoint{2.714672in}{1.666935in}} %
\pgfusepath{clip}%
\pgfsetrectcap%
\pgfsetroundjoin%
\pgfsetlinewidth{1.505625pt}%
\definecolor{currentstroke}{rgb}{0.239216,0.239216,0.239216}%
\pgfsetstrokecolor{currentstroke}%
\pgfsetdash{}{0pt}%
\pgfpathmoveto{\pgfqpoint{2.700077in}{0.313903in}}%
\pgfpathlineto{\pgfqpoint{2.700077in}{0.313903in}}%
\pgfusepath{stroke}%
\end{pgfscope}%
\begin{pgfscope}%
\pgfpathrectangle{\pgfqpoint{0.437850in}{0.238134in}}{\pgfqpoint{2.714672in}{1.666935in}} %
\pgfusepath{clip}%
\pgfsetrectcap%
\pgfsetroundjoin%
\pgfsetlinewidth{1.505625pt}%
\definecolor{currentstroke}{rgb}{0.239216,0.239216,0.239216}%
\pgfsetstrokecolor{currentstroke}%
\pgfsetdash{}{0pt}%
\pgfpathmoveto{\pgfqpoint{2.700077in}{0.868160in}}%
\pgfpathlineto{\pgfqpoint{2.700077in}{1.660922in}}%
\pgfusepath{stroke}%
\end{pgfscope}%
\begin{pgfscope}%
\pgfpathrectangle{\pgfqpoint{0.437850in}{0.238134in}}{\pgfqpoint{2.714672in}{1.666935in}} %
\pgfusepath{clip}%
\pgfsetrectcap%
\pgfsetroundjoin%
\pgfsetlinewidth{1.505625pt}%
\definecolor{currentstroke}{rgb}{0.239216,0.239216,0.239216}%
\pgfsetstrokecolor{currentstroke}%
\pgfsetdash{}{0pt}%
\pgfpathmoveto{\pgfqpoint{2.519098in}{0.313903in}}%
\pgfpathlineto{\pgfqpoint{2.881055in}{0.313903in}}%
\pgfusepath{stroke}%
\end{pgfscope}%
\begin{pgfscope}%
\pgfpathrectangle{\pgfqpoint{0.437850in}{0.238134in}}{\pgfqpoint{2.714672in}{1.666935in}} %
\pgfusepath{clip}%
\pgfsetrectcap%
\pgfsetroundjoin%
\pgfsetlinewidth{1.505625pt}%
\definecolor{currentstroke}{rgb}{0.239216,0.239216,0.239216}%
\pgfsetstrokecolor{currentstroke}%
\pgfsetdash{}{0pt}%
\pgfpathmoveto{\pgfqpoint{2.519098in}{1.660922in}}%
\pgfpathlineto{\pgfqpoint{2.881055in}{1.660922in}}%
\pgfusepath{stroke}%
\end{pgfscope}%
\begin{pgfscope}%
\pgfpathrectangle{\pgfqpoint{0.437850in}{0.238134in}}{\pgfqpoint{2.714672in}{1.666935in}} %
\pgfusepath{clip}%
\pgfsetbuttcap%
\pgfsetmiterjoin%
\definecolor{currentfill}{rgb}{0.239216,0.239216,0.239216}%
\pgfsetfillcolor{currentfill}%
\pgfsetlinewidth{1.003750pt}%
\definecolor{currentstroke}{rgb}{0.239216,0.239216,0.239216}%
\pgfsetstrokecolor{currentstroke}%
\pgfsetdash{}{0pt}%
\pgfsys@defobject{currentmarker}{\pgfqpoint{-0.029463in}{-0.049105in}}{\pgfqpoint{0.029463in}{0.049105in}}{%
\pgfpathmoveto{\pgfqpoint{-0.000000in}{-0.049105in}}%
\pgfpathlineto{\pgfqpoint{0.029463in}{0.000000in}}%
\pgfpathlineto{\pgfqpoint{0.000000in}{0.049105in}}%
\pgfpathlineto{\pgfqpoint{-0.029463in}{0.000000in}}%
\pgfpathclose%
\pgfusepath{stroke,fill}%
}%
\begin{pgfscope}%
\pgfsys@transformshift{2.700077in}{1.829299in}%
\pgfsys@useobject{currentmarker}{}%
\end{pgfscope}%
\begin{pgfscope}%
\pgfsys@transformshift{2.700077in}{1.740158in}%
\pgfsys@useobject{currentmarker}{}%
\end{pgfscope}%
\end{pgfscope}%
\begin{pgfscope}%
\pgfpathrectangle{\pgfqpoint{0.437850in}{0.238134in}}{\pgfqpoint{2.714672in}{1.666935in}} %
\pgfusepath{clip}%
\pgfsetrectcap%
\pgfsetroundjoin%
\pgfsetlinewidth{1.505625pt}%
\definecolor{currentstroke}{rgb}{0.239216,0.239216,0.239216}%
\pgfsetstrokecolor{currentstroke}%
\pgfsetdash{}{0pt}%
\pgfpathmoveto{\pgfqpoint{0.528339in}{0.617548in}}%
\pgfpathlineto{\pgfqpoint{1.252252in}{0.617548in}}%
\pgfusepath{stroke}%
\end{pgfscope}%
\begin{pgfscope}%
\pgfpathrectangle{\pgfqpoint{0.437850in}{0.238134in}}{\pgfqpoint{2.714672in}{1.666935in}} %
\pgfusepath{clip}%
\pgfsetrectcap%
\pgfsetroundjoin%
\pgfsetlinewidth{1.505625pt}%
\definecolor{currentstroke}{rgb}{0.239216,0.239216,0.239216}%
\pgfsetstrokecolor{currentstroke}%
\pgfsetdash{}{0pt}%
\pgfpathmoveto{\pgfqpoint{1.433230in}{0.792003in}}%
\pgfpathlineto{\pgfqpoint{2.157142in}{0.792003in}}%
\pgfusepath{stroke}%
\end{pgfscope}%
\begin{pgfscope}%
\pgfpathrectangle{\pgfqpoint{0.437850in}{0.238134in}}{\pgfqpoint{2.714672in}{1.666935in}} %
\pgfusepath{clip}%
\pgfsetrectcap%
\pgfsetroundjoin%
\pgfsetlinewidth{1.505625pt}%
\definecolor{currentstroke}{rgb}{0.239216,0.239216,0.239216}%
\pgfsetstrokecolor{currentstroke}%
\pgfsetdash{}{0pt}%
\pgfpathmoveto{\pgfqpoint{2.338120in}{0.684098in}}%
\pgfpathlineto{\pgfqpoint{3.062033in}{0.684098in}}%
\pgfusepath{stroke}%
\end{pgfscope}%
\begin{pgfscope}%
\pgfsetrectcap%
\pgfsetmiterjoin%
\pgfsetlinewidth{0.803000pt}%
\definecolor{currentstroke}{rgb}{0.000000,0.000000,0.000000}%
\pgfsetstrokecolor{currentstroke}%
\pgfsetdash{}{0pt}%
\pgfpathmoveto{\pgfqpoint{0.437850in}{0.238134in}}%
\pgfpathlineto{\pgfqpoint{0.437850in}{1.905069in}}%
\pgfusepath{stroke}%
\end{pgfscope}%
\begin{pgfscope}%
\pgfsetrectcap%
\pgfsetmiterjoin%
\pgfsetlinewidth{0.803000pt}%
\definecolor{currentstroke}{rgb}{0.000000,0.000000,0.000000}%
\pgfsetstrokecolor{currentstroke}%
\pgfsetdash{}{0pt}%
\pgfpathmoveto{\pgfqpoint{3.152522in}{0.238134in}}%
\pgfpathlineto{\pgfqpoint{3.152522in}{1.905069in}}%
\pgfusepath{stroke}%
\end{pgfscope}%
\begin{pgfscope}%
\pgfsetrectcap%
\pgfsetmiterjoin%
\pgfsetlinewidth{0.803000pt}%
\definecolor{currentstroke}{rgb}{0.000000,0.000000,0.000000}%
\pgfsetstrokecolor{currentstroke}%
\pgfsetdash{}{0pt}%
\pgfpathmoveto{\pgfqpoint{0.437850in}{0.238134in}}%
\pgfpathlineto{\pgfqpoint{3.152522in}{0.238134in}}%
\pgfusepath{stroke}%
\end{pgfscope}%
\begin{pgfscope}%
\pgfsetrectcap%
\pgfsetmiterjoin%
\pgfsetlinewidth{0.803000pt}%
\definecolor{currentstroke}{rgb}{0.000000,0.000000,0.000000}%
\pgfsetstrokecolor{currentstroke}%
\pgfsetdash{}{0pt}%
\pgfpathmoveto{\pgfqpoint{0.437850in}{1.905069in}}%
\pgfpathlineto{\pgfqpoint{3.152522in}{1.905069in}}%
\pgfusepath{stroke}%
\end{pgfscope}%
\end{pgfpicture}%
\makeatother%
\endgroup%

%% file: figures/image/train_test_stats.pgf
\begingroup%
\makeatletter%
\begin{pgfpicture}%
\pgfpathrectangle{\pgfpointorigin}{\pgfqpoint{2.101681in}{1.298910in}}%
\pgfusepath{use as bounding box, clip}%
\begin{pgfscope}%
\pgfsetbuttcap%
\pgfsetmiterjoin%
\pgfsetlinewidth{0.000000pt}%
\definecolor{currentstroke}{rgb}{1.000000,1.000000,1.000000}%
\pgfsetstrokecolor{currentstroke}%
\pgfsetstrokeopacity{0.000000}%
\pgfsetdash{}{0pt}%
\pgfpathmoveto{\pgfqpoint{0.000000in}{0.000000in}}%
\pgfpathlineto{\pgfqpoint{2.101681in}{0.000000in}}%
\pgfpathlineto{\pgfqpoint{2.101681in}{1.298910in}}%
\pgfpathlineto{\pgfqpoint{0.000000in}{1.298910in}}%
\pgfpathclose%
\pgfusepath{}%
\end{pgfscope}%
\begin{pgfscope}%
\pgfsetbuttcap%
\pgfsetmiterjoin%
\definecolor{currentfill}{rgb}{1.000000,1.000000,1.000000}%
\pgfsetfillcolor{currentfill}%
\pgfsetlinewidth{0.000000pt}%
\definecolor{currentstroke}{rgb}{0.000000,0.000000,0.000000}%
\pgfsetstrokecolor{currentstroke}%
\pgfsetstrokeopacity{0.000000}%
\pgfsetdash{}{0pt}%
\pgfpathmoveto{\pgfqpoint{0.375397in}{0.318319in}}%
\pgfpathlineto{\pgfqpoint{1.931681in}{0.318319in}}%
\pgfpathlineto{\pgfqpoint{1.931681in}{1.128910in}}%
\pgfpathlineto{\pgfqpoint{0.375397in}{1.128910in}}%
\pgfpathclose%
\pgfusepath{fill}%
\end{pgfscope}%
\begin{pgfscope}%
\pgfpathrectangle{\pgfqpoint{0.375397in}{0.318319in}}{\pgfqpoint{1.556284in}{0.810592in}} %
\pgfusepath{clip}%
\pgfsetbuttcap%
\pgfsetmiterjoin%
\definecolor{currentfill}{rgb}{0.194608,0.453431,0.632843}%
\pgfsetfillcolor{currentfill}%
\pgfsetlinewidth{1.505625pt}%
\definecolor{currentstroke}{rgb}{0.247059,0.247059,0.247059}%
\pgfsetstrokecolor{currentstroke}%
\pgfsetdash{}{0pt}%
\pgfpathmoveto{\pgfqpoint{0.406523in}{0.430908in}}%
\pgfpathlineto{\pgfqpoint{0.655528in}{0.430908in}}%
\pgfpathlineto{\pgfqpoint{0.655528in}{0.667378in}}%
\pgfpathlineto{\pgfqpoint{0.406523in}{0.667378in}}%
\pgfpathlineto{\pgfqpoint{0.406523in}{0.430908in}}%
\pgfpathclose%
\pgfusepath{stroke,fill}%
\end{pgfscope}%
\begin{pgfscope}%
\pgfpathrectangle{\pgfqpoint{0.375397in}{0.318319in}}{\pgfqpoint{1.556284in}{0.810592in}} %
\pgfusepath{clip}%
\pgfsetbuttcap%
\pgfsetmiterjoin%
\definecolor{currentfill}{rgb}{0.194608,0.453431,0.632843}%
\pgfsetfillcolor{currentfill}%
\pgfsetlinewidth{1.505625pt}%
\definecolor{currentstroke}{rgb}{0.247059,0.247059,0.247059}%
\pgfsetstrokecolor{currentstroke}%
\pgfsetdash{}{0pt}%
\pgfpathmoveto{\pgfqpoint{0.717780in}{0.434270in}}%
\pgfpathlineto{\pgfqpoint{0.966785in}{0.434270in}}%
\pgfpathlineto{\pgfqpoint{0.966785in}{0.920313in}}%
\pgfpathlineto{\pgfqpoint{0.717780in}{0.920313in}}%
\pgfpathlineto{\pgfqpoint{0.717780in}{0.434270in}}%
\pgfpathclose%
\pgfusepath{stroke,fill}%
\end{pgfscope}%
\begin{pgfscope}%
\pgfpathrectangle{\pgfqpoint{0.375397in}{0.318319in}}{\pgfqpoint{1.556284in}{0.810592in}} %
\pgfusepath{clip}%
\pgfsetbuttcap%
\pgfsetmiterjoin%
\definecolor{currentfill}{rgb}{0.194608,0.453431,0.632843}%
\pgfsetfillcolor{currentfill}%
\pgfsetlinewidth{1.505625pt}%
\definecolor{currentstroke}{rgb}{0.247059,0.247059,0.247059}%
\pgfsetstrokecolor{currentstroke}%
\pgfsetdash{}{0pt}%
\pgfpathmoveto{\pgfqpoint{1.029037in}{0.400958in}}%
\pgfpathlineto{\pgfqpoint{1.278042in}{0.400958in}}%
\pgfpathlineto{\pgfqpoint{1.278042in}{0.520633in}}%
\pgfpathlineto{\pgfqpoint{1.029037in}{0.520633in}}%
\pgfpathlineto{\pgfqpoint{1.029037in}{0.400958in}}%
\pgfpathclose%
\pgfusepath{stroke,fill}%
\end{pgfscope}%
\begin{pgfscope}%
\pgfpathrectangle{\pgfqpoint{0.375397in}{0.318319in}}{\pgfqpoint{1.556284in}{0.810592in}} %
\pgfusepath{clip}%
\pgfsetbuttcap%
\pgfsetmiterjoin%
\definecolor{currentfill}{rgb}{0.194608,0.453431,0.632843}%
\pgfsetfillcolor{currentfill}%
\pgfsetlinewidth{1.505625pt}%
\definecolor{currentstroke}{rgb}{0.247059,0.247059,0.247059}%
\pgfsetstrokecolor{currentstroke}%
\pgfsetdash{}{0pt}%
\pgfpathmoveto{\pgfqpoint{1.340293in}{0.417331in}}%
\pgfpathlineto{\pgfqpoint{1.589299in}{0.417331in}}%
\pgfpathlineto{\pgfqpoint{1.589299in}{0.529110in}}%
\pgfpathlineto{\pgfqpoint{1.340293in}{0.529110in}}%
\pgfpathlineto{\pgfqpoint{1.340293in}{0.417331in}}%
\pgfpathclose%
\pgfusepath{stroke,fill}%
\end{pgfscope}%
\begin{pgfscope}%
\pgfpathrectangle{\pgfqpoint{0.375397in}{0.318319in}}{\pgfqpoint{1.556284in}{0.810592in}} %
\pgfusepath{clip}%
\pgfsetbuttcap%
\pgfsetmiterjoin%
\definecolor{currentfill}{rgb}{0.194608,0.453431,0.632843}%
\pgfsetfillcolor{currentfill}%
\pgfsetlinewidth{1.505625pt}%
\definecolor{currentstroke}{rgb}{0.247059,0.247059,0.247059}%
\pgfsetstrokecolor{currentstroke}%
\pgfsetdash{}{0pt}%
\pgfpathmoveto{\pgfqpoint{1.651550in}{0.429467in}}%
\pgfpathlineto{\pgfqpoint{1.900556in}{0.429467in}}%
\pgfpathlineto{\pgfqpoint{1.900556in}{0.721052in}}%
\pgfpathlineto{\pgfqpoint{1.651550in}{0.721052in}}%
\pgfpathlineto{\pgfqpoint{1.651550in}{0.429467in}}%
\pgfpathclose%
\pgfusepath{stroke,fill}%
\end{pgfscope}%
\begin{pgfscope}%
\pgfsetbuttcap%
\pgfsetroundjoin%
\definecolor{currentfill}{rgb}{0.000000,0.000000,0.000000}%
\pgfsetfillcolor{currentfill}%
\pgfsetlinewidth{0.803000pt}%
\definecolor{currentstroke}{rgb}{0.000000,0.000000,0.000000}%
\pgfsetstrokecolor{currentstroke}%
\pgfsetdash{}{0pt}%
\pgfsys@defobject{currentmarker}{\pgfqpoint{0.000000in}{-0.048611in}}{\pgfqpoint{0.000000in}{0.000000in}}{%
\pgfpathmoveto{\pgfqpoint{0.000000in}{0.000000in}}%
\pgfpathlineto{\pgfqpoint{0.000000in}{-0.048611in}}%
\pgfusepath{stroke,fill}%
}%
\begin{pgfscope}%
\pgfsys@transformshift{0.531026in}{0.318319in}%
\pgfsys@useobject{currentmarker}{}%
\end{pgfscope}%
\end{pgfscope}%
\begin{pgfscope}%
\pgftext[x=0.531026in,y=0.221097in,,top]{\rmfamily\fontsize{7.000000}{8.400000}\selectfont 0}%
\end{pgfscope}%
\begin{pgfscope}%
\pgfsetbuttcap%
\pgfsetroundjoin%
\definecolor{currentfill}{rgb}{0.000000,0.000000,0.000000}%
\pgfsetfillcolor{currentfill}%
\pgfsetlinewidth{0.803000pt}%
\definecolor{currentstroke}{rgb}{0.000000,0.000000,0.000000}%
\pgfsetstrokecolor{currentstroke}%
\pgfsetdash{}{0pt}%
\pgfsys@defobject{currentmarker}{\pgfqpoint{0.000000in}{-0.048611in}}{\pgfqpoint{0.000000in}{0.000000in}}{%
\pgfpathmoveto{\pgfqpoint{0.000000in}{0.000000in}}%
\pgfpathlineto{\pgfqpoint{0.000000in}{-0.048611in}}%
\pgfusepath{stroke,fill}%
}%
\begin{pgfscope}%
\pgfsys@transformshift{0.842282in}{0.318319in}%
\pgfsys@useobject{currentmarker}{}%
\end{pgfscope}%
\end{pgfscope}%
\begin{pgfscope}%
\pgftext[x=0.842282in,y=0.221097in,,top]{\rmfamily\fontsize{7.000000}{8.400000}\selectfont 1}%
\end{pgfscope}%
\begin{pgfscope}%
\pgfsetbuttcap%
\pgfsetroundjoin%
\definecolor{currentfill}{rgb}{0.000000,0.000000,0.000000}%
\pgfsetfillcolor{currentfill}%
\pgfsetlinewidth{0.803000pt}%
\definecolor{currentstroke}{rgb}{0.000000,0.000000,0.000000}%
\pgfsetstrokecolor{currentstroke}%
\pgfsetdash{}{0pt}%
\pgfsys@defobject{currentmarker}{\pgfqpoint{0.000000in}{-0.048611in}}{\pgfqpoint{0.000000in}{0.000000in}}{%
\pgfpathmoveto{\pgfqpoint{0.000000in}{0.000000in}}%
\pgfpathlineto{\pgfqpoint{0.000000in}{-0.048611in}}%
\pgfusepath{stroke,fill}%
}%
\begin{pgfscope}%
\pgfsys@transformshift{1.153539in}{0.318319in}%
\pgfsys@useobject{currentmarker}{}%
\end{pgfscope}%
\end{pgfscope}%
\begin{pgfscope}%
\pgftext[x=1.153539in,y=0.221097in,,top]{\rmfamily\fontsize{7.000000}{8.400000}\selectfont 2}%
\end{pgfscope}%
\begin{pgfscope}%
\pgfsetbuttcap%
\pgfsetroundjoin%
\definecolor{currentfill}{rgb}{0.000000,0.000000,0.000000}%
\pgfsetfillcolor{currentfill}%
\pgfsetlinewidth{0.803000pt}%
\definecolor{currentstroke}{rgb}{0.000000,0.000000,0.000000}%
\pgfsetstrokecolor{currentstroke}%
\pgfsetdash{}{0pt}%
\pgfsys@defobject{currentmarker}{\pgfqpoint{0.000000in}{-0.048611in}}{\pgfqpoint{0.000000in}{0.000000in}}{%
\pgfpathmoveto{\pgfqpoint{0.000000in}{0.000000in}}%
\pgfpathlineto{\pgfqpoint{0.000000in}{-0.048611in}}%
\pgfusepath{stroke,fill}%
}%
\begin{pgfscope}%
\pgfsys@transformshift{1.464796in}{0.318319in}%
\pgfsys@useobject{currentmarker}{}%
\end{pgfscope}%
\end{pgfscope}%
\begin{pgfscope}%
\pgftext[x=1.464796in,y=0.221097in,,top]{\rmfamily\fontsize{7.000000}{8.400000}\selectfont 3}%
\end{pgfscope}%
\begin{pgfscope}%
\pgfsetbuttcap%
\pgfsetroundjoin%
\definecolor{currentfill}{rgb}{0.000000,0.000000,0.000000}%
\pgfsetfillcolor{currentfill}%
\pgfsetlinewidth{0.803000pt}%
\definecolor{currentstroke}{rgb}{0.000000,0.000000,0.000000}%
\pgfsetstrokecolor{currentstroke}%
\pgfsetdash{}{0pt}%
\pgfsys@defobject{currentmarker}{\pgfqpoint{0.000000in}{-0.048611in}}{\pgfqpoint{0.000000in}{0.000000in}}{%
\pgfpathmoveto{\pgfqpoint{0.000000in}{0.000000in}}%
\pgfpathlineto{\pgfqpoint{0.000000in}{-0.048611in}}%
\pgfusepath{stroke,fill}%
}%
\begin{pgfscope}%
\pgfsys@transformshift{1.776053in}{0.318319in}%
\pgfsys@useobject{currentmarker}{}%
\end{pgfscope}%
\end{pgfscope}%
\begin{pgfscope}%
\pgftext[x=1.776053in,y=0.221097in,,top]{\rmfamily\fontsize{7.000000}{8.400000}\selectfont 4}%
\end{pgfscope}%
\begin{pgfscope}%
\pgfsetbuttcap%
\pgfsetroundjoin%
\definecolor{currentfill}{rgb}{0.000000,0.000000,0.000000}%
\pgfsetfillcolor{currentfill}%
\pgfsetlinewidth{0.803000pt}%
\definecolor{currentstroke}{rgb}{0.000000,0.000000,0.000000}%
\pgfsetstrokecolor{currentstroke}%
\pgfsetdash{}{0pt}%
\pgfsys@defobject{currentmarker}{\pgfqpoint{-0.048611in}{0.000000in}}{\pgfqpoint{0.000000in}{0.000000in}}{%
\pgfpathmoveto{\pgfqpoint{0.000000in}{0.000000in}}%
\pgfpathlineto{\pgfqpoint{-0.048611in}{0.000000in}}%
\pgfusepath{stroke,fill}%
}%
\begin{pgfscope}%
\pgfsys@transformshift{0.375397in}{0.355164in}%
\pgfsys@useobject{currentmarker}{}%
\end{pgfscope}%
\end{pgfscope}%
\begin{pgfscope}%
\pgftext[x=0.134463in,y=0.321684in,left,base]{\rmfamily\fontsize{7.000000}{8.400000}\selectfont \(\displaystyle 0.0\)}%
\end{pgfscope}%
\begin{pgfscope}%
\pgfsetbuttcap%
\pgfsetroundjoin%
\definecolor{currentfill}{rgb}{0.000000,0.000000,0.000000}%
\pgfsetfillcolor{currentfill}%
\pgfsetlinewidth{0.803000pt}%
\definecolor{currentstroke}{rgb}{0.000000,0.000000,0.000000}%
\pgfsetstrokecolor{currentstroke}%
\pgfsetdash{}{0pt}%
\pgfsys@defobject{currentmarker}{\pgfqpoint{-0.048611in}{0.000000in}}{\pgfqpoint{0.000000in}{0.000000in}}{%
\pgfpathmoveto{\pgfqpoint{0.000000in}{0.000000in}}%
\pgfpathlineto{\pgfqpoint{-0.048611in}{0.000000in}}%
\pgfusepath{stroke,fill}%
}%
\begin{pgfscope}%
\pgfsys@transformshift{0.375397in}{0.598142in}%
\pgfsys@useobject{currentmarker}{}%
\end{pgfscope}%
\end{pgfscope}%
\begin{pgfscope}%
\pgftext[x=0.134463in,y=0.564663in,left,base]{\rmfamily\fontsize{7.000000}{8.400000}\selectfont \(\displaystyle 0.2\)}%
\end{pgfscope}%
\begin{pgfscope}%
\pgfsetbuttcap%
\pgfsetroundjoin%
\definecolor{currentfill}{rgb}{0.000000,0.000000,0.000000}%
\pgfsetfillcolor{currentfill}%
\pgfsetlinewidth{0.803000pt}%
\definecolor{currentstroke}{rgb}{0.000000,0.000000,0.000000}%
\pgfsetstrokecolor{currentstroke}%
\pgfsetdash{}{0pt}%
\pgfsys@defobject{currentmarker}{\pgfqpoint{-0.048611in}{0.000000in}}{\pgfqpoint{0.000000in}{0.000000in}}{%
\pgfpathmoveto{\pgfqpoint{0.000000in}{0.000000in}}%
\pgfpathlineto{\pgfqpoint{-0.048611in}{0.000000in}}%
\pgfusepath{stroke,fill}%
}%
\begin{pgfscope}%
\pgfsys@transformshift{0.375397in}{0.841121in}%
\pgfsys@useobject{currentmarker}{}%
\end{pgfscope}%
\end{pgfscope}%
\begin{pgfscope}%
\pgftext[x=0.134463in,y=0.807641in,left,base]{\rmfamily\fontsize{7.000000}{8.400000}\selectfont \(\displaystyle 0.4\)}%
\end{pgfscope}%
\begin{pgfscope}%
\pgfsetbuttcap%
\pgfsetroundjoin%
\definecolor{currentfill}{rgb}{0.000000,0.000000,0.000000}%
\pgfsetfillcolor{currentfill}%
\pgfsetlinewidth{0.803000pt}%
\definecolor{currentstroke}{rgb}{0.000000,0.000000,0.000000}%
\pgfsetstrokecolor{currentstroke}%
\pgfsetdash{}{0pt}%
\pgfsys@defobject{currentmarker}{\pgfqpoint{-0.048611in}{0.000000in}}{\pgfqpoint{0.000000in}{0.000000in}}{%
\pgfpathmoveto{\pgfqpoint{0.000000in}{0.000000in}}%
\pgfpathlineto{\pgfqpoint{-0.048611in}{0.000000in}}%
\pgfusepath{stroke,fill}%
}%
\begin{pgfscope}%
\pgfsys@transformshift{0.375397in}{1.084099in}%
\pgfsys@useobject{currentmarker}{}%
\end{pgfscope}%
\end{pgfscope}%
\begin{pgfscope}%
\pgftext[x=0.134463in,y=1.050619in,left,base]{\rmfamily\fontsize{7.000000}{8.400000}\selectfont \(\displaystyle 0.6\)}%
\end{pgfscope}%
\begin{pgfscope}%
\pgfpathrectangle{\pgfqpoint{0.375397in}{0.318319in}}{\pgfqpoint{1.556284in}{0.810592in}} %
\pgfusepath{clip}%
\pgfsetrectcap%
\pgfsetroundjoin%
\pgfsetlinewidth{1.505625pt}%
\definecolor{currentstroke}{rgb}{0.247059,0.247059,0.247059}%
\pgfsetstrokecolor{currentstroke}%
\pgfsetdash{}{0pt}%
\pgfpathmoveto{\pgfqpoint{0.531026in}{0.430908in}}%
\pgfpathlineto{\pgfqpoint{0.531026in}{0.355164in}}%
\pgfusepath{stroke}%
\end{pgfscope}%
\begin{pgfscope}%
\pgfpathrectangle{\pgfqpoint{0.375397in}{0.318319in}}{\pgfqpoint{1.556284in}{0.810592in}} %
\pgfusepath{clip}%
\pgfsetrectcap%
\pgfsetroundjoin%
\pgfsetlinewidth{1.505625pt}%
\definecolor{currentstroke}{rgb}{0.247059,0.247059,0.247059}%
\pgfsetstrokecolor{currentstroke}%
\pgfsetdash{}{0pt}%
\pgfpathmoveto{\pgfqpoint{0.531026in}{0.667378in}}%
\pgfpathlineto{\pgfqpoint{0.531026in}{0.795562in}}%
\pgfusepath{stroke}%
\end{pgfscope}%
\begin{pgfscope}%
\pgfpathrectangle{\pgfqpoint{0.375397in}{0.318319in}}{\pgfqpoint{1.556284in}{0.810592in}} %
\pgfusepath{clip}%
\pgfsetrectcap%
\pgfsetroundjoin%
\pgfsetlinewidth{1.505625pt}%
\definecolor{currentstroke}{rgb}{0.247059,0.247059,0.247059}%
\pgfsetstrokecolor{currentstroke}%
\pgfsetdash{}{0pt}%
\pgfpathmoveto{\pgfqpoint{0.468774in}{0.355164in}}%
\pgfpathlineto{\pgfqpoint{0.593277in}{0.355164in}}%
\pgfusepath{stroke}%
\end{pgfscope}%
\begin{pgfscope}%
\pgfpathrectangle{\pgfqpoint{0.375397in}{0.318319in}}{\pgfqpoint{1.556284in}{0.810592in}} %
\pgfusepath{clip}%
\pgfsetrectcap%
\pgfsetroundjoin%
\pgfsetlinewidth{1.505625pt}%
\definecolor{currentstroke}{rgb}{0.247059,0.247059,0.247059}%
\pgfsetstrokecolor{currentstroke}%
\pgfsetdash{}{0pt}%
\pgfpathmoveto{\pgfqpoint{0.468774in}{0.795562in}}%
\pgfpathlineto{\pgfqpoint{0.593277in}{0.795562in}}%
\pgfusepath{stroke}%
\end{pgfscope}%
\begin{pgfscope}%
\pgfpathrectangle{\pgfqpoint{0.375397in}{0.318319in}}{\pgfqpoint{1.556284in}{0.810592in}} %
\pgfusepath{clip}%
\pgfsetbuttcap%
\pgfsetmiterjoin%
\definecolor{currentfill}{rgb}{0.247059,0.247059,0.247059}%
\pgfsetfillcolor{currentfill}%
\pgfsetlinewidth{1.003750pt}%
\definecolor{currentstroke}{rgb}{0.247059,0.247059,0.247059}%
\pgfsetstrokecolor{currentstroke}%
\pgfsetdash{}{0pt}%
\pgfsys@defobject{currentmarker}{\pgfqpoint{-0.029463in}{-0.049105in}}{\pgfqpoint{0.029463in}{0.049105in}}{%
\pgfpathmoveto{\pgfqpoint{-0.000000in}{-0.049105in}}%
\pgfpathlineto{\pgfqpoint{0.029463in}{0.000000in}}%
\pgfpathlineto{\pgfqpoint{0.000000in}{0.049105in}}%
\pgfpathlineto{\pgfqpoint{-0.029463in}{0.000000in}}%
\pgfpathclose%
\pgfusepath{stroke,fill}%
}%
\begin{pgfscope}%
\pgfsys@transformshift{0.531026in}{1.022499in}%
\pgfsys@useobject{currentmarker}{}%
\end{pgfscope}%
\end{pgfscope}%
\begin{pgfscope}%
\pgfpathrectangle{\pgfqpoint{0.375397in}{0.318319in}}{\pgfqpoint{1.556284in}{0.810592in}} %
\pgfusepath{clip}%
\pgfsetrectcap%
\pgfsetroundjoin%
\pgfsetlinewidth{1.505625pt}%
\definecolor{currentstroke}{rgb}{0.247059,0.247059,0.247059}%
\pgfsetstrokecolor{currentstroke}%
\pgfsetdash{}{0pt}%
\pgfpathmoveto{\pgfqpoint{0.842282in}{0.434270in}}%
\pgfpathlineto{\pgfqpoint{0.842282in}{0.355164in}}%
\pgfusepath{stroke}%
\end{pgfscope}%
\begin{pgfscope}%
\pgfpathrectangle{\pgfqpoint{0.375397in}{0.318319in}}{\pgfqpoint{1.556284in}{0.810592in}} %
\pgfusepath{clip}%
\pgfsetrectcap%
\pgfsetroundjoin%
\pgfsetlinewidth{1.505625pt}%
\definecolor{currentstroke}{rgb}{0.247059,0.247059,0.247059}%
\pgfsetstrokecolor{currentstroke}%
\pgfsetdash{}{0pt}%
\pgfpathmoveto{\pgfqpoint{0.842282in}{0.920313in}}%
\pgfpathlineto{\pgfqpoint{0.842282in}{1.092065in}}%
\pgfusepath{stroke}%
\end{pgfscope}%
\begin{pgfscope}%
\pgfpathrectangle{\pgfqpoint{0.375397in}{0.318319in}}{\pgfqpoint{1.556284in}{0.810592in}} %
\pgfusepath{clip}%
\pgfsetrectcap%
\pgfsetroundjoin%
\pgfsetlinewidth{1.505625pt}%
\definecolor{currentstroke}{rgb}{0.247059,0.247059,0.247059}%
\pgfsetstrokecolor{currentstroke}%
\pgfsetdash{}{0pt}%
\pgfpathmoveto{\pgfqpoint{0.780031in}{0.355164in}}%
\pgfpathlineto{\pgfqpoint{0.904534in}{0.355164in}}%
\pgfusepath{stroke}%
\end{pgfscope}%
\begin{pgfscope}%
\pgfpathrectangle{\pgfqpoint{0.375397in}{0.318319in}}{\pgfqpoint{1.556284in}{0.810592in}} %
\pgfusepath{clip}%
\pgfsetrectcap%
\pgfsetroundjoin%
\pgfsetlinewidth{1.505625pt}%
\definecolor{currentstroke}{rgb}{0.247059,0.247059,0.247059}%
\pgfsetstrokecolor{currentstroke}%
\pgfsetdash{}{0pt}%
\pgfpathmoveto{\pgfqpoint{0.780031in}{1.092065in}}%
\pgfpathlineto{\pgfqpoint{0.904534in}{1.092065in}}%
\pgfusepath{stroke}%
\end{pgfscope}%
\begin{pgfscope}%
\pgfpathrectangle{\pgfqpoint{0.375397in}{0.318319in}}{\pgfqpoint{1.556284in}{0.810592in}} %
\pgfusepath{clip}%
\pgfsetrectcap%
\pgfsetroundjoin%
\pgfsetlinewidth{1.505625pt}%
\definecolor{currentstroke}{rgb}{0.247059,0.247059,0.247059}%
\pgfsetstrokecolor{currentstroke}%
\pgfsetdash{}{0pt}%
\pgfpathmoveto{\pgfqpoint{1.153539in}{0.400958in}}%
\pgfpathlineto{\pgfqpoint{1.153539in}{0.355164in}}%
\pgfusepath{stroke}%
\end{pgfscope}%
\begin{pgfscope}%
\pgfpathrectangle{\pgfqpoint{0.375397in}{0.318319in}}{\pgfqpoint{1.556284in}{0.810592in}} %
\pgfusepath{clip}%
\pgfsetrectcap%
\pgfsetroundjoin%
\pgfsetlinewidth{1.505625pt}%
\definecolor{currentstroke}{rgb}{0.247059,0.247059,0.247059}%
\pgfsetstrokecolor{currentstroke}%
\pgfsetdash{}{0pt}%
\pgfpathmoveto{\pgfqpoint{1.153539in}{0.520633in}}%
\pgfpathlineto{\pgfqpoint{1.153539in}{0.650040in}}%
\pgfusepath{stroke}%
\end{pgfscope}%
\begin{pgfscope}%
\pgfpathrectangle{\pgfqpoint{0.375397in}{0.318319in}}{\pgfqpoint{1.556284in}{0.810592in}} %
\pgfusepath{clip}%
\pgfsetrectcap%
\pgfsetroundjoin%
\pgfsetlinewidth{1.505625pt}%
\definecolor{currentstroke}{rgb}{0.247059,0.247059,0.247059}%
\pgfsetstrokecolor{currentstroke}%
\pgfsetdash{}{0pt}%
\pgfpathmoveto{\pgfqpoint{1.091288in}{0.355164in}}%
\pgfpathlineto{\pgfqpoint{1.215791in}{0.355164in}}%
\pgfusepath{stroke}%
\end{pgfscope}%
\begin{pgfscope}%
\pgfpathrectangle{\pgfqpoint{0.375397in}{0.318319in}}{\pgfqpoint{1.556284in}{0.810592in}} %
\pgfusepath{clip}%
\pgfsetrectcap%
\pgfsetroundjoin%
\pgfsetlinewidth{1.505625pt}%
\definecolor{currentstroke}{rgb}{0.247059,0.247059,0.247059}%
\pgfsetstrokecolor{currentstroke}%
\pgfsetdash{}{0pt}%
\pgfpathmoveto{\pgfqpoint{1.091288in}{0.650040in}}%
\pgfpathlineto{\pgfqpoint{1.215791in}{0.650040in}}%
\pgfusepath{stroke}%
\end{pgfscope}%
\begin{pgfscope}%
\pgfpathrectangle{\pgfqpoint{0.375397in}{0.318319in}}{\pgfqpoint{1.556284in}{0.810592in}} %
\pgfusepath{clip}%
\pgfsetbuttcap%
\pgfsetmiterjoin%
\definecolor{currentfill}{rgb}{0.247059,0.247059,0.247059}%
\pgfsetfillcolor{currentfill}%
\pgfsetlinewidth{1.003750pt}%
\definecolor{currentstroke}{rgb}{0.247059,0.247059,0.247059}%
\pgfsetstrokecolor{currentstroke}%
\pgfsetdash{}{0pt}%
\pgfsys@defobject{currentmarker}{\pgfqpoint{-0.029463in}{-0.049105in}}{\pgfqpoint{0.029463in}{0.049105in}}{%
\pgfpathmoveto{\pgfqpoint{-0.000000in}{-0.049105in}}%
\pgfpathlineto{\pgfqpoint{0.029463in}{0.000000in}}%
\pgfpathlineto{\pgfqpoint{0.000000in}{0.049105in}}%
\pgfpathlineto{\pgfqpoint{-0.029463in}{0.000000in}}%
\pgfpathclose%
\pgfusepath{stroke,fill}%
}%
\begin{pgfscope}%
\pgfsys@transformshift{1.153539in}{0.729850in}%
\pgfsys@useobject{currentmarker}{}%
\end{pgfscope}%
\begin{pgfscope}%
\pgfsys@transformshift{1.153539in}{0.850712in}%
\pgfsys@useobject{currentmarker}{}%
\end{pgfscope}%
\end{pgfscope}%
\begin{pgfscope}%
\pgfpathrectangle{\pgfqpoint{0.375397in}{0.318319in}}{\pgfqpoint{1.556284in}{0.810592in}} %
\pgfusepath{clip}%
\pgfsetrectcap%
\pgfsetroundjoin%
\pgfsetlinewidth{1.505625pt}%
\definecolor{currentstroke}{rgb}{0.247059,0.247059,0.247059}%
\pgfsetstrokecolor{currentstroke}%
\pgfsetdash{}{0pt}%
\pgfpathmoveto{\pgfqpoint{1.464796in}{0.417331in}}%
\pgfpathlineto{\pgfqpoint{1.464796in}{0.355164in}}%
\pgfusepath{stroke}%
\end{pgfscope}%
\begin{pgfscope}%
\pgfpathrectangle{\pgfqpoint{0.375397in}{0.318319in}}{\pgfqpoint{1.556284in}{0.810592in}} %
\pgfusepath{clip}%
\pgfsetrectcap%
\pgfsetroundjoin%
\pgfsetlinewidth{1.505625pt}%
\definecolor{currentstroke}{rgb}{0.247059,0.247059,0.247059}%
\pgfsetstrokecolor{currentstroke}%
\pgfsetdash{}{0pt}%
\pgfpathmoveto{\pgfqpoint{1.464796in}{0.529110in}}%
\pgfpathlineto{\pgfqpoint{1.464796in}{0.685134in}}%
\pgfusepath{stroke}%
\end{pgfscope}%
\begin{pgfscope}%
\pgfpathrectangle{\pgfqpoint{0.375397in}{0.318319in}}{\pgfqpoint{1.556284in}{0.810592in}} %
\pgfusepath{clip}%
\pgfsetrectcap%
\pgfsetroundjoin%
\pgfsetlinewidth{1.505625pt}%
\definecolor{currentstroke}{rgb}{0.247059,0.247059,0.247059}%
\pgfsetstrokecolor{currentstroke}%
\pgfsetdash{}{0pt}%
\pgfpathmoveto{\pgfqpoint{1.402545in}{0.355164in}}%
\pgfpathlineto{\pgfqpoint{1.527047in}{0.355164in}}%
\pgfusepath{stroke}%
\end{pgfscope}%
\begin{pgfscope}%
\pgfpathrectangle{\pgfqpoint{0.375397in}{0.318319in}}{\pgfqpoint{1.556284in}{0.810592in}} %
\pgfusepath{clip}%
\pgfsetrectcap%
\pgfsetroundjoin%
\pgfsetlinewidth{1.505625pt}%
\definecolor{currentstroke}{rgb}{0.247059,0.247059,0.247059}%
\pgfsetstrokecolor{currentstroke}%
\pgfsetdash{}{0pt}%
\pgfpathmoveto{\pgfqpoint{1.402545in}{0.685134in}}%
\pgfpathlineto{\pgfqpoint{1.527047in}{0.685134in}}%
\pgfusepath{stroke}%
\end{pgfscope}%
\begin{pgfscope}%
\pgfpathrectangle{\pgfqpoint{0.375397in}{0.318319in}}{\pgfqpoint{1.556284in}{0.810592in}} %
\pgfusepath{clip}%
\pgfsetbuttcap%
\pgfsetmiterjoin%
\definecolor{currentfill}{rgb}{0.247059,0.247059,0.247059}%
\pgfsetfillcolor{currentfill}%
\pgfsetlinewidth{1.003750pt}%
\definecolor{currentstroke}{rgb}{0.247059,0.247059,0.247059}%
\pgfsetstrokecolor{currentstroke}%
\pgfsetdash{}{0pt}%
\pgfsys@defobject{currentmarker}{\pgfqpoint{-0.029463in}{-0.049105in}}{\pgfqpoint{0.029463in}{0.049105in}}{%
\pgfpathmoveto{\pgfqpoint{-0.000000in}{-0.049105in}}%
\pgfpathlineto{\pgfqpoint{0.029463in}{0.000000in}}%
\pgfpathlineto{\pgfqpoint{0.000000in}{0.049105in}}%
\pgfpathlineto{\pgfqpoint{-0.029463in}{0.000000in}}%
\pgfpathclose%
\pgfusepath{stroke,fill}%
}%
\begin{pgfscope}%
\pgfsys@transformshift{1.464796in}{1.034680in}%
\pgfsys@useobject{currentmarker}{}%
\end{pgfscope}%
\end{pgfscope}%
\begin{pgfscope}%
\pgfpathrectangle{\pgfqpoint{0.375397in}{0.318319in}}{\pgfqpoint{1.556284in}{0.810592in}} %
\pgfusepath{clip}%
\pgfsetrectcap%
\pgfsetroundjoin%
\pgfsetlinewidth{1.505625pt}%
\definecolor{currentstroke}{rgb}{0.247059,0.247059,0.247059}%
\pgfsetstrokecolor{currentstroke}%
\pgfsetdash{}{0pt}%
\pgfpathmoveto{\pgfqpoint{1.776053in}{0.429467in}}%
\pgfpathlineto{\pgfqpoint{1.776053in}{0.357271in}}%
\pgfusepath{stroke}%
\end{pgfscope}%
\begin{pgfscope}%
\pgfpathrectangle{\pgfqpoint{0.375397in}{0.318319in}}{\pgfqpoint{1.556284in}{0.810592in}} %
\pgfusepath{clip}%
\pgfsetrectcap%
\pgfsetroundjoin%
\pgfsetlinewidth{1.505625pt}%
\definecolor{currentstroke}{rgb}{0.247059,0.247059,0.247059}%
\pgfsetstrokecolor{currentstroke}%
\pgfsetdash{}{0pt}%
\pgfpathmoveto{\pgfqpoint{1.776053in}{0.721052in}}%
\pgfpathlineto{\pgfqpoint{1.776053in}{1.084099in}}%
\pgfusepath{stroke}%
\end{pgfscope}%
\begin{pgfscope}%
\pgfpathrectangle{\pgfqpoint{0.375397in}{0.318319in}}{\pgfqpoint{1.556284in}{0.810592in}} %
\pgfusepath{clip}%
\pgfsetrectcap%
\pgfsetroundjoin%
\pgfsetlinewidth{1.505625pt}%
\definecolor{currentstroke}{rgb}{0.247059,0.247059,0.247059}%
\pgfsetstrokecolor{currentstroke}%
\pgfsetdash{}{0pt}%
\pgfpathmoveto{\pgfqpoint{1.713801in}{0.357271in}}%
\pgfpathlineto{\pgfqpoint{1.838304in}{0.357271in}}%
\pgfusepath{stroke}%
\end{pgfscope}%
\begin{pgfscope}%
\pgfpathrectangle{\pgfqpoint{0.375397in}{0.318319in}}{\pgfqpoint{1.556284in}{0.810592in}} %
\pgfusepath{clip}%
\pgfsetrectcap%
\pgfsetroundjoin%
\pgfsetlinewidth{1.505625pt}%
\definecolor{currentstroke}{rgb}{0.247059,0.247059,0.247059}%
\pgfsetstrokecolor{currentstroke}%
\pgfsetdash{}{0pt}%
\pgfpathmoveto{\pgfqpoint{1.713801in}{1.084099in}}%
\pgfpathlineto{\pgfqpoint{1.838304in}{1.084099in}}%
\pgfusepath{stroke}%
\end{pgfscope}%
\begin{pgfscope}%
\pgfpathrectangle{\pgfqpoint{0.375397in}{0.318319in}}{\pgfqpoint{1.556284in}{0.810592in}} %
\pgfusepath{clip}%
\pgfsetrectcap%
\pgfsetroundjoin%
\pgfsetlinewidth{1.505625pt}%
\definecolor{currentstroke}{rgb}{0.247059,0.247059,0.247059}%
\pgfsetstrokecolor{currentstroke}%
\pgfsetdash{}{0pt}%
\pgfpathmoveto{\pgfqpoint{0.406523in}{0.518846in}}%
\pgfpathlineto{\pgfqpoint{0.655528in}{0.518846in}}%
\pgfusepath{stroke}%
\end{pgfscope}%
\begin{pgfscope}%
\pgfpathrectangle{\pgfqpoint{0.375397in}{0.318319in}}{\pgfqpoint{1.556284in}{0.810592in}} %
\pgfusepath{clip}%
\pgfsetrectcap%
\pgfsetroundjoin%
\pgfsetlinewidth{1.505625pt}%
\definecolor{currentstroke}{rgb}{0.247059,0.247059,0.247059}%
\pgfsetstrokecolor{currentstroke}%
\pgfsetdash{}{0pt}%
\pgfpathmoveto{\pgfqpoint{0.717780in}{0.510530in}}%
\pgfpathlineto{\pgfqpoint{0.966785in}{0.510530in}}%
\pgfusepath{stroke}%
\end{pgfscope}%
\begin{pgfscope}%
\pgfpathrectangle{\pgfqpoint{0.375397in}{0.318319in}}{\pgfqpoint{1.556284in}{0.810592in}} %
\pgfusepath{clip}%
\pgfsetrectcap%
\pgfsetroundjoin%
\pgfsetlinewidth{1.505625pt}%
\definecolor{currentstroke}{rgb}{0.247059,0.247059,0.247059}%
\pgfsetstrokecolor{currentstroke}%
\pgfsetdash{}{0pt}%
\pgfpathmoveto{\pgfqpoint{1.029037in}{0.435593in}}%
\pgfpathlineto{\pgfqpoint{1.278042in}{0.435593in}}%
\pgfusepath{stroke}%
\end{pgfscope}%
\begin{pgfscope}%
\pgfpathrectangle{\pgfqpoint{0.375397in}{0.318319in}}{\pgfqpoint{1.556284in}{0.810592in}} %
\pgfusepath{clip}%
\pgfsetrectcap%
\pgfsetroundjoin%
\pgfsetlinewidth{1.505625pt}%
\definecolor{currentstroke}{rgb}{0.247059,0.247059,0.247059}%
\pgfsetstrokecolor{currentstroke}%
\pgfsetdash{}{0pt}%
\pgfpathmoveto{\pgfqpoint{1.340293in}{0.433476in}}%
\pgfpathlineto{\pgfqpoint{1.589299in}{0.433476in}}%
\pgfusepath{stroke}%
\end{pgfscope}%
\begin{pgfscope}%
\pgfpathrectangle{\pgfqpoint{0.375397in}{0.318319in}}{\pgfqpoint{1.556284in}{0.810592in}} %
\pgfusepath{clip}%
\pgfsetrectcap%
\pgfsetroundjoin%
\pgfsetlinewidth{1.505625pt}%
\definecolor{currentstroke}{rgb}{0.247059,0.247059,0.247059}%
\pgfsetstrokecolor{currentstroke}%
\pgfsetdash{}{0pt}%
\pgfpathmoveto{\pgfqpoint{1.651550in}{0.549998in}}%
\pgfpathlineto{\pgfqpoint{1.900556in}{0.549998in}}%
\pgfusepath{stroke}%
\end{pgfscope}%
\begin{pgfscope}%
\pgfsetrectcap%
\pgfsetmiterjoin%
\pgfsetlinewidth{0.803000pt}%
\definecolor{currentstroke}{rgb}{0.000000,0.000000,0.000000}%
\pgfsetstrokecolor{currentstroke}%
\pgfsetdash{}{0pt}%
\pgfpathmoveto{\pgfqpoint{0.375397in}{0.318319in}}%
\pgfpathlineto{\pgfqpoint{0.375397in}{1.128910in}}%
\pgfusepath{stroke}%
\end{pgfscope}%
\begin{pgfscope}%
\pgfsetrectcap%
\pgfsetmiterjoin%
\pgfsetlinewidth{0.803000pt}%
\definecolor{currentstroke}{rgb}{0.000000,0.000000,0.000000}%
\pgfsetstrokecolor{currentstroke}%
\pgfsetdash{}{0pt}%
\pgfpathmoveto{\pgfqpoint{1.931681in}{0.318319in}}%
\pgfpathlineto{\pgfqpoint{1.931681in}{1.128910in}}%
\pgfusepath{stroke}%
\end{pgfscope}%
\begin{pgfscope}%
\pgfsetrectcap%
\pgfsetmiterjoin%
\pgfsetlinewidth{0.803000pt}%
\definecolor{currentstroke}{rgb}{0.000000,0.000000,0.000000}%
\pgfsetstrokecolor{currentstroke}%
\pgfsetdash{}{0pt}%
\pgfpathmoveto{\pgfqpoint{0.375397in}{0.318319in}}%
\pgfpathlineto{\pgfqpoint{1.931681in}{0.318319in}}%
\pgfusepath{stroke}%
\end{pgfscope}%
\begin{pgfscope}%
\pgfsetrectcap%
\pgfsetmiterjoin%
\pgfsetlinewidth{0.803000pt}%
\definecolor{currentstroke}{rgb}{0.000000,0.000000,0.000000}%
\pgfsetstrokecolor{currentstroke}%
\pgfsetdash{}{0pt}%
\pgfpathmoveto{\pgfqpoint{0.375397in}{1.128910in}}%
\pgfpathlineto{\pgfqpoint{1.931681in}{1.128910in}}%
\pgfusepath{stroke}%
\end{pgfscope}%
\end{pgfpicture}%
\makeatother%
\endgroup%

%% file: figures/nlp_translation_sentiment_ner/train_test_stats.pgf
\begingroup%
\makeatletter%
\begin{pgfpicture}%
\pgfpathrectangle{\pgfpointorigin}{\pgfqpoint{2.101681in}{1.298910in}}%
\pgfusepath{use as bounding box, clip}%
\begin{pgfscope}%
\pgfsetbuttcap%
\pgfsetmiterjoin%
\pgfsetlinewidth{0.000000pt}%
\definecolor{currentstroke}{rgb}{1.000000,1.000000,1.000000}%
\pgfsetstrokecolor{currentstroke}%
\pgfsetstrokeopacity{0.000000}%
\pgfsetdash{}{0pt}%
\pgfpathmoveto{\pgfqpoint{0.000000in}{0.000000in}}%
\pgfpathlineto{\pgfqpoint{2.101681in}{0.000000in}}%
\pgfpathlineto{\pgfqpoint{2.101681in}{1.298910in}}%
\pgfpathlineto{\pgfqpoint{0.000000in}{1.298910in}}%
\pgfpathclose%
\pgfusepath{}%
\end{pgfscope}%
\begin{pgfscope}%
\pgfsetbuttcap%
\pgfsetmiterjoin%
\definecolor{currentfill}{rgb}{1.000000,1.000000,1.000000}%
\pgfsetfillcolor{currentfill}%
\pgfsetlinewidth{0.000000pt}%
\definecolor{currentstroke}{rgb}{0.000000,0.000000,0.000000}%
\pgfsetstrokecolor{currentstroke}%
\pgfsetstrokeopacity{0.000000}%
\pgfsetdash{}{0pt}%
\pgfpathmoveto{\pgfqpoint{0.375397in}{0.318319in}}%
\pgfpathlineto{\pgfqpoint{1.931681in}{0.318319in}}%
\pgfpathlineto{\pgfqpoint{1.931681in}{1.128910in}}%
\pgfpathlineto{\pgfqpoint{0.375397in}{1.128910in}}%
\pgfpathclose%
\pgfusepath{fill}%
\end{pgfscope}%
\begin{pgfscope}%
\pgfpathrectangle{\pgfqpoint{0.375397in}{0.318319in}}{\pgfqpoint{1.556284in}{0.810592in}} %
\pgfusepath{clip}%
\pgfsetbuttcap%
\pgfsetmiterjoin%
\definecolor{currentfill}{rgb}{0.881863,0.505392,0.173039}%
\pgfsetfillcolor{currentfill}%
\pgfsetlinewidth{1.505625pt}%
\definecolor{currentstroke}{rgb}{0.317647,0.317647,0.317647}%
\pgfsetstrokecolor{currentstroke}%
\pgfsetdash{}{0pt}%
\pgfpathmoveto{\pgfqpoint{0.406523in}{0.556144in}}%
\pgfpathlineto{\pgfqpoint{0.655528in}{0.556144in}}%
\pgfpathlineto{\pgfqpoint{0.655528in}{0.609636in}}%
\pgfpathlineto{\pgfqpoint{0.406523in}{0.609636in}}%
\pgfpathlineto{\pgfqpoint{0.406523in}{0.556144in}}%
\pgfpathclose%
\pgfusepath{stroke,fill}%
\end{pgfscope}%
\begin{pgfscope}%
\pgfpathrectangle{\pgfqpoint{0.375397in}{0.318319in}}{\pgfqpoint{1.556284in}{0.810592in}} %
\pgfusepath{clip}%
\pgfsetbuttcap%
\pgfsetmiterjoin%
\definecolor{currentfill}{rgb}{0.881863,0.505392,0.173039}%
\pgfsetfillcolor{currentfill}%
\pgfsetlinewidth{1.505625pt}%
\definecolor{currentstroke}{rgb}{0.317647,0.317647,0.317647}%
\pgfsetstrokecolor{currentstroke}%
\pgfsetdash{}{0pt}%
\pgfpathmoveto{\pgfqpoint{0.717780in}{0.602061in}}%
\pgfpathlineto{\pgfqpoint{0.966785in}{0.602061in}}%
\pgfpathlineto{\pgfqpoint{0.966785in}{0.718546in}}%
\pgfpathlineto{\pgfqpoint{0.717780in}{0.718546in}}%
\pgfpathlineto{\pgfqpoint{0.717780in}{0.602061in}}%
\pgfpathclose%
\pgfusepath{stroke,fill}%
\end{pgfscope}%
\begin{pgfscope}%
\pgfpathrectangle{\pgfqpoint{0.375397in}{0.318319in}}{\pgfqpoint{1.556284in}{0.810592in}} %
\pgfusepath{clip}%
\pgfsetbuttcap%
\pgfsetmiterjoin%
\definecolor{currentfill}{rgb}{0.881863,0.505392,0.173039}%
\pgfsetfillcolor{currentfill}%
\pgfsetlinewidth{1.505625pt}%
\definecolor{currentstroke}{rgb}{0.317647,0.317647,0.317647}%
\pgfsetstrokecolor{currentstroke}%
\pgfsetdash{}{0pt}%
\pgfpathmoveto{\pgfqpoint{1.029037in}{0.667220in}}%
\pgfpathlineto{\pgfqpoint{1.278042in}{0.667220in}}%
\pgfpathlineto{\pgfqpoint{1.278042in}{0.800817in}}%
\pgfpathlineto{\pgfqpoint{1.029037in}{0.800817in}}%
\pgfpathlineto{\pgfqpoint{1.029037in}{0.667220in}}%
\pgfpathclose%
\pgfusepath{stroke,fill}%
\end{pgfscope}%
\begin{pgfscope}%
\pgfpathrectangle{\pgfqpoint{0.375397in}{0.318319in}}{\pgfqpoint{1.556284in}{0.810592in}} %
\pgfusepath{clip}%
\pgfsetbuttcap%
\pgfsetmiterjoin%
\definecolor{currentfill}{rgb}{0.881863,0.505392,0.173039}%
\pgfsetfillcolor{currentfill}%
\pgfsetlinewidth{1.505625pt}%
\definecolor{currentstroke}{rgb}{0.317647,0.317647,0.317647}%
\pgfsetstrokecolor{currentstroke}%
\pgfsetdash{}{0pt}%
\pgfpathmoveto{\pgfqpoint{1.340293in}{0.585998in}}%
\pgfpathlineto{\pgfqpoint{1.589299in}{0.585998in}}%
\pgfpathlineto{\pgfqpoint{1.589299in}{0.687648in}}%
\pgfpathlineto{\pgfqpoint{1.340293in}{0.687648in}}%
\pgfpathlineto{\pgfqpoint{1.340293in}{0.585998in}}%
\pgfpathclose%
\pgfusepath{stroke,fill}%
\end{pgfscope}%
\begin{pgfscope}%
\pgfpathrectangle{\pgfqpoint{0.375397in}{0.318319in}}{\pgfqpoint{1.556284in}{0.810592in}} %
\pgfusepath{clip}%
\pgfsetbuttcap%
\pgfsetmiterjoin%
\definecolor{currentfill}{rgb}{0.881863,0.505392,0.173039}%
\pgfsetfillcolor{currentfill}%
\pgfsetlinewidth{1.505625pt}%
\definecolor{currentstroke}{rgb}{0.317647,0.317647,0.317647}%
\pgfsetstrokecolor{currentstroke}%
\pgfsetdash{}{0pt}%
\pgfpathmoveto{\pgfqpoint{1.651550in}{0.587854in}}%
\pgfpathlineto{\pgfqpoint{1.900556in}{0.587854in}}%
\pgfpathlineto{\pgfqpoint{1.900556in}{0.690263in}}%
\pgfpathlineto{\pgfqpoint{1.651550in}{0.690263in}}%
\pgfpathlineto{\pgfqpoint{1.651550in}{0.587854in}}%
\pgfpathclose%
\pgfusepath{stroke,fill}%
\end{pgfscope}%
\begin{pgfscope}%
\pgfsetbuttcap%
\pgfsetroundjoin%
\definecolor{currentfill}{rgb}{0.000000,0.000000,0.000000}%
\pgfsetfillcolor{currentfill}%
\pgfsetlinewidth{0.803000pt}%
\definecolor{currentstroke}{rgb}{0.000000,0.000000,0.000000}%
\pgfsetstrokecolor{currentstroke}%
\pgfsetdash{}{0pt}%
\pgfsys@defobject{currentmarker}{\pgfqpoint{0.000000in}{-0.048611in}}{\pgfqpoint{0.000000in}{0.000000in}}{%
\pgfpathmoveto{\pgfqpoint{0.000000in}{0.000000in}}%
\pgfpathlineto{\pgfqpoint{0.000000in}{-0.048611in}}%
\pgfusepath{stroke,fill}%
}%
\begin{pgfscope}%
\pgfsys@transformshift{0.531026in}{0.318319in}%
\pgfsys@useobject{currentmarker}{}%
\end{pgfscope}%
\end{pgfscope}%
\begin{pgfscope}%
\pgftext[x=0.531026in,y=0.221097in,,top]{\rmfamily\fontsize{7.000000}{8.400000}\selectfont 0}%
\end{pgfscope}%
\begin{pgfscope}%
\pgfsetbuttcap%
\pgfsetroundjoin%
\definecolor{currentfill}{rgb}{0.000000,0.000000,0.000000}%
\pgfsetfillcolor{currentfill}%
\pgfsetlinewidth{0.803000pt}%
\definecolor{currentstroke}{rgb}{0.000000,0.000000,0.000000}%
\pgfsetstrokecolor{currentstroke}%
\pgfsetdash{}{0pt}%
\pgfsys@defobject{currentmarker}{\pgfqpoint{0.000000in}{-0.048611in}}{\pgfqpoint{0.000000in}{0.000000in}}{%
\pgfpathmoveto{\pgfqpoint{0.000000in}{0.000000in}}%
\pgfpathlineto{\pgfqpoint{0.000000in}{-0.048611in}}%
\pgfusepath{stroke,fill}%
}%
\begin{pgfscope}%
\pgfsys@transformshift{0.842282in}{0.318319in}%
\pgfsys@useobject{currentmarker}{}%
\end{pgfscope}%
\end{pgfscope}%
\begin{pgfscope}%
\pgftext[x=0.842282in,y=0.221097in,,top]{\rmfamily\fontsize{7.000000}{8.400000}\selectfont 1}%
\end{pgfscope}%
\begin{pgfscope}%
\pgfsetbuttcap%
\pgfsetroundjoin%
\definecolor{currentfill}{rgb}{0.000000,0.000000,0.000000}%
\pgfsetfillcolor{currentfill}%
\pgfsetlinewidth{0.803000pt}%
\definecolor{currentstroke}{rgb}{0.000000,0.000000,0.000000}%
\pgfsetstrokecolor{currentstroke}%
\pgfsetdash{}{0pt}%
\pgfsys@defobject{currentmarker}{\pgfqpoint{0.000000in}{-0.048611in}}{\pgfqpoint{0.000000in}{0.000000in}}{%
\pgfpathmoveto{\pgfqpoint{0.000000in}{0.000000in}}%
\pgfpathlineto{\pgfqpoint{0.000000in}{-0.048611in}}%
\pgfusepath{stroke,fill}%
}%
\begin{pgfscope}%
\pgfsys@transformshift{1.153539in}{0.318319in}%
\pgfsys@useobject{currentmarker}{}%
\end{pgfscope}%
\end{pgfscope}%
\begin{pgfscope}%
\pgftext[x=1.153539in,y=0.221097in,,top]{\rmfamily\fontsize{7.000000}{8.400000}\selectfont 2}%
\end{pgfscope}%
\begin{pgfscope}%
\pgfsetbuttcap%
\pgfsetroundjoin%
\definecolor{currentfill}{rgb}{0.000000,0.000000,0.000000}%
\pgfsetfillcolor{currentfill}%
\pgfsetlinewidth{0.803000pt}%
\definecolor{currentstroke}{rgb}{0.000000,0.000000,0.000000}%
\pgfsetstrokecolor{currentstroke}%
\pgfsetdash{}{0pt}%
\pgfsys@defobject{currentmarker}{\pgfqpoint{0.000000in}{-0.048611in}}{\pgfqpoint{0.000000in}{0.000000in}}{%
\pgfpathmoveto{\pgfqpoint{0.000000in}{0.000000in}}%
\pgfpathlineto{\pgfqpoint{0.000000in}{-0.048611in}}%
\pgfusepath{stroke,fill}%
}%
\begin{pgfscope}%
\pgfsys@transformshift{1.464796in}{0.318319in}%
\pgfsys@useobject{currentmarker}{}%
\end{pgfscope}%
\end{pgfscope}%
\begin{pgfscope}%
\pgftext[x=1.464796in,y=0.221097in,,top]{\rmfamily\fontsize{7.000000}{8.400000}\selectfont 3}%
\end{pgfscope}%
\begin{pgfscope}%
\pgfsetbuttcap%
\pgfsetroundjoin%
\definecolor{currentfill}{rgb}{0.000000,0.000000,0.000000}%
\pgfsetfillcolor{currentfill}%
\pgfsetlinewidth{0.803000pt}%
\definecolor{currentstroke}{rgb}{0.000000,0.000000,0.000000}%
\pgfsetstrokecolor{currentstroke}%
\pgfsetdash{}{0pt}%
\pgfsys@defobject{currentmarker}{\pgfqpoint{0.000000in}{-0.048611in}}{\pgfqpoint{0.000000in}{0.000000in}}{%
\pgfpathmoveto{\pgfqpoint{0.000000in}{0.000000in}}%
\pgfpathlineto{\pgfqpoint{0.000000in}{-0.048611in}}%
\pgfusepath{stroke,fill}%
}%
\begin{pgfscope}%
\pgfsys@transformshift{1.776053in}{0.318319in}%
\pgfsys@useobject{currentmarker}{}%
\end{pgfscope}%
\end{pgfscope}%
\begin{pgfscope}%
\pgftext[x=1.776053in,y=0.221097in,,top]{\rmfamily\fontsize{7.000000}{8.400000}\selectfont 4}%
\end{pgfscope}%
\begin{pgfscope}%
\pgfsetbuttcap%
\pgfsetroundjoin%
\definecolor{currentfill}{rgb}{0.000000,0.000000,0.000000}%
\pgfsetfillcolor{currentfill}%
\pgfsetlinewidth{0.803000pt}%
\definecolor{currentstroke}{rgb}{0.000000,0.000000,0.000000}%
\pgfsetstrokecolor{currentstroke}%
\pgfsetdash{}{0pt}%
\pgfsys@defobject{currentmarker}{\pgfqpoint{-0.048611in}{0.000000in}}{\pgfqpoint{0.000000in}{0.000000in}}{%
\pgfpathmoveto{\pgfqpoint{0.000000in}{0.000000in}}%
\pgfpathlineto{\pgfqpoint{-0.048611in}{0.000000in}}%
\pgfusepath{stroke,fill}%
}%
\begin{pgfscope}%
\pgfsys@transformshift{0.375397in}{0.355164in}%
\pgfsys@useobject{currentmarker}{}%
\end{pgfscope}%
\end{pgfscope}%
\begin{pgfscope}%
\pgftext[x=0.134463in,y=0.321684in,left,base]{\rmfamily\fontsize{7.000000}{8.400000}\selectfont \(\displaystyle 0.0\)}%
\end{pgfscope}%
\begin{pgfscope}%
\pgfsetbuttcap%
\pgfsetroundjoin%
\definecolor{currentfill}{rgb}{0.000000,0.000000,0.000000}%
\pgfsetfillcolor{currentfill}%
\pgfsetlinewidth{0.803000pt}%
\definecolor{currentstroke}{rgb}{0.000000,0.000000,0.000000}%
\pgfsetstrokecolor{currentstroke}%
\pgfsetdash{}{0pt}%
\pgfsys@defobject{currentmarker}{\pgfqpoint{-0.048611in}{0.000000in}}{\pgfqpoint{0.000000in}{0.000000in}}{%
\pgfpathmoveto{\pgfqpoint{0.000000in}{0.000000in}}%
\pgfpathlineto{\pgfqpoint{-0.048611in}{0.000000in}}%
\pgfusepath{stroke,fill}%
}%
\begin{pgfscope}%
\pgfsys@transformshift{0.375397in}{0.573164in}%
\pgfsys@useobject{currentmarker}{}%
\end{pgfscope}%
\end{pgfscope}%
\begin{pgfscope}%
\pgftext[x=0.134463in,y=0.539684in,left,base]{\rmfamily\fontsize{7.000000}{8.400000}\selectfont \(\displaystyle 0.2\)}%
\end{pgfscope}%
\begin{pgfscope}%
\pgfsetbuttcap%
\pgfsetroundjoin%
\definecolor{currentfill}{rgb}{0.000000,0.000000,0.000000}%
\pgfsetfillcolor{currentfill}%
\pgfsetlinewidth{0.803000pt}%
\definecolor{currentstroke}{rgb}{0.000000,0.000000,0.000000}%
\pgfsetstrokecolor{currentstroke}%
\pgfsetdash{}{0pt}%
\pgfsys@defobject{currentmarker}{\pgfqpoint{-0.048611in}{0.000000in}}{\pgfqpoint{0.000000in}{0.000000in}}{%
\pgfpathmoveto{\pgfqpoint{0.000000in}{0.000000in}}%
\pgfpathlineto{\pgfqpoint{-0.048611in}{0.000000in}}%
\pgfusepath{stroke,fill}%
}%
\begin{pgfscope}%
\pgfsys@transformshift{0.375397in}{0.791164in}%
\pgfsys@useobject{currentmarker}{}%
\end{pgfscope}%
\end{pgfscope}%
\begin{pgfscope}%
\pgftext[x=0.134463in,y=0.757684in,left,base]{\rmfamily\fontsize{7.000000}{8.400000}\selectfont \(\displaystyle 0.4\)}%
\end{pgfscope}%
\begin{pgfscope}%
\pgfsetbuttcap%
\pgfsetroundjoin%
\definecolor{currentfill}{rgb}{0.000000,0.000000,0.000000}%
\pgfsetfillcolor{currentfill}%
\pgfsetlinewidth{0.803000pt}%
\definecolor{currentstroke}{rgb}{0.000000,0.000000,0.000000}%
\pgfsetstrokecolor{currentstroke}%
\pgfsetdash{}{0pt}%
\pgfsys@defobject{currentmarker}{\pgfqpoint{-0.048611in}{0.000000in}}{\pgfqpoint{0.000000in}{0.000000in}}{%
\pgfpathmoveto{\pgfqpoint{0.000000in}{0.000000in}}%
\pgfpathlineto{\pgfqpoint{-0.048611in}{0.000000in}}%
\pgfusepath{stroke,fill}%
}%
\begin{pgfscope}%
\pgfsys@transformshift{0.375397in}{1.009164in}%
\pgfsys@useobject{currentmarker}{}%
\end{pgfscope}%
\end{pgfscope}%
\begin{pgfscope}%
\pgftext[x=0.134463in,y=0.975684in,left,base]{\rmfamily\fontsize{7.000000}{8.400000}\selectfont \(\displaystyle 0.6\)}%
\end{pgfscope}%
\begin{pgfscope}%
\pgfpathrectangle{\pgfqpoint{0.375397in}{0.318319in}}{\pgfqpoint{1.556284in}{0.810592in}} %
\pgfusepath{clip}%
\pgfsetrectcap%
\pgfsetroundjoin%
\pgfsetlinewidth{1.505625pt}%
\definecolor{currentstroke}{rgb}{0.317647,0.317647,0.317647}%
\pgfsetstrokecolor{currentstroke}%
\pgfsetdash{}{0pt}%
\pgfpathmoveto{\pgfqpoint{0.531026in}{0.556144in}}%
\pgfpathlineto{\pgfqpoint{0.531026in}{0.517966in}}%
\pgfusepath{stroke}%
\end{pgfscope}%
\begin{pgfscope}%
\pgfpathrectangle{\pgfqpoint{0.375397in}{0.318319in}}{\pgfqpoint{1.556284in}{0.810592in}} %
\pgfusepath{clip}%
\pgfsetrectcap%
\pgfsetroundjoin%
\pgfsetlinewidth{1.505625pt}%
\definecolor{currentstroke}{rgb}{0.317647,0.317647,0.317647}%
\pgfsetstrokecolor{currentstroke}%
\pgfsetdash{}{0pt}%
\pgfpathmoveto{\pgfqpoint{0.531026in}{0.609636in}}%
\pgfpathlineto{\pgfqpoint{0.531026in}{0.624602in}}%
\pgfusepath{stroke}%
\end{pgfscope}%
\begin{pgfscope}%
\pgfpathrectangle{\pgfqpoint{0.375397in}{0.318319in}}{\pgfqpoint{1.556284in}{0.810592in}} %
\pgfusepath{clip}%
\pgfsetrectcap%
\pgfsetroundjoin%
\pgfsetlinewidth{1.505625pt}%
\definecolor{currentstroke}{rgb}{0.317647,0.317647,0.317647}%
\pgfsetstrokecolor{currentstroke}%
\pgfsetdash{}{0pt}%
\pgfpathmoveto{\pgfqpoint{0.468774in}{0.517966in}}%
\pgfpathlineto{\pgfqpoint{0.593277in}{0.517966in}}%
\pgfusepath{stroke}%
\end{pgfscope}%
\begin{pgfscope}%
\pgfpathrectangle{\pgfqpoint{0.375397in}{0.318319in}}{\pgfqpoint{1.556284in}{0.810592in}} %
\pgfusepath{clip}%
\pgfsetrectcap%
\pgfsetroundjoin%
\pgfsetlinewidth{1.505625pt}%
\definecolor{currentstroke}{rgb}{0.317647,0.317647,0.317647}%
\pgfsetstrokecolor{currentstroke}%
\pgfsetdash{}{0pt}%
\pgfpathmoveto{\pgfqpoint{0.468774in}{0.624602in}}%
\pgfpathlineto{\pgfqpoint{0.593277in}{0.624602in}}%
\pgfusepath{stroke}%
\end{pgfscope}%
\begin{pgfscope}%
\pgfpathrectangle{\pgfqpoint{0.375397in}{0.318319in}}{\pgfqpoint{1.556284in}{0.810592in}} %
\pgfusepath{clip}%
\pgfsetrectcap%
\pgfsetroundjoin%
\pgfsetlinewidth{1.505625pt}%
\definecolor{currentstroke}{rgb}{0.317647,0.317647,0.317647}%
\pgfsetstrokecolor{currentstroke}%
\pgfsetdash{}{0pt}%
\pgfpathmoveto{\pgfqpoint{0.842282in}{0.602061in}}%
\pgfpathlineto{\pgfqpoint{0.842282in}{0.584477in}}%
\pgfusepath{stroke}%
\end{pgfscope}%
\begin{pgfscope}%
\pgfpathrectangle{\pgfqpoint{0.375397in}{0.318319in}}{\pgfqpoint{1.556284in}{0.810592in}} %
\pgfusepath{clip}%
\pgfsetrectcap%
\pgfsetroundjoin%
\pgfsetlinewidth{1.505625pt}%
\definecolor{currentstroke}{rgb}{0.317647,0.317647,0.317647}%
\pgfsetstrokecolor{currentstroke}%
\pgfsetdash{}{0pt}%
\pgfpathmoveto{\pgfqpoint{0.842282in}{0.718546in}}%
\pgfpathlineto{\pgfqpoint{0.842282in}{0.732208in}}%
\pgfusepath{stroke}%
\end{pgfscope}%
\begin{pgfscope}%
\pgfpathrectangle{\pgfqpoint{0.375397in}{0.318319in}}{\pgfqpoint{1.556284in}{0.810592in}} %
\pgfusepath{clip}%
\pgfsetrectcap%
\pgfsetroundjoin%
\pgfsetlinewidth{1.505625pt}%
\definecolor{currentstroke}{rgb}{0.317647,0.317647,0.317647}%
\pgfsetstrokecolor{currentstroke}%
\pgfsetdash{}{0pt}%
\pgfpathmoveto{\pgfqpoint{0.780031in}{0.584477in}}%
\pgfpathlineto{\pgfqpoint{0.904534in}{0.584477in}}%
\pgfusepath{stroke}%
\end{pgfscope}%
\begin{pgfscope}%
\pgfpathrectangle{\pgfqpoint{0.375397in}{0.318319in}}{\pgfqpoint{1.556284in}{0.810592in}} %
\pgfusepath{clip}%
\pgfsetrectcap%
\pgfsetroundjoin%
\pgfsetlinewidth{1.505625pt}%
\definecolor{currentstroke}{rgb}{0.317647,0.317647,0.317647}%
\pgfsetstrokecolor{currentstroke}%
\pgfsetdash{}{0pt}%
\pgfpathmoveto{\pgfqpoint{0.780031in}{0.732208in}}%
\pgfpathlineto{\pgfqpoint{0.904534in}{0.732208in}}%
\pgfusepath{stroke}%
\end{pgfscope}%
\begin{pgfscope}%
\pgfpathrectangle{\pgfqpoint{0.375397in}{0.318319in}}{\pgfqpoint{1.556284in}{0.810592in}} %
\pgfusepath{clip}%
\pgfsetbuttcap%
\pgfsetmiterjoin%
\definecolor{currentfill}{rgb}{0.317647,0.317647,0.317647}%
\pgfsetfillcolor{currentfill}%
\pgfsetlinewidth{1.003750pt}%
\definecolor{currentstroke}{rgb}{0.317647,0.317647,0.317647}%
\pgfsetstrokecolor{currentstroke}%
\pgfsetdash{}{0pt}%
\pgfsys@defobject{currentmarker}{\pgfqpoint{-0.029463in}{-0.049105in}}{\pgfqpoint{0.029463in}{0.049105in}}{%
\pgfpathmoveto{\pgfqpoint{-0.000000in}{-0.049105in}}%
\pgfpathlineto{\pgfqpoint{0.029463in}{0.000000in}}%
\pgfpathlineto{\pgfqpoint{0.000000in}{0.049105in}}%
\pgfpathlineto{\pgfqpoint{-0.029463in}{0.000000in}}%
\pgfpathclose%
\pgfusepath{stroke,fill}%
}%
\begin{pgfscope}%
\pgfsys@transformshift{0.842282in}{0.355164in}%
\pgfsys@useobject{currentmarker}{}%
\end{pgfscope}%
\begin{pgfscope}%
\pgfsys@transformshift{0.842282in}{1.092065in}%
\pgfsys@useobject{currentmarker}{}%
\end{pgfscope}%
\end{pgfscope}%
\begin{pgfscope}%
\pgfpathrectangle{\pgfqpoint{0.375397in}{0.318319in}}{\pgfqpoint{1.556284in}{0.810592in}} %
\pgfusepath{clip}%
\pgfsetrectcap%
\pgfsetroundjoin%
\pgfsetlinewidth{1.505625pt}%
\definecolor{currentstroke}{rgb}{0.317647,0.317647,0.317647}%
\pgfsetstrokecolor{currentstroke}%
\pgfsetdash{}{0pt}%
\pgfpathmoveto{\pgfqpoint{1.153539in}{0.667220in}}%
\pgfpathlineto{\pgfqpoint{1.153539in}{0.589721in}}%
\pgfusepath{stroke}%
\end{pgfscope}%
\begin{pgfscope}%
\pgfpathrectangle{\pgfqpoint{0.375397in}{0.318319in}}{\pgfqpoint{1.556284in}{0.810592in}} %
\pgfusepath{clip}%
\pgfsetrectcap%
\pgfsetroundjoin%
\pgfsetlinewidth{1.505625pt}%
\definecolor{currentstroke}{rgb}{0.317647,0.317647,0.317647}%
\pgfsetstrokecolor{currentstroke}%
\pgfsetdash{}{0pt}%
\pgfpathmoveto{\pgfqpoint{1.153539in}{0.800817in}}%
\pgfpathlineto{\pgfqpoint{1.153539in}{0.923355in}}%
\pgfusepath{stroke}%
\end{pgfscope}%
\begin{pgfscope}%
\pgfpathrectangle{\pgfqpoint{0.375397in}{0.318319in}}{\pgfqpoint{1.556284in}{0.810592in}} %
\pgfusepath{clip}%
\pgfsetrectcap%
\pgfsetroundjoin%
\pgfsetlinewidth{1.505625pt}%
\definecolor{currentstroke}{rgb}{0.317647,0.317647,0.317647}%
\pgfsetstrokecolor{currentstroke}%
\pgfsetdash{}{0pt}%
\pgfpathmoveto{\pgfqpoint{1.091288in}{0.589721in}}%
\pgfpathlineto{\pgfqpoint{1.215791in}{0.589721in}}%
\pgfusepath{stroke}%
\end{pgfscope}%
\begin{pgfscope}%
\pgfpathrectangle{\pgfqpoint{0.375397in}{0.318319in}}{\pgfqpoint{1.556284in}{0.810592in}} %
\pgfusepath{clip}%
\pgfsetrectcap%
\pgfsetroundjoin%
\pgfsetlinewidth{1.505625pt}%
\definecolor{currentstroke}{rgb}{0.317647,0.317647,0.317647}%
\pgfsetstrokecolor{currentstroke}%
\pgfsetdash{}{0pt}%
\pgfpathmoveto{\pgfqpoint{1.091288in}{0.923355in}}%
\pgfpathlineto{\pgfqpoint{1.215791in}{0.923355in}}%
\pgfusepath{stroke}%
\end{pgfscope}%
\begin{pgfscope}%
\pgfpathrectangle{\pgfqpoint{0.375397in}{0.318319in}}{\pgfqpoint{1.556284in}{0.810592in}} %
\pgfusepath{clip}%
\pgfsetrectcap%
\pgfsetroundjoin%
\pgfsetlinewidth{1.505625pt}%
\definecolor{currentstroke}{rgb}{0.317647,0.317647,0.317647}%
\pgfsetstrokecolor{currentstroke}%
\pgfsetdash{}{0pt}%
\pgfpathmoveto{\pgfqpoint{1.464796in}{0.585998in}}%
\pgfpathlineto{\pgfqpoint{1.464796in}{0.582247in}}%
\pgfusepath{stroke}%
\end{pgfscope}%
\begin{pgfscope}%
\pgfpathrectangle{\pgfqpoint{0.375397in}{0.318319in}}{\pgfqpoint{1.556284in}{0.810592in}} %
\pgfusepath{clip}%
\pgfsetrectcap%
\pgfsetroundjoin%
\pgfsetlinewidth{1.505625pt}%
\definecolor{currentstroke}{rgb}{0.317647,0.317647,0.317647}%
\pgfsetstrokecolor{currentstroke}%
\pgfsetdash{}{0pt}%
\pgfpathmoveto{\pgfqpoint{1.464796in}{0.687648in}}%
\pgfpathlineto{\pgfqpoint{1.464796in}{0.718497in}}%
\pgfusepath{stroke}%
\end{pgfscope}%
\begin{pgfscope}%
\pgfpathrectangle{\pgfqpoint{0.375397in}{0.318319in}}{\pgfqpoint{1.556284in}{0.810592in}} %
\pgfusepath{clip}%
\pgfsetrectcap%
\pgfsetroundjoin%
\pgfsetlinewidth{1.505625pt}%
\definecolor{currentstroke}{rgb}{0.317647,0.317647,0.317647}%
\pgfsetstrokecolor{currentstroke}%
\pgfsetdash{}{0pt}%
\pgfpathmoveto{\pgfqpoint{1.402545in}{0.582247in}}%
\pgfpathlineto{\pgfqpoint{1.527047in}{0.582247in}}%
\pgfusepath{stroke}%
\end{pgfscope}%
\begin{pgfscope}%
\pgfpathrectangle{\pgfqpoint{0.375397in}{0.318319in}}{\pgfqpoint{1.556284in}{0.810592in}} %
\pgfusepath{clip}%
\pgfsetrectcap%
\pgfsetroundjoin%
\pgfsetlinewidth{1.505625pt}%
\definecolor{currentstroke}{rgb}{0.317647,0.317647,0.317647}%
\pgfsetstrokecolor{currentstroke}%
\pgfsetdash{}{0pt}%
\pgfpathmoveto{\pgfqpoint{1.402545in}{0.718497in}}%
\pgfpathlineto{\pgfqpoint{1.527047in}{0.718497in}}%
\pgfusepath{stroke}%
\end{pgfscope}%
\begin{pgfscope}%
\pgfpathrectangle{\pgfqpoint{0.375397in}{0.318319in}}{\pgfqpoint{1.556284in}{0.810592in}} %
\pgfusepath{clip}%
\pgfsetbuttcap%
\pgfsetmiterjoin%
\definecolor{currentfill}{rgb}{0.317647,0.317647,0.317647}%
\pgfsetfillcolor{currentfill}%
\pgfsetlinewidth{1.003750pt}%
\definecolor{currentstroke}{rgb}{0.317647,0.317647,0.317647}%
\pgfsetstrokecolor{currentstroke}%
\pgfsetdash{}{0pt}%
\pgfsys@defobject{currentmarker}{\pgfqpoint{-0.029463in}{-0.049105in}}{\pgfqpoint{0.029463in}{0.049105in}}{%
\pgfpathmoveto{\pgfqpoint{-0.000000in}{-0.049105in}}%
\pgfpathlineto{\pgfqpoint{0.029463in}{0.000000in}}%
\pgfpathlineto{\pgfqpoint{0.000000in}{0.049105in}}%
\pgfpathlineto{\pgfqpoint{-0.029463in}{0.000000in}}%
\pgfpathclose%
\pgfusepath{stroke,fill}%
}%
\begin{pgfscope}%
\pgfsys@transformshift{1.464796in}{0.851175in}%
\pgfsys@useobject{currentmarker}{}%
\end{pgfscope}%
\end{pgfscope}%
\begin{pgfscope}%
\pgfpathrectangle{\pgfqpoint{0.375397in}{0.318319in}}{\pgfqpoint{1.556284in}{0.810592in}} %
\pgfusepath{clip}%
\pgfsetrectcap%
\pgfsetroundjoin%
\pgfsetlinewidth{1.505625pt}%
\definecolor{currentstroke}{rgb}{0.317647,0.317647,0.317647}%
\pgfsetstrokecolor{currentstroke}%
\pgfsetdash{}{0pt}%
\pgfpathmoveto{\pgfqpoint{1.776053in}{0.587854in}}%
\pgfpathlineto{\pgfqpoint{1.776053in}{0.498951in}}%
\pgfusepath{stroke}%
\end{pgfscope}%
\begin{pgfscope}%
\pgfpathrectangle{\pgfqpoint{0.375397in}{0.318319in}}{\pgfqpoint{1.556284in}{0.810592in}} %
\pgfusepath{clip}%
\pgfsetrectcap%
\pgfsetroundjoin%
\pgfsetlinewidth{1.505625pt}%
\definecolor{currentstroke}{rgb}{0.317647,0.317647,0.317647}%
\pgfsetstrokecolor{currentstroke}%
\pgfsetdash{}{0pt}%
\pgfpathmoveto{\pgfqpoint{1.776053in}{0.690263in}}%
\pgfpathlineto{\pgfqpoint{1.776053in}{0.692910in}}%
\pgfusepath{stroke}%
\end{pgfscope}%
\begin{pgfscope}%
\pgfpathrectangle{\pgfqpoint{0.375397in}{0.318319in}}{\pgfqpoint{1.556284in}{0.810592in}} %
\pgfusepath{clip}%
\pgfsetrectcap%
\pgfsetroundjoin%
\pgfsetlinewidth{1.505625pt}%
\definecolor{currentstroke}{rgb}{0.317647,0.317647,0.317647}%
\pgfsetstrokecolor{currentstroke}%
\pgfsetdash{}{0pt}%
\pgfpathmoveto{\pgfqpoint{1.713801in}{0.498951in}}%
\pgfpathlineto{\pgfqpoint{1.838304in}{0.498951in}}%
\pgfusepath{stroke}%
\end{pgfscope}%
\begin{pgfscope}%
\pgfpathrectangle{\pgfqpoint{0.375397in}{0.318319in}}{\pgfqpoint{1.556284in}{0.810592in}} %
\pgfusepath{clip}%
\pgfsetrectcap%
\pgfsetroundjoin%
\pgfsetlinewidth{1.505625pt}%
\definecolor{currentstroke}{rgb}{0.317647,0.317647,0.317647}%
\pgfsetstrokecolor{currentstroke}%
\pgfsetdash{}{0pt}%
\pgfpathmoveto{\pgfqpoint{1.713801in}{0.692910in}}%
\pgfpathlineto{\pgfqpoint{1.838304in}{0.692910in}}%
\pgfusepath{stroke}%
\end{pgfscope}%
\begin{pgfscope}%
\pgfpathrectangle{\pgfqpoint{0.375397in}{0.318319in}}{\pgfqpoint{1.556284in}{0.810592in}} %
\pgfusepath{clip}%
\pgfsetbuttcap%
\pgfsetmiterjoin%
\definecolor{currentfill}{rgb}{0.317647,0.317647,0.317647}%
\pgfsetfillcolor{currentfill}%
\pgfsetlinewidth{1.003750pt}%
\definecolor{currentstroke}{rgb}{0.317647,0.317647,0.317647}%
\pgfsetstrokecolor{currentstroke}%
\pgfsetdash{}{0pt}%
\pgfsys@defobject{currentmarker}{\pgfqpoint{-0.029463in}{-0.049105in}}{\pgfqpoint{0.029463in}{0.049105in}}{%
\pgfpathmoveto{\pgfqpoint{-0.000000in}{-0.049105in}}%
\pgfpathlineto{\pgfqpoint{0.029463in}{0.000000in}}%
\pgfpathlineto{\pgfqpoint{0.000000in}{0.049105in}}%
\pgfpathlineto{\pgfqpoint{-0.029463in}{0.000000in}}%
\pgfpathclose%
\pgfusepath{stroke,fill}%
}%
\begin{pgfscope}%
\pgfsys@transformshift{1.776053in}{0.892488in}%
\pgfsys@useobject{currentmarker}{}%
\end{pgfscope}%
\end{pgfscope}%
\begin{pgfscope}%
\pgfpathrectangle{\pgfqpoint{0.375397in}{0.318319in}}{\pgfqpoint{1.556284in}{0.810592in}} %
\pgfusepath{clip}%
\pgfsetrectcap%
\pgfsetroundjoin%
\pgfsetlinewidth{1.505625pt}%
\definecolor{currentstroke}{rgb}{0.317647,0.317647,0.317647}%
\pgfsetstrokecolor{currentstroke}%
\pgfsetdash{}{0pt}%
\pgfpathmoveto{\pgfqpoint{0.406523in}{0.575431in}}%
\pgfpathlineto{\pgfqpoint{0.655528in}{0.575431in}}%
\pgfusepath{stroke}%
\end{pgfscope}%
\begin{pgfscope}%
\pgfpathrectangle{\pgfqpoint{0.375397in}{0.318319in}}{\pgfqpoint{1.556284in}{0.810592in}} %
\pgfusepath{clip}%
\pgfsetrectcap%
\pgfsetroundjoin%
\pgfsetlinewidth{1.505625pt}%
\definecolor{currentstroke}{rgb}{0.317647,0.317647,0.317647}%
\pgfsetstrokecolor{currentstroke}%
\pgfsetdash{}{0pt}%
\pgfpathmoveto{\pgfqpoint{0.717780in}{0.666186in}}%
\pgfpathlineto{\pgfqpoint{0.966785in}{0.666186in}}%
\pgfusepath{stroke}%
\end{pgfscope}%
\begin{pgfscope}%
\pgfpathrectangle{\pgfqpoint{0.375397in}{0.318319in}}{\pgfqpoint{1.556284in}{0.810592in}} %
\pgfusepath{clip}%
\pgfsetrectcap%
\pgfsetroundjoin%
\pgfsetlinewidth{1.505625pt}%
\definecolor{currentstroke}{rgb}{0.317647,0.317647,0.317647}%
\pgfsetstrokecolor{currentstroke}%
\pgfsetdash{}{0pt}%
\pgfpathmoveto{\pgfqpoint{1.029037in}{0.723395in}}%
\pgfpathlineto{\pgfqpoint{1.278042in}{0.723395in}}%
\pgfusepath{stroke}%
\end{pgfscope}%
\begin{pgfscope}%
\pgfpathrectangle{\pgfqpoint{0.375397in}{0.318319in}}{\pgfqpoint{1.556284in}{0.810592in}} %
\pgfusepath{clip}%
\pgfsetrectcap%
\pgfsetroundjoin%
\pgfsetlinewidth{1.505625pt}%
\definecolor{currentstroke}{rgb}{0.317647,0.317647,0.317647}%
\pgfsetstrokecolor{currentstroke}%
\pgfsetdash{}{0pt}%
\pgfpathmoveto{\pgfqpoint{1.340293in}{0.591918in}}%
\pgfpathlineto{\pgfqpoint{1.589299in}{0.591918in}}%
\pgfusepath{stroke}%
\end{pgfscope}%
\begin{pgfscope}%
\pgfpathrectangle{\pgfqpoint{0.375397in}{0.318319in}}{\pgfqpoint{1.556284in}{0.810592in}} %
\pgfusepath{clip}%
\pgfsetrectcap%
\pgfsetroundjoin%
\pgfsetlinewidth{1.505625pt}%
\definecolor{currentstroke}{rgb}{0.317647,0.317647,0.317647}%
\pgfsetstrokecolor{currentstroke}%
\pgfsetdash{}{0pt}%
\pgfpathmoveto{\pgfqpoint{1.651550in}{0.639238in}}%
\pgfpathlineto{\pgfqpoint{1.900556in}{0.639238in}}%
\pgfusepath{stroke}%
\end{pgfscope}%
\begin{pgfscope}%
\pgfsetrectcap%
\pgfsetmiterjoin%
\pgfsetlinewidth{0.803000pt}%
\definecolor{currentstroke}{rgb}{0.000000,0.000000,0.000000}%
\pgfsetstrokecolor{currentstroke}%
\pgfsetdash{}{0pt}%
\pgfpathmoveto{\pgfqpoint{0.375397in}{0.318319in}}%
\pgfpathlineto{\pgfqpoint{0.375397in}{1.128910in}}%
\pgfusepath{stroke}%
\end{pgfscope}%
\begin{pgfscope}%
\pgfsetrectcap%
\pgfsetmiterjoin%
\pgfsetlinewidth{0.803000pt}%
\definecolor{currentstroke}{rgb}{0.000000,0.000000,0.000000}%
\pgfsetstrokecolor{currentstroke}%
\pgfsetdash{}{0pt}%
\pgfpathmoveto{\pgfqpoint{1.931681in}{0.318319in}}%
\pgfpathlineto{\pgfqpoint{1.931681in}{1.128910in}}%
\pgfusepath{stroke}%
\end{pgfscope}%
\begin{pgfscope}%
\pgfsetrectcap%
\pgfsetmiterjoin%
\pgfsetlinewidth{0.803000pt}%
\definecolor{currentstroke}{rgb}{0.000000,0.000000,0.000000}%
\pgfsetstrokecolor{currentstroke}%
\pgfsetdash{}{0pt}%
\pgfpathmoveto{\pgfqpoint{0.375397in}{0.318319in}}%
\pgfpathlineto{\pgfqpoint{1.931681in}{0.318319in}}%
\pgfusepath{stroke}%
\end{pgfscope}%
\begin{pgfscope}%
\pgfsetrectcap%
\pgfsetmiterjoin%
\pgfsetlinewidth{0.803000pt}%
\definecolor{currentstroke}{rgb}{0.000000,0.000000,0.000000}%
\pgfsetstrokecolor{currentstroke}%
\pgfsetdash{}{0pt}%
\pgfpathmoveto{\pgfqpoint{0.375397in}{1.128910in}}%
\pgfpathlineto{\pgfqpoint{1.931681in}{1.128910in}}%
\pgfusepath{stroke}%
\end{pgfscope}%
\end{pgfpicture}%
\makeatother%
\endgroup%

%% file: figures/reinforcement/train_test_stats.pgf
\begingroup%
\makeatletter%
\begin{pgfpicture}%
\pgfpathrectangle{\pgfpointorigin}{\pgfqpoint{2.101681in}{1.298910in}}%
\pgfusepath{use as bounding box, clip}%
\begin{pgfscope}%
\pgfsetbuttcap%
\pgfsetmiterjoin%
\pgfsetlinewidth{0.000000pt}%
\definecolor{currentstroke}{rgb}{1.000000,1.000000,1.000000}%
\pgfsetstrokecolor{currentstroke}%
\pgfsetstrokeopacity{0.000000}%
\pgfsetdash{}{0pt}%
\pgfpathmoveto{\pgfqpoint{0.000000in}{0.000000in}}%
\pgfpathlineto{\pgfqpoint{2.101681in}{0.000000in}}%
\pgfpathlineto{\pgfqpoint{2.101681in}{1.298910in}}%
\pgfpathlineto{\pgfqpoint{0.000000in}{1.298910in}}%
\pgfpathclose%
\pgfusepath{}%
\end{pgfscope}%
\begin{pgfscope}%
\pgfsetbuttcap%
\pgfsetmiterjoin%
\definecolor{currentfill}{rgb}{1.000000,1.000000,1.000000}%
\pgfsetfillcolor{currentfill}%
\pgfsetlinewidth{0.000000pt}%
\definecolor{currentstroke}{rgb}{0.000000,0.000000,0.000000}%
\pgfsetstrokecolor{currentstroke}%
\pgfsetstrokeopacity{0.000000}%
\pgfsetdash{}{0pt}%
\pgfpathmoveto{\pgfqpoint{0.375397in}{0.318319in}}%
\pgfpathlineto{\pgfqpoint{1.931681in}{0.318319in}}%
\pgfpathlineto{\pgfqpoint{1.931681in}{1.128910in}}%
\pgfpathlineto{\pgfqpoint{0.375397in}{1.128910in}}%
\pgfpathclose%
\pgfusepath{fill}%
\end{pgfscope}%
\begin{pgfscope}%
\pgfpathrectangle{\pgfqpoint{0.375397in}{0.318319in}}{\pgfqpoint{1.556284in}{0.810592in}} %
\pgfusepath{clip}%
\pgfsetbuttcap%
\pgfsetmiterjoin%
\definecolor{currentfill}{rgb}{0.229412,0.570588,0.229412}%
\pgfsetfillcolor{currentfill}%
\pgfsetlinewidth{1.505625pt}%
\definecolor{currentstroke}{rgb}{0.239216,0.239216,0.239216}%
\pgfsetstrokecolor{currentstroke}%
\pgfsetdash{}{0pt}%
\pgfpathmoveto{\pgfqpoint{0.406523in}{0.355164in}}%
\pgfpathlineto{\pgfqpoint{0.655528in}{0.355164in}}%
\pgfpathlineto{\pgfqpoint{0.655528in}{0.721808in}}%
\pgfpathlineto{\pgfqpoint{0.406523in}{0.721808in}}%
\pgfpathlineto{\pgfqpoint{0.406523in}{0.355164in}}%
\pgfpathclose%
\pgfusepath{stroke,fill}%
\end{pgfscope}%
\begin{pgfscope}%
\pgfpathrectangle{\pgfqpoint{0.375397in}{0.318319in}}{\pgfqpoint{1.556284in}{0.810592in}} %
\pgfusepath{clip}%
\pgfsetbuttcap%
\pgfsetmiterjoin%
\definecolor{currentfill}{rgb}{0.229412,0.570588,0.229412}%
\pgfsetfillcolor{currentfill}%
\pgfsetlinewidth{1.505625pt}%
\definecolor{currentstroke}{rgb}{0.239216,0.239216,0.239216}%
\pgfsetstrokecolor{currentstroke}%
\pgfsetdash{}{0pt}%
\pgfpathmoveto{\pgfqpoint{0.717780in}{0.355164in}}%
\pgfpathlineto{\pgfqpoint{0.966785in}{0.355164in}}%
\pgfpathlineto{\pgfqpoint{0.966785in}{0.663633in}}%
\pgfpathlineto{\pgfqpoint{0.717780in}{0.663633in}}%
\pgfpathlineto{\pgfqpoint{0.717780in}{0.355164in}}%
\pgfpathclose%
\pgfusepath{stroke,fill}%
\end{pgfscope}%
\begin{pgfscope}%
\pgfpathrectangle{\pgfqpoint{0.375397in}{0.318319in}}{\pgfqpoint{1.556284in}{0.810592in}} %
\pgfusepath{clip}%
\pgfsetbuttcap%
\pgfsetmiterjoin%
\definecolor{currentfill}{rgb}{0.229412,0.570588,0.229412}%
\pgfsetfillcolor{currentfill}%
\pgfsetlinewidth{1.505625pt}%
\definecolor{currentstroke}{rgb}{0.239216,0.239216,0.239216}%
\pgfsetstrokecolor{currentstroke}%
\pgfsetdash{}{0pt}%
\pgfpathmoveto{\pgfqpoint{1.029037in}{0.355164in}}%
\pgfpathlineto{\pgfqpoint{1.278042in}{0.355164in}}%
\pgfpathlineto{\pgfqpoint{1.278042in}{0.725711in}}%
\pgfpathlineto{\pgfqpoint{1.029037in}{0.725711in}}%
\pgfpathlineto{\pgfqpoint{1.029037in}{0.355164in}}%
\pgfpathclose%
\pgfusepath{stroke,fill}%
\end{pgfscope}%
\begin{pgfscope}%
\pgfpathrectangle{\pgfqpoint{0.375397in}{0.318319in}}{\pgfqpoint{1.556284in}{0.810592in}} %
\pgfusepath{clip}%
\pgfsetbuttcap%
\pgfsetmiterjoin%
\definecolor{currentfill}{rgb}{0.229412,0.570588,0.229412}%
\pgfsetfillcolor{currentfill}%
\pgfsetlinewidth{1.505625pt}%
\definecolor{currentstroke}{rgb}{0.239216,0.239216,0.239216}%
\pgfsetstrokecolor{currentstroke}%
\pgfsetdash{}{0pt}%
\pgfpathmoveto{\pgfqpoint{1.340293in}{0.561022in}}%
\pgfpathlineto{\pgfqpoint{1.589299in}{0.561022in}}%
\pgfpathlineto{\pgfqpoint{1.589299in}{0.693276in}}%
\pgfpathlineto{\pgfqpoint{1.340293in}{0.693276in}}%
\pgfpathlineto{\pgfqpoint{1.340293in}{0.561022in}}%
\pgfpathclose%
\pgfusepath{stroke,fill}%
\end{pgfscope}%
\begin{pgfscope}%
\pgfpathrectangle{\pgfqpoint{0.375397in}{0.318319in}}{\pgfqpoint{1.556284in}{0.810592in}} %
\pgfusepath{clip}%
\pgfsetbuttcap%
\pgfsetmiterjoin%
\definecolor{currentfill}{rgb}{0.229412,0.570588,0.229412}%
\pgfsetfillcolor{currentfill}%
\pgfsetlinewidth{1.505625pt}%
\definecolor{currentstroke}{rgb}{0.239216,0.239216,0.239216}%
\pgfsetstrokecolor{currentstroke}%
\pgfsetdash{}{0pt}%
\pgfpathmoveto{\pgfqpoint{1.651550in}{0.355164in}}%
\pgfpathlineto{\pgfqpoint{1.900556in}{0.355164in}}%
\pgfpathlineto{\pgfqpoint{1.900556in}{0.671448in}}%
\pgfpathlineto{\pgfqpoint{1.651550in}{0.671448in}}%
\pgfpathlineto{\pgfqpoint{1.651550in}{0.355164in}}%
\pgfpathclose%
\pgfusepath{stroke,fill}%
\end{pgfscope}%
\begin{pgfscope}%
\pgfsetbuttcap%
\pgfsetroundjoin%
\definecolor{currentfill}{rgb}{0.000000,0.000000,0.000000}%
\pgfsetfillcolor{currentfill}%
\pgfsetlinewidth{0.803000pt}%
\definecolor{currentstroke}{rgb}{0.000000,0.000000,0.000000}%
\pgfsetstrokecolor{currentstroke}%
\pgfsetdash{}{0pt}%
\pgfsys@defobject{currentmarker}{\pgfqpoint{0.000000in}{-0.048611in}}{\pgfqpoint{0.000000in}{0.000000in}}{%
\pgfpathmoveto{\pgfqpoint{0.000000in}{0.000000in}}%
\pgfpathlineto{\pgfqpoint{0.000000in}{-0.048611in}}%
\pgfusepath{stroke,fill}%
}%
\begin{pgfscope}%
\pgfsys@transformshift{0.531026in}{0.318319in}%
\pgfsys@useobject{currentmarker}{}%
\end{pgfscope}%
\end{pgfscope}%
\begin{pgfscope}%
\pgftext[x=0.531026in,y=0.221097in,,top]{\rmfamily\fontsize{7.000000}{8.400000}\selectfont 0}%
\end{pgfscope}%
\begin{pgfscope}%
\pgfsetbuttcap%
\pgfsetroundjoin%
\definecolor{currentfill}{rgb}{0.000000,0.000000,0.000000}%
\pgfsetfillcolor{currentfill}%
\pgfsetlinewidth{0.803000pt}%
\definecolor{currentstroke}{rgb}{0.000000,0.000000,0.000000}%
\pgfsetstrokecolor{currentstroke}%
\pgfsetdash{}{0pt}%
\pgfsys@defobject{currentmarker}{\pgfqpoint{0.000000in}{-0.048611in}}{\pgfqpoint{0.000000in}{0.000000in}}{%
\pgfpathmoveto{\pgfqpoint{0.000000in}{0.000000in}}%
\pgfpathlineto{\pgfqpoint{0.000000in}{-0.048611in}}%
\pgfusepath{stroke,fill}%
}%
\begin{pgfscope}%
\pgfsys@transformshift{0.842282in}{0.318319in}%
\pgfsys@useobject{currentmarker}{}%
\end{pgfscope}%
\end{pgfscope}%
\begin{pgfscope}%
\pgftext[x=0.842282in,y=0.221097in,,top]{\rmfamily\fontsize{7.000000}{8.400000}\selectfont 1}%
\end{pgfscope}%
\begin{pgfscope}%
\pgfsetbuttcap%
\pgfsetroundjoin%
\definecolor{currentfill}{rgb}{0.000000,0.000000,0.000000}%
\pgfsetfillcolor{currentfill}%
\pgfsetlinewidth{0.803000pt}%
\definecolor{currentstroke}{rgb}{0.000000,0.000000,0.000000}%
\pgfsetstrokecolor{currentstroke}%
\pgfsetdash{}{0pt}%
\pgfsys@defobject{currentmarker}{\pgfqpoint{0.000000in}{-0.048611in}}{\pgfqpoint{0.000000in}{0.000000in}}{%
\pgfpathmoveto{\pgfqpoint{0.000000in}{0.000000in}}%
\pgfpathlineto{\pgfqpoint{0.000000in}{-0.048611in}}%
\pgfusepath{stroke,fill}%
}%
\begin{pgfscope}%
\pgfsys@transformshift{1.153539in}{0.318319in}%
\pgfsys@useobject{currentmarker}{}%
\end{pgfscope}%
\end{pgfscope}%
\begin{pgfscope}%
\pgftext[x=1.153539in,y=0.221097in,,top]{\rmfamily\fontsize{7.000000}{8.400000}\selectfont 2}%
\end{pgfscope}%
\begin{pgfscope}%
\pgfsetbuttcap%
\pgfsetroundjoin%
\definecolor{currentfill}{rgb}{0.000000,0.000000,0.000000}%
\pgfsetfillcolor{currentfill}%
\pgfsetlinewidth{0.803000pt}%
\definecolor{currentstroke}{rgb}{0.000000,0.000000,0.000000}%
\pgfsetstrokecolor{currentstroke}%
\pgfsetdash{}{0pt}%
\pgfsys@defobject{currentmarker}{\pgfqpoint{0.000000in}{-0.048611in}}{\pgfqpoint{0.000000in}{0.000000in}}{%
\pgfpathmoveto{\pgfqpoint{0.000000in}{0.000000in}}%
\pgfpathlineto{\pgfqpoint{0.000000in}{-0.048611in}}%
\pgfusepath{stroke,fill}%
}%
\begin{pgfscope}%
\pgfsys@transformshift{1.464796in}{0.318319in}%
\pgfsys@useobject{currentmarker}{}%
\end{pgfscope}%
\end{pgfscope}%
\begin{pgfscope}%
\pgftext[x=1.464796in,y=0.221097in,,top]{\rmfamily\fontsize{7.000000}{8.400000}\selectfont 3}%
\end{pgfscope}%
\begin{pgfscope}%
\pgfsetbuttcap%
\pgfsetroundjoin%
\definecolor{currentfill}{rgb}{0.000000,0.000000,0.000000}%
\pgfsetfillcolor{currentfill}%
\pgfsetlinewidth{0.803000pt}%
\definecolor{currentstroke}{rgb}{0.000000,0.000000,0.000000}%
\pgfsetstrokecolor{currentstroke}%
\pgfsetdash{}{0pt}%
\pgfsys@defobject{currentmarker}{\pgfqpoint{0.000000in}{-0.048611in}}{\pgfqpoint{0.000000in}{0.000000in}}{%
\pgfpathmoveto{\pgfqpoint{0.000000in}{0.000000in}}%
\pgfpathlineto{\pgfqpoint{0.000000in}{-0.048611in}}%
\pgfusepath{stroke,fill}%
}%
\begin{pgfscope}%
\pgfsys@transformshift{1.776053in}{0.318319in}%
\pgfsys@useobject{currentmarker}{}%
\end{pgfscope}%
\end{pgfscope}%
\begin{pgfscope}%
\pgftext[x=1.776053in,y=0.221097in,,top]{\rmfamily\fontsize{7.000000}{8.400000}\selectfont 4}%
\end{pgfscope}%
\begin{pgfscope}%
\pgfsetbuttcap%
\pgfsetroundjoin%
\definecolor{currentfill}{rgb}{0.000000,0.000000,0.000000}%
\pgfsetfillcolor{currentfill}%
\pgfsetlinewidth{0.803000pt}%
\definecolor{currentstroke}{rgb}{0.000000,0.000000,0.000000}%
\pgfsetstrokecolor{currentstroke}%
\pgfsetdash{}{0pt}%
\pgfsys@defobject{currentmarker}{\pgfqpoint{-0.048611in}{0.000000in}}{\pgfqpoint{0.000000in}{0.000000in}}{%
\pgfpathmoveto{\pgfqpoint{0.000000in}{0.000000in}}%
\pgfpathlineto{\pgfqpoint{-0.048611in}{0.000000in}}%
\pgfusepath{stroke,fill}%
}%
\begin{pgfscope}%
\pgfsys@transformshift{0.375397in}{0.355164in}%
\pgfsys@useobject{currentmarker}{}%
\end{pgfscope}%
\end{pgfscope}%
\begin{pgfscope}%
\pgftext[x=0.134463in,y=0.321684in,left,base]{\rmfamily\fontsize{7.000000}{8.400000}\selectfont \(\displaystyle 0.0\)}%
\end{pgfscope}%
\begin{pgfscope}%
\pgfsetbuttcap%
\pgfsetroundjoin%
\definecolor{currentfill}{rgb}{0.000000,0.000000,0.000000}%
\pgfsetfillcolor{currentfill}%
\pgfsetlinewidth{0.803000pt}%
\definecolor{currentstroke}{rgb}{0.000000,0.000000,0.000000}%
\pgfsetstrokecolor{currentstroke}%
\pgfsetdash{}{0pt}%
\pgfsys@defobject{currentmarker}{\pgfqpoint{-0.048611in}{0.000000in}}{\pgfqpoint{0.000000in}{0.000000in}}{%
\pgfpathmoveto{\pgfqpoint{0.000000in}{0.000000in}}%
\pgfpathlineto{\pgfqpoint{-0.048611in}{0.000000in}}%
\pgfusepath{stroke,fill}%
}%
\begin{pgfscope}%
\pgfsys@transformshift{0.375397in}{0.609268in}%
\pgfsys@useobject{currentmarker}{}%
\end{pgfscope}%
\end{pgfscope}%
\begin{pgfscope}%
\pgftext[x=0.134463in,y=0.575788in,left,base]{\rmfamily\fontsize{7.000000}{8.400000}\selectfont \(\displaystyle 0.2\)}%
\end{pgfscope}%
\begin{pgfscope}%
\pgfsetbuttcap%
\pgfsetroundjoin%
\definecolor{currentfill}{rgb}{0.000000,0.000000,0.000000}%
\pgfsetfillcolor{currentfill}%
\pgfsetlinewidth{0.803000pt}%
\definecolor{currentstroke}{rgb}{0.000000,0.000000,0.000000}%
\pgfsetstrokecolor{currentstroke}%
\pgfsetdash{}{0pt}%
\pgfsys@defobject{currentmarker}{\pgfqpoint{-0.048611in}{0.000000in}}{\pgfqpoint{0.000000in}{0.000000in}}{%
\pgfpathmoveto{\pgfqpoint{0.000000in}{0.000000in}}%
\pgfpathlineto{\pgfqpoint{-0.048611in}{0.000000in}}%
\pgfusepath{stroke,fill}%
}%
\begin{pgfscope}%
\pgfsys@transformshift{0.375397in}{0.863372in}%
\pgfsys@useobject{currentmarker}{}%
\end{pgfscope}%
\end{pgfscope}%
\begin{pgfscope}%
\pgftext[x=0.134463in,y=0.829892in,left,base]{\rmfamily\fontsize{7.000000}{8.400000}\selectfont \(\displaystyle 0.4\)}%
\end{pgfscope}%
\begin{pgfscope}%
\pgfsetbuttcap%
\pgfsetroundjoin%
\definecolor{currentfill}{rgb}{0.000000,0.000000,0.000000}%
\pgfsetfillcolor{currentfill}%
\pgfsetlinewidth{0.803000pt}%
\definecolor{currentstroke}{rgb}{0.000000,0.000000,0.000000}%
\pgfsetstrokecolor{currentstroke}%
\pgfsetdash{}{0pt}%
\pgfsys@defobject{currentmarker}{\pgfqpoint{-0.048611in}{0.000000in}}{\pgfqpoint{0.000000in}{0.000000in}}{%
\pgfpathmoveto{\pgfqpoint{0.000000in}{0.000000in}}%
\pgfpathlineto{\pgfqpoint{-0.048611in}{0.000000in}}%
\pgfusepath{stroke,fill}%
}%
\begin{pgfscope}%
\pgfsys@transformshift{0.375397in}{1.117476in}%
\pgfsys@useobject{currentmarker}{}%
\end{pgfscope}%
\end{pgfscope}%
\begin{pgfscope}%
\pgftext[x=0.134463in,y=1.083996in,left,base]{\rmfamily\fontsize{7.000000}{8.400000}\selectfont \(\displaystyle 0.6\)}%
\end{pgfscope}%
\begin{pgfscope}%
\pgfpathrectangle{\pgfqpoint{0.375397in}{0.318319in}}{\pgfqpoint{1.556284in}{0.810592in}} %
\pgfusepath{clip}%
\pgfsetrectcap%
\pgfsetroundjoin%
\pgfsetlinewidth{1.505625pt}%
\definecolor{currentstroke}{rgb}{0.239216,0.239216,0.239216}%
\pgfsetstrokecolor{currentstroke}%
\pgfsetdash{}{0pt}%
\pgfpathmoveto{\pgfqpoint{0.531026in}{0.355164in}}%
\pgfpathlineto{\pgfqpoint{0.531026in}{0.355164in}}%
\pgfusepath{stroke}%
\end{pgfscope}%
\begin{pgfscope}%
\pgfpathrectangle{\pgfqpoint{0.375397in}{0.318319in}}{\pgfqpoint{1.556284in}{0.810592in}} %
\pgfusepath{clip}%
\pgfsetrectcap%
\pgfsetroundjoin%
\pgfsetlinewidth{1.505625pt}%
\definecolor{currentstroke}{rgb}{0.239216,0.239216,0.239216}%
\pgfsetstrokecolor{currentstroke}%
\pgfsetdash{}{0pt}%
\pgfpathmoveto{\pgfqpoint{0.531026in}{0.721808in}}%
\pgfpathlineto{\pgfqpoint{0.531026in}{1.092065in}}%
\pgfusepath{stroke}%
\end{pgfscope}%
\begin{pgfscope}%
\pgfpathrectangle{\pgfqpoint{0.375397in}{0.318319in}}{\pgfqpoint{1.556284in}{0.810592in}} %
\pgfusepath{clip}%
\pgfsetrectcap%
\pgfsetroundjoin%
\pgfsetlinewidth{1.505625pt}%
\definecolor{currentstroke}{rgb}{0.239216,0.239216,0.239216}%
\pgfsetstrokecolor{currentstroke}%
\pgfsetdash{}{0pt}%
\pgfpathmoveto{\pgfqpoint{0.468774in}{0.355164in}}%
\pgfpathlineto{\pgfqpoint{0.593277in}{0.355164in}}%
\pgfusepath{stroke}%
\end{pgfscope}%
\begin{pgfscope}%
\pgfpathrectangle{\pgfqpoint{0.375397in}{0.318319in}}{\pgfqpoint{1.556284in}{0.810592in}} %
\pgfusepath{clip}%
\pgfsetrectcap%
\pgfsetroundjoin%
\pgfsetlinewidth{1.505625pt}%
\definecolor{currentstroke}{rgb}{0.239216,0.239216,0.239216}%
\pgfsetstrokecolor{currentstroke}%
\pgfsetdash{}{0pt}%
\pgfpathmoveto{\pgfqpoint{0.468774in}{1.092065in}}%
\pgfpathlineto{\pgfqpoint{0.593277in}{1.092065in}}%
\pgfusepath{stroke}%
\end{pgfscope}%
\begin{pgfscope}%
\pgfpathrectangle{\pgfqpoint{0.375397in}{0.318319in}}{\pgfqpoint{1.556284in}{0.810592in}} %
\pgfusepath{clip}%
\pgfsetrectcap%
\pgfsetroundjoin%
\pgfsetlinewidth{1.505625pt}%
\definecolor{currentstroke}{rgb}{0.239216,0.239216,0.239216}%
\pgfsetstrokecolor{currentstroke}%
\pgfsetdash{}{0pt}%
\pgfpathmoveto{\pgfqpoint{0.842282in}{0.355164in}}%
\pgfpathlineto{\pgfqpoint{0.842282in}{0.355164in}}%
\pgfusepath{stroke}%
\end{pgfscope}%
\begin{pgfscope}%
\pgfpathrectangle{\pgfqpoint{0.375397in}{0.318319in}}{\pgfqpoint{1.556284in}{0.810592in}} %
\pgfusepath{clip}%
\pgfsetrectcap%
\pgfsetroundjoin%
\pgfsetlinewidth{1.505625pt}%
\definecolor{currentstroke}{rgb}{0.239216,0.239216,0.239216}%
\pgfsetstrokecolor{currentstroke}%
\pgfsetdash{}{0pt}%
\pgfpathmoveto{\pgfqpoint{0.842282in}{0.663633in}}%
\pgfpathlineto{\pgfqpoint{0.842282in}{1.063723in}}%
\pgfusepath{stroke}%
\end{pgfscope}%
\begin{pgfscope}%
\pgfpathrectangle{\pgfqpoint{0.375397in}{0.318319in}}{\pgfqpoint{1.556284in}{0.810592in}} %
\pgfusepath{clip}%
\pgfsetrectcap%
\pgfsetroundjoin%
\pgfsetlinewidth{1.505625pt}%
\definecolor{currentstroke}{rgb}{0.239216,0.239216,0.239216}%
\pgfsetstrokecolor{currentstroke}%
\pgfsetdash{}{0pt}%
\pgfpathmoveto{\pgfqpoint{0.780031in}{0.355164in}}%
\pgfpathlineto{\pgfqpoint{0.904534in}{0.355164in}}%
\pgfusepath{stroke}%
\end{pgfscope}%
\begin{pgfscope}%
\pgfpathrectangle{\pgfqpoint{0.375397in}{0.318319in}}{\pgfqpoint{1.556284in}{0.810592in}} %
\pgfusepath{clip}%
\pgfsetrectcap%
\pgfsetroundjoin%
\pgfsetlinewidth{1.505625pt}%
\definecolor{currentstroke}{rgb}{0.239216,0.239216,0.239216}%
\pgfsetstrokecolor{currentstroke}%
\pgfsetdash{}{0pt}%
\pgfpathmoveto{\pgfqpoint{0.780031in}{1.063723in}}%
\pgfpathlineto{\pgfqpoint{0.904534in}{1.063723in}}%
\pgfusepath{stroke}%
\end{pgfscope}%
\begin{pgfscope}%
\pgfpathrectangle{\pgfqpoint{0.375397in}{0.318319in}}{\pgfqpoint{1.556284in}{0.810592in}} %
\pgfusepath{clip}%
\pgfsetrectcap%
\pgfsetroundjoin%
\pgfsetlinewidth{1.505625pt}%
\definecolor{currentstroke}{rgb}{0.239216,0.239216,0.239216}%
\pgfsetstrokecolor{currentstroke}%
\pgfsetdash{}{0pt}%
\pgfpathmoveto{\pgfqpoint{1.153539in}{0.355164in}}%
\pgfpathlineto{\pgfqpoint{1.153539in}{0.355164in}}%
\pgfusepath{stroke}%
\end{pgfscope}%
\begin{pgfscope}%
\pgfpathrectangle{\pgfqpoint{0.375397in}{0.318319in}}{\pgfqpoint{1.556284in}{0.810592in}} %
\pgfusepath{clip}%
\pgfsetrectcap%
\pgfsetroundjoin%
\pgfsetlinewidth{1.505625pt}%
\definecolor{currentstroke}{rgb}{0.239216,0.239216,0.239216}%
\pgfsetstrokecolor{currentstroke}%
\pgfsetdash{}{0pt}%
\pgfpathmoveto{\pgfqpoint{1.153539in}{0.725711in}}%
\pgfpathlineto{\pgfqpoint{1.153539in}{1.037480in}}%
\pgfusepath{stroke}%
\end{pgfscope}%
\begin{pgfscope}%
\pgfpathrectangle{\pgfqpoint{0.375397in}{0.318319in}}{\pgfqpoint{1.556284in}{0.810592in}} %
\pgfusepath{clip}%
\pgfsetrectcap%
\pgfsetroundjoin%
\pgfsetlinewidth{1.505625pt}%
\definecolor{currentstroke}{rgb}{0.239216,0.239216,0.239216}%
\pgfsetstrokecolor{currentstroke}%
\pgfsetdash{}{0pt}%
\pgfpathmoveto{\pgfqpoint{1.091288in}{0.355164in}}%
\pgfpathlineto{\pgfqpoint{1.215791in}{0.355164in}}%
\pgfusepath{stroke}%
\end{pgfscope}%
\begin{pgfscope}%
\pgfpathrectangle{\pgfqpoint{0.375397in}{0.318319in}}{\pgfqpoint{1.556284in}{0.810592in}} %
\pgfusepath{clip}%
\pgfsetrectcap%
\pgfsetroundjoin%
\pgfsetlinewidth{1.505625pt}%
\definecolor{currentstroke}{rgb}{0.239216,0.239216,0.239216}%
\pgfsetstrokecolor{currentstroke}%
\pgfsetdash{}{0pt}%
\pgfpathmoveto{\pgfqpoint{1.091288in}{1.037480in}}%
\pgfpathlineto{\pgfqpoint{1.215791in}{1.037480in}}%
\pgfusepath{stroke}%
\end{pgfscope}%
\begin{pgfscope}%
\pgfpathrectangle{\pgfqpoint{0.375397in}{0.318319in}}{\pgfqpoint{1.556284in}{0.810592in}} %
\pgfusepath{clip}%
\pgfsetrectcap%
\pgfsetroundjoin%
\pgfsetlinewidth{1.505625pt}%
\definecolor{currentstroke}{rgb}{0.239216,0.239216,0.239216}%
\pgfsetstrokecolor{currentstroke}%
\pgfsetdash{}{0pt}%
\pgfpathmoveto{\pgfqpoint{1.464796in}{0.561022in}}%
\pgfpathlineto{\pgfqpoint{1.464796in}{0.464066in}}%
\pgfusepath{stroke}%
\end{pgfscope}%
\begin{pgfscope}%
\pgfpathrectangle{\pgfqpoint{0.375397in}{0.318319in}}{\pgfqpoint{1.556284in}{0.810592in}} %
\pgfusepath{clip}%
\pgfsetrectcap%
\pgfsetroundjoin%
\pgfsetlinewidth{1.505625pt}%
\definecolor{currentstroke}{rgb}{0.239216,0.239216,0.239216}%
\pgfsetstrokecolor{currentstroke}%
\pgfsetdash{}{0pt}%
\pgfpathmoveto{\pgfqpoint{1.464796in}{0.693276in}}%
\pgfpathlineto{\pgfqpoint{1.464796in}{0.815727in}}%
\pgfusepath{stroke}%
\end{pgfscope}%
\begin{pgfscope}%
\pgfpathrectangle{\pgfqpoint{0.375397in}{0.318319in}}{\pgfqpoint{1.556284in}{0.810592in}} %
\pgfusepath{clip}%
\pgfsetrectcap%
\pgfsetroundjoin%
\pgfsetlinewidth{1.505625pt}%
\definecolor{currentstroke}{rgb}{0.239216,0.239216,0.239216}%
\pgfsetstrokecolor{currentstroke}%
\pgfsetdash{}{0pt}%
\pgfpathmoveto{\pgfqpoint{1.402545in}{0.464066in}}%
\pgfpathlineto{\pgfqpoint{1.527047in}{0.464066in}}%
\pgfusepath{stroke}%
\end{pgfscope}%
\begin{pgfscope}%
\pgfpathrectangle{\pgfqpoint{0.375397in}{0.318319in}}{\pgfqpoint{1.556284in}{0.810592in}} %
\pgfusepath{clip}%
\pgfsetrectcap%
\pgfsetroundjoin%
\pgfsetlinewidth{1.505625pt}%
\definecolor{currentstroke}{rgb}{0.239216,0.239216,0.239216}%
\pgfsetstrokecolor{currentstroke}%
\pgfsetdash{}{0pt}%
\pgfpathmoveto{\pgfqpoint{1.402545in}{0.815727in}}%
\pgfpathlineto{\pgfqpoint{1.527047in}{0.815727in}}%
\pgfusepath{stroke}%
\end{pgfscope}%
\begin{pgfscope}%
\pgfpathrectangle{\pgfqpoint{0.375397in}{0.318319in}}{\pgfqpoint{1.556284in}{0.810592in}} %
\pgfusepath{clip}%
\pgfsetbuttcap%
\pgfsetmiterjoin%
\definecolor{currentfill}{rgb}{0.239216,0.239216,0.239216}%
\pgfsetfillcolor{currentfill}%
\pgfsetlinewidth{1.003750pt}%
\definecolor{currentstroke}{rgb}{0.239216,0.239216,0.239216}%
\pgfsetstrokecolor{currentstroke}%
\pgfsetdash{}{0pt}%
\pgfsys@defobject{currentmarker}{\pgfqpoint{-0.029463in}{-0.049105in}}{\pgfqpoint{0.029463in}{0.049105in}}{%
\pgfpathmoveto{\pgfqpoint{-0.000000in}{-0.049105in}}%
\pgfpathlineto{\pgfqpoint{0.029463in}{0.000000in}}%
\pgfpathlineto{\pgfqpoint{0.000000in}{0.049105in}}%
\pgfpathlineto{\pgfqpoint{-0.029463in}{0.000000in}}%
\pgfpathclose%
\pgfusepath{stroke,fill}%
}%
\begin{pgfscope}%
\pgfsys@transformshift{1.464796in}{0.356443in}%
\pgfsys@useobject{currentmarker}{}%
\end{pgfscope}%
\begin{pgfscope}%
\pgfsys@transformshift{1.464796in}{0.355164in}%
\pgfsys@useobject{currentmarker}{}%
\end{pgfscope}%
\begin{pgfscope}%
\pgfsys@transformshift{1.464796in}{0.355164in}%
\pgfsys@useobject{currentmarker}{}%
\end{pgfscope}%
\end{pgfscope}%
\begin{pgfscope}%
\pgfpathrectangle{\pgfqpoint{0.375397in}{0.318319in}}{\pgfqpoint{1.556284in}{0.810592in}} %
\pgfusepath{clip}%
\pgfsetrectcap%
\pgfsetroundjoin%
\pgfsetlinewidth{1.505625pt}%
\definecolor{currentstroke}{rgb}{0.239216,0.239216,0.239216}%
\pgfsetstrokecolor{currentstroke}%
\pgfsetdash{}{0pt}%
\pgfpathmoveto{\pgfqpoint{1.776053in}{0.355164in}}%
\pgfpathlineto{\pgfqpoint{1.776053in}{0.355164in}}%
\pgfusepath{stroke}%
\end{pgfscope}%
\begin{pgfscope}%
\pgfpathrectangle{\pgfqpoint{0.375397in}{0.318319in}}{\pgfqpoint{1.556284in}{0.810592in}} %
\pgfusepath{clip}%
\pgfsetrectcap%
\pgfsetroundjoin%
\pgfsetlinewidth{1.505625pt}%
\definecolor{currentstroke}{rgb}{0.239216,0.239216,0.239216}%
\pgfsetstrokecolor{currentstroke}%
\pgfsetdash{}{0pt}%
\pgfpathmoveto{\pgfqpoint{1.776053in}{0.671448in}}%
\pgfpathlineto{\pgfqpoint{1.776053in}{1.013112in}}%
\pgfusepath{stroke}%
\end{pgfscope}%
\begin{pgfscope}%
\pgfpathrectangle{\pgfqpoint{0.375397in}{0.318319in}}{\pgfqpoint{1.556284in}{0.810592in}} %
\pgfusepath{clip}%
\pgfsetrectcap%
\pgfsetroundjoin%
\pgfsetlinewidth{1.505625pt}%
\definecolor{currentstroke}{rgb}{0.239216,0.239216,0.239216}%
\pgfsetstrokecolor{currentstroke}%
\pgfsetdash{}{0pt}%
\pgfpathmoveto{\pgfqpoint{1.713801in}{0.355164in}}%
\pgfpathlineto{\pgfqpoint{1.838304in}{0.355164in}}%
\pgfusepath{stroke}%
\end{pgfscope}%
\begin{pgfscope}%
\pgfpathrectangle{\pgfqpoint{0.375397in}{0.318319in}}{\pgfqpoint{1.556284in}{0.810592in}} %
\pgfusepath{clip}%
\pgfsetrectcap%
\pgfsetroundjoin%
\pgfsetlinewidth{1.505625pt}%
\definecolor{currentstroke}{rgb}{0.239216,0.239216,0.239216}%
\pgfsetstrokecolor{currentstroke}%
\pgfsetdash{}{0pt}%
\pgfpathmoveto{\pgfqpoint{1.713801in}{1.013112in}}%
\pgfpathlineto{\pgfqpoint{1.838304in}{1.013112in}}%
\pgfusepath{stroke}%
\end{pgfscope}%
\begin{pgfscope}%
\pgfpathrectangle{\pgfqpoint{0.375397in}{0.318319in}}{\pgfqpoint{1.556284in}{0.810592in}} %
\pgfusepath{clip}%
\pgfsetrectcap%
\pgfsetroundjoin%
\pgfsetlinewidth{1.505625pt}%
\definecolor{currentstroke}{rgb}{0.239216,0.239216,0.239216}%
\pgfsetstrokecolor{currentstroke}%
\pgfsetdash{}{0pt}%
\pgfpathmoveto{\pgfqpoint{0.406523in}{0.568148in}}%
\pgfpathlineto{\pgfqpoint{0.655528in}{0.568148in}}%
\pgfusepath{stroke}%
\end{pgfscope}%
\begin{pgfscope}%
\pgfpathrectangle{\pgfqpoint{0.375397in}{0.318319in}}{\pgfqpoint{1.556284in}{0.810592in}} %
\pgfusepath{clip}%
\pgfsetrectcap%
\pgfsetroundjoin%
\pgfsetlinewidth{1.505625pt}%
\definecolor{currentstroke}{rgb}{0.239216,0.239216,0.239216}%
\pgfsetstrokecolor{currentstroke}%
\pgfsetdash{}{0pt}%
\pgfpathmoveto{\pgfqpoint{0.717780in}{0.600373in}}%
\pgfpathlineto{\pgfqpoint{0.966785in}{0.600373in}}%
\pgfusepath{stroke}%
\end{pgfscope}%
\begin{pgfscope}%
\pgfpathrectangle{\pgfqpoint{0.375397in}{0.318319in}}{\pgfqpoint{1.556284in}{0.810592in}} %
\pgfusepath{clip}%
\pgfsetrectcap%
\pgfsetroundjoin%
\pgfsetlinewidth{1.505625pt}%
\definecolor{currentstroke}{rgb}{0.239216,0.239216,0.239216}%
\pgfsetstrokecolor{currentstroke}%
\pgfsetdash{}{0pt}%
\pgfpathmoveto{\pgfqpoint{1.029037in}{0.456387in}}%
\pgfpathlineto{\pgfqpoint{1.278042in}{0.456387in}}%
\pgfusepath{stroke}%
\end{pgfscope}%
\begin{pgfscope}%
\pgfpathrectangle{\pgfqpoint{0.375397in}{0.318319in}}{\pgfqpoint{1.556284in}{0.810592in}} %
\pgfusepath{clip}%
\pgfsetrectcap%
\pgfsetroundjoin%
\pgfsetlinewidth{1.505625pt}%
\definecolor{currentstroke}{rgb}{0.239216,0.239216,0.239216}%
\pgfsetstrokecolor{currentstroke}%
\pgfsetdash{}{0pt}%
\pgfpathmoveto{\pgfqpoint{1.340293in}{0.636440in}}%
\pgfpathlineto{\pgfqpoint{1.589299in}{0.636440in}}%
\pgfusepath{stroke}%
\end{pgfscope}%
\begin{pgfscope}%
\pgfpathrectangle{\pgfqpoint{0.375397in}{0.318319in}}{\pgfqpoint{1.556284in}{0.810592in}} %
\pgfusepath{clip}%
\pgfsetrectcap%
\pgfsetroundjoin%
\pgfsetlinewidth{1.505625pt}%
\definecolor{currentstroke}{rgb}{0.239216,0.239216,0.239216}%
\pgfsetstrokecolor{currentstroke}%
\pgfsetdash{}{0pt}%
\pgfpathmoveto{\pgfqpoint{1.651550in}{0.457511in}}%
\pgfpathlineto{\pgfqpoint{1.900556in}{0.457511in}}%
\pgfusepath{stroke}%
\end{pgfscope}%
\begin{pgfscope}%
\pgfsetrectcap%
\pgfsetmiterjoin%
\pgfsetlinewidth{0.803000pt}%
\definecolor{currentstroke}{rgb}{0.000000,0.000000,0.000000}%
\pgfsetstrokecolor{currentstroke}%
\pgfsetdash{}{0pt}%
\pgfpathmoveto{\pgfqpoint{0.375397in}{0.318319in}}%
\pgfpathlineto{\pgfqpoint{0.375397in}{1.128910in}}%
\pgfusepath{stroke}%
\end{pgfscope}%
\begin{pgfscope}%
\pgfsetrectcap%
\pgfsetmiterjoin%
\pgfsetlinewidth{0.803000pt}%
\definecolor{currentstroke}{rgb}{0.000000,0.000000,0.000000}%
\pgfsetstrokecolor{currentstroke}%
\pgfsetdash{}{0pt}%
\pgfpathmoveto{\pgfqpoint{1.931681in}{0.318319in}}%
\pgfpathlineto{\pgfqpoint{1.931681in}{1.128910in}}%
\pgfusepath{stroke}%
\end{pgfscope}%
\begin{pgfscope}%
\pgfsetrectcap%
\pgfsetmiterjoin%
\pgfsetlinewidth{0.803000pt}%
\definecolor{currentstroke}{rgb}{0.000000,0.000000,0.000000}%
\pgfsetstrokecolor{currentstroke}%
\pgfsetdash{}{0pt}%
\pgfpathmoveto{\pgfqpoint{0.375397in}{0.318319in}}%
\pgfpathlineto{\pgfqpoint{1.931681in}{0.318319in}}%
\pgfusepath{stroke}%
\end{pgfscope}%
\begin{pgfscope}%
\pgfsetrectcap%
\pgfsetmiterjoin%
\pgfsetlinewidth{0.803000pt}%
\definecolor{currentstroke}{rgb}{0.000000,0.000000,0.000000}%
\pgfsetstrokecolor{currentstroke}%
\pgfsetdash{}{0pt}%
\pgfpathmoveto{\pgfqpoint{0.375397in}{1.128910in}}%
\pgfpathlineto{\pgfqpoint{1.931681in}{1.128910in}}%
\pgfusepath{stroke}%
\end{pgfscope}%
\end{pgfpicture}%
\makeatother%
\endgroup%

%% file: conclusion.tex
\section{Conclusion}

We introduced GitGraph, the first corpus of neural computation graphs. The first goal of GitGraph is to serve as a knowledge repository that allows for an automated search of neural architectures that solve a specific problem. 
Using the search functionality, we can obtain a set of distinct architectures for problems related to the searched one.
From the found architectures, in the form of computation graphs, we created a method of generating unique relevant common subgraphs.

The main aim of finding problem-specific GitGraph common subgraphs is to create a neural search space. Instead of searching or using reinforcement or evolution strategies using the basic building blocks, we reduce the complexity and cost of the subsequent neural architecture creation policy by optimizing the search space itself.

We show that the GitGraph common subgraphs cover between 20 and 40\% of the nodes in their source graphs. 
Given the obtained complexity reduction, we believe they will be a basis for large problem-specific modules in future automated neural creation strategies.

%% file: appendix.tex
\newpage
\vspace{-1em}
\appendix

\section{Database of tensorflow graphs}

All the meta checkpoints (from which we can extract the Tensorflow graphs) are contained in the following multi-part zip.

Download all the files below in the same directory and uncompress the main zip (Approx: 3go compressed , 15go uncompressed)
\\
\href{https://www.mycloud.ch/s/S00694BE7C11AA77D3491DEB8CE9BF68689E9D407AC}{Main zip}
\\
\href{https://www.mycloud.ch/s/S00633B82DF94EED9C535C7471B662725AC5ABE5951}{Part 1}
\\
\href{https://www.mycloud.ch/s/S0048436DE52EAFD5A20333311EEA260ED533B5938B}{Part 2}
\\
\href{https://www.mycloud.ch/s/S0032BE7A647D2EBB7D1A3035A13601F6727AF4147E}{Part 3}
\\
\href{https://www.mycloud.ch/s/S0080E12DDFEE75605057E600200206FC853ED88EB9}{Part 4}
\\
\href{https://www.mycloud.ch/s/S003617AC44A165E2170B2C5A6DFA4022E4B5CB6639}{Part 5}
\\
\href{https://www.mycloud.ch/s/S003AD2D8D49A8B386061712470908334DC691B6238}{Part 6}

\section{Examples of frequent subgraphs}
\label{subfreq_example}

\begin{figure*}[!ht]
\begin{center}
\subfloat[]{
	\includegraphics[width=0.5\textwidth]{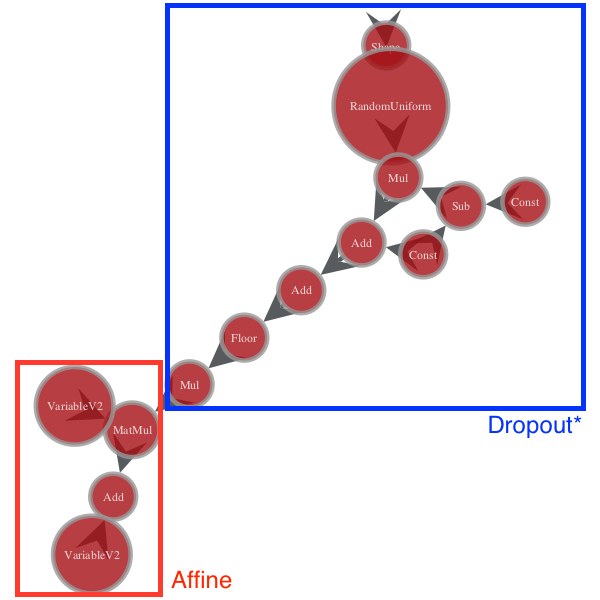}
}
\subfloat[]{
	\includegraphics[width=0.5\textwidth]{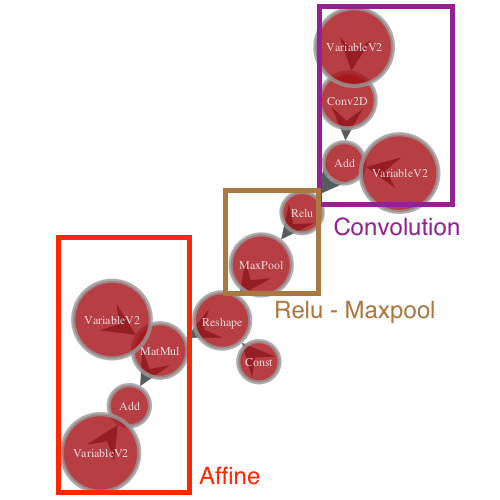}
	\label{conv_example}
}
\end{center}
\caption{Example of frequent subgraphs for the image dataset}
\end{figure*}

\begin{figure*}
\begin{center}
\subfloat[]{
	\includegraphics[width=0.5\textwidth]{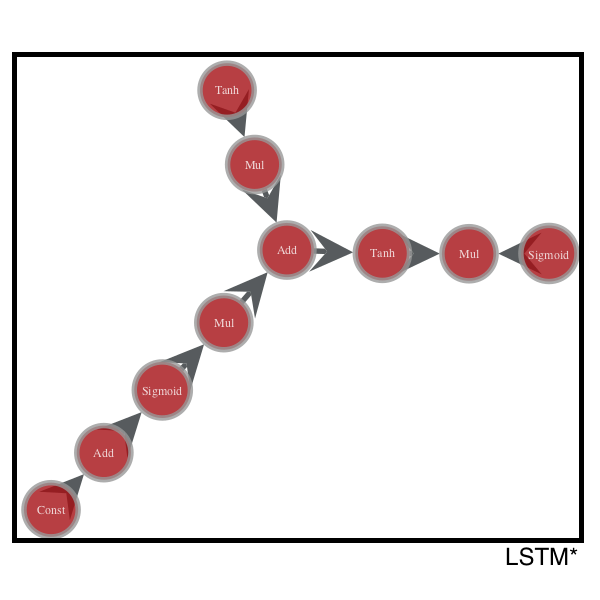}
	\label{fig-example-LSTM}
}
\subfloat[]{
	\includegraphics[width=0.5\textwidth]{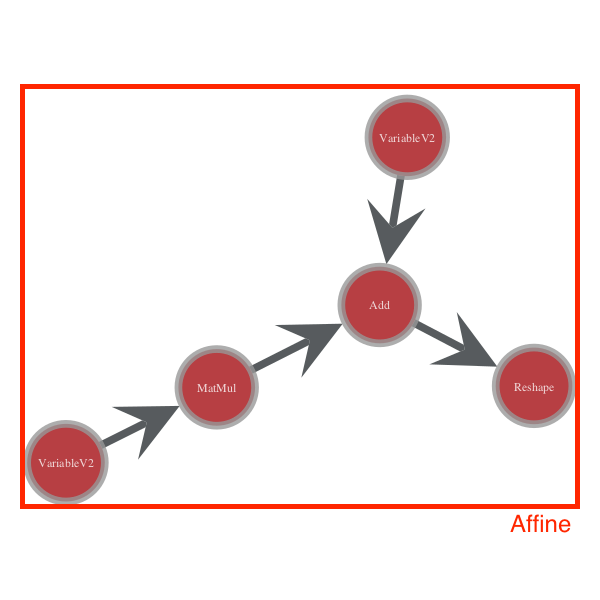}
}
\end{center}
\caption{Example of frequent subgraphs for the text dataset}
\end{figure*}

\begin{figure}
\subfloat[]{
	\includegraphics[width=0.5\textwidth]{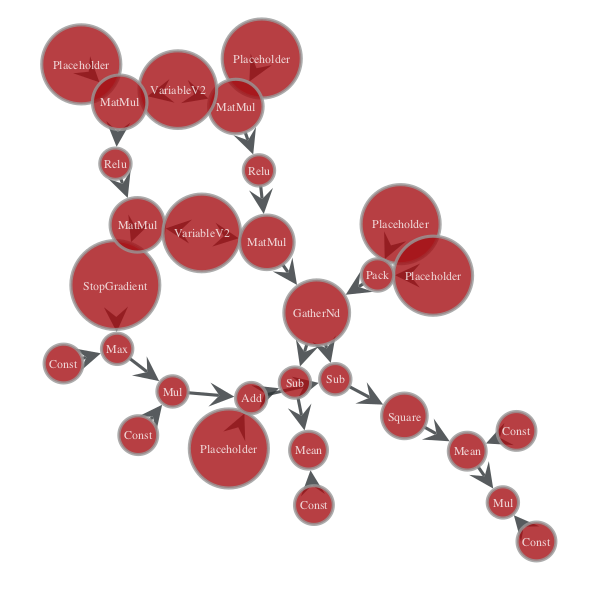}
	\label{fig-example-reinforcement}
}
\caption{Frequent subgraph for the reinforcement dataset}
\end{figure}

